\pdfoutput=1
\documentclass[11pt]{article}

\usepackage{EACL2023}

\usepackage{times}
\usepackage{latexsym}
\usepackage{graphicx}
\usepackage{booktabs}
\usepackage{enumitem}
\usepackage{multirow}

\usepackage[T1]{fontenc}
\usepackage{inputenc}

\usepackage{microtype}

\usepackage{inconsolata}

\usepackage{lipsum}
\usepackage{amsmath}

\DeclareMathOperator*{\argmax}{arg\,max} % Jan Hlavacek

\setlist{topsep=1pt,itemsep=1pt,partopsep=1pt, parsep=1pt}

\title{
LLMs Are Few-Shot In-Context Low-Resource Language Learners
}

\author{Samuel Cahyawijaya \\
  HKUST \\
  \texttt{scahyawijaya@connect.ust.hk} \\\And
  Holy Lovenia \\
  AI Singapore \\
  \texttt{holy@aisingapore.org} \\\And
  Pascale Fung \\
  HKUST \\
  \texttt{pascale@ece.ust.hk} \\ \\}

\begin{document}

\maketitle
\begin{abstract}

In-context learning (ICL) empowers large language models (LLMs) to perform diverse tasks in underrepresented languages using only short in-context information, offering a crucial avenue for narrowing the gap between high-resource and low-resource languages.
Nonetheless, there is only a handful of works explored ICL for low-resource languages with most of them focusing on relatively high-resource languages, such as French and Spanish. 
In this work, we extensively study ICL and its cross-lingual variation (X-ICL) on 25 low-resource and 7 relatively higher-resource languages.
Our study not only assesses the effectiveness of ICL with LLMs in low-resource languages but also identifies the shortcomings of in-context label alignment, and introduces a more effective alternative: query alignment. Moreover, we provide valuable insights into various facets of ICL for low-resource languages.
Our study concludes the significance of few-shot in-context information on enhancing the low-resource understanding quality of LLMs through semantically relevant information by closing the language gap in the target language and aligning the semantics between the targeted low-resource and the high-resource language that the model is proficient in. Our work highlights the importance of advancing ICL research, particularly for low-resource languages. Our code is publicly released at \url{https://github.com/SamuelCahyawijaya/in-context-alignment}.

% \pascale{We study the effectiveness of cross-lingual in-context learning (X-ICL) with large language models (LLMs) on low source languages with a new proposed query alignment method. X-ICL empowers LLMs to generalize to low resource languages using only a few cross-lingual samples, narrowing the gap between high and low resource languages. In this work, we extensively study X-ICL on 25 low-resource languages and 7 high-resource languages with multiple approaches of X-ICL. We show that, unlike previous work, ???? compared to ??  }
\end{abstract}

% we show that coupling similarly semantic exemplars in the

\section{Introduction}

Large language models (LLMs) have displayed remarkable generalization capability in various tasks~\cite{brown2020language,kojima2022large,wei2022chain,smith2022megatron,rae2022scaling,chowdhery2022palm,scao2022bloom,liang2023holistic,srivastava2023beyond,lovenia2023negative,bang2023multitask}. Nonetheless, these models face difficulties in generalizing across different languages, leading to performance disparity, particularly for low-resource languages~\cite{aji2022one,ebrahimi2022americasnli,adelani2022masakhaner2,cahyawijaya2023nusacrowd,cahyawijaya2023nusawrites,asai2023buffet}. A myriad of research works address this problem through language-specific fine-tuning~\cite{wilie2020indonlu,kakwani2020indicnlpsuite,cahyawijaya2021indonlg,adelani2021masakhaner,kumar2022indicnlg} which often leads to catastrophic forgetting~\cite{robert1993catastrophic,chaudhry2019tiny,david2019experiencereplay}. Another line of work utilizes continual learning and adapter-based methods to inject new languages to existing LLMs~\cite{yong2022bloom1,cahyawijaya2023instructalign,jin2023dataless}. 
% Closing the performance gap between languages is essential for expanding inclusivity and diversity in Natural Language Processing (NLP), particularly for low-resource and low-resource languages~\cite{cahyawijaya2021indonlg,aji2022one,adelani2022masakhaner2,kumar2022indicnlg,ebrahimi2022americasnli}. 
% Nevertheless, injecting new languages to LLMs through fine-tuning is computationally expensive and often lead to catastrophic forgetting~\cite{yong2022bloom1,cahyawijaya2023instructalign,jin2023dataless}. While continual learning methods can alleviate catastrophic forgetting, these approaches incur significant computational overhead prominently for LLMs. 
Nevertheless, these methods rely on performing multiple steps of parameter updates which require huge computational budgets, particularly for very large LLMs with hundreds of billion parameters. 

To cope with this problem, prior works~\cite{winata2021language,lin2022shot,shi2023language,zhang2023multilingual} explore cross-lingual in-context learning (X-ICL) methods, an extension from in-context learning (ICL), that allow LLMs to generate better response quality in low-resource languages without the need for parameter tuning. 
% Through cross-lingual in-context learning, the model is able to better understand the task by learning from the source language exemplars.
In X-ICL, source language exemplars are incorporated into the input context allowing the model to transfer the task understanding capability from the source, commonly high-resource, language into the target language query~\cite{winata2021language,shi2023language}. However, X-ICL still fails to compete with a simple translate-test baseline, prominently for low-resource languages. A recent work~\cite{tanwar2023multilingual} further enhances X-ICL through semantically similar cross-lingual exemplars and in-context label alignment\footnote{In \citet{tanwar2023multilingual}, label alignment is referred to as task alignment. In this work, we distinguish two types of task alignments, i.e., query alignment and label alignment (\S\ref{sec:cross-lingual-alignment}).}, yielding a large gain over the baselines on relatively high-resource languages such as French, Spanish, Chinese, and Japanese.

% aside from the in-context cross-lingual exemplars allowing the model to align the label of the source language to the target language. This method shows significant performance improvement compared to regular cross-lingual in-context learning approaches on various cross-lingual classification tasks.

% cross-lingual semantic similarity to search the source language exemplars and 

% and we refer to this mapping phenomenon as in-context mapping. This behavior is similar to (Copy Mechanism), which.....

% Despite the great potential, naively adapting new languages to existing LLMs is costly and often leads to catastrophic forgetting~\cite{yong2022bloom1,cahyawijaya2023instructalign}. Prior works have explored various methods to alleviate these problems, including adapter-based approach~\cite{yong2022bloom1}, cross-lingual alignment~\cite{cahyawijaya2023instructalign}, and distillation~\cite{li2023bactrianx}. However, these approaches still require a relatively huge computation budget for adapting new languages resulting in limited adoption of LLMs, particularly in low-resource languages.

% Limitation

% Solution - ICT?

% In this work, we extend the idea of cross-lingual semantic similarity and in-context label alignment to low-resource languages. We hypothesize that this method will not work as well in low-resource languages due to the weak label and sentence representation of the target languages.
In this work, we expand upon the concept of cross-lingual semantic similarity and in-context alignment, specifically focusing on low-resource languages. Our hypothesis posits that their effectiveness may be compromised in low-resource languages due to the weak representation of the labels and sentences for the target languages.
To test our hypothesis, we explore cross-lingual in-context learning (X-ICL) covering 25 low-resource languages from various language families and compare them with the performance of 7 relatively higher resource languages, including French (fra), Spanish (spa), German (deu), Italian (ita), Portuguese (por), Arabic (arb), and Hindi (hin). Our result suggests that the X-ICL performance decays correlate to the size of pre-training data of the target languages, which aligns with our hypothesis. Moreover, to our surprise, contrary to the results reported in~\cite{tanwar2023multilingual}, we found that in-context label alignment does not work for all the languages under study and introduced an alternative alignment method namely in-context query alignment, which significantly improves the alignment quality compared to the in-context label alignment.

To this end, we explore alternatives for X-ICL approaches covering variations of in-context alignment information, prompt, label encoding, and strategy for selecting in-context learning exemplars. We extensively analyze all these factors and their effect on the downstream task performance of all the languages under study. Our results and analysis highlight the following key takeaways: 

\begin{itemize}
    \item Contrary to prior work~\cite{tanwar2023multilingual}, we found that label alignment undermines the performance in most languages. Keeping uniform labels from the high-resource language often yields the best results.
    \item We introduce a new approach for cross-lingual alignment, i.e., query alignment, which is more effective than label alignment and can substitute or complement X-ICL.
    \item We analyze the effect of improving prompt format consistency on low-resource languages. However, despite improving performance for higher-resource languages, format consistency does not yield any benefit to the low-resource languages under study.
    % \item We showcase the competing performance between the use of ICL and X-ICL for improving LLMs' understanding of low-resource languages.
    % Our result displays the potential of incorporating X-ICL in the case where there is no in-language data and machine translation available for the specified language.
    % Our findings reveal the efficacy of using X-ICL, particularly when there is no available high-quality machine translation model for the specified language.
    \item We present a comprehensive in-context learning framework for better low-resource language understanding under various constrained conditions, concluding the significance of few-shot in-context information on enhancing the low-resource understanding quality of LLMs through semantically relevant information, where \textbf{monolingual ICL} does so by closing the language and domain gap on the targeted downstream task, while \textbf{X-ICL} closes the domain gap to the target downstream task, and \textbf{in-context alignment} closes the semantic gap between the targeted low-resource and the high-resource language that the model is proficient in.
    
    % \item We present a framework for improving low-resource language understanding capability under various constrained conditions, where we can apply: 1) \textbf{translate-test ICL} when a strong machine translation system is available for the target low-resource language to a high-resource language, 2) \textbf{ICL} and \textbf{X-ICL} with \textbf{semantic-similarity-based exemplar selection} when only a monolingual or cross-lingual semantic similarity model and task-specific data are available, 3) \textbf{ICL} and \textbf{X-ICL} with \textbf{random exemplar selection} when both machine translation and semantic similarity models are not available but monolingual or cross-lingual task-specific data is available, and 4) \textbf{in-context query alignment} when only a handful of translation data from the target low-resource to other high-resource language is  available.
    
    % \item We analyze the quality of various semantic similarity methods, and we showcase their performance bottleneck on low-resource languages. Despite the bottleneck, we find that simply incorporating  different semantic similarity models bring significant benefits for in-context learning, for both low-resource and relatively high-resource languages.
\end{itemize}

\section{Related Work}

\subsection{In-Context Learning}

The in-context learning paradigm, originally introduced by \citet{brown2020gpt3}, has significantly advanced our understanding of LLMs' capabilities. It demonstrated that LLMs can effectively perform complex tasks through in-context learning with just task-specific formatting and a few task-specific examples (few-shot) or none at all (zero-shot). This ability is facilitated by the LLMs' increasing capacity for generalization across diverse tasks, e.g., machine translation, question answering, and domain adaptation, without gradient updates.

Another line of work expands the scope of the study to multilingual generative LLMs, i.e., BLOOM~\cite{scao2022bloom}, trained on the ROOTS corpus covering 46 natural and 13 programming languages, and XGLM~\cite{lin2022fewshot}, trained on 500B tokens comprising 30 languages, which exhibit robust zero-shot and few-shot performances on multilingual NLP tasks. Furthermore,~\citet{lin2022fewshot} address the imbalance in language representation by up-sampling the less-resourced languages.
% XGLM exhibits robust zero-shot and few-shot performances across 20 languages in tasks such as multilingual commonsense reasoning, natural language inference, and machine translation.
\citet{bandarkar2023belebele} then expand the language coverage of the in-context learning evaluation to 122 languages through Belebele, a wide-scale multilingual multiple-choice machine reading comprehension benchmark comprising short passages from FLORES-200~\cite{goyal-etal-2022-flores}.

% - GPT3

% - BLOOM

% - XGLM

% - Falcon

% - LLaMA2

% - Belebele

\begin{figure*}[t]
    \centering
    \resizebox{\linewidth}{!}{
     \includegraphics[trim=0 0.6em 0 0,width=\textwidth,clip]{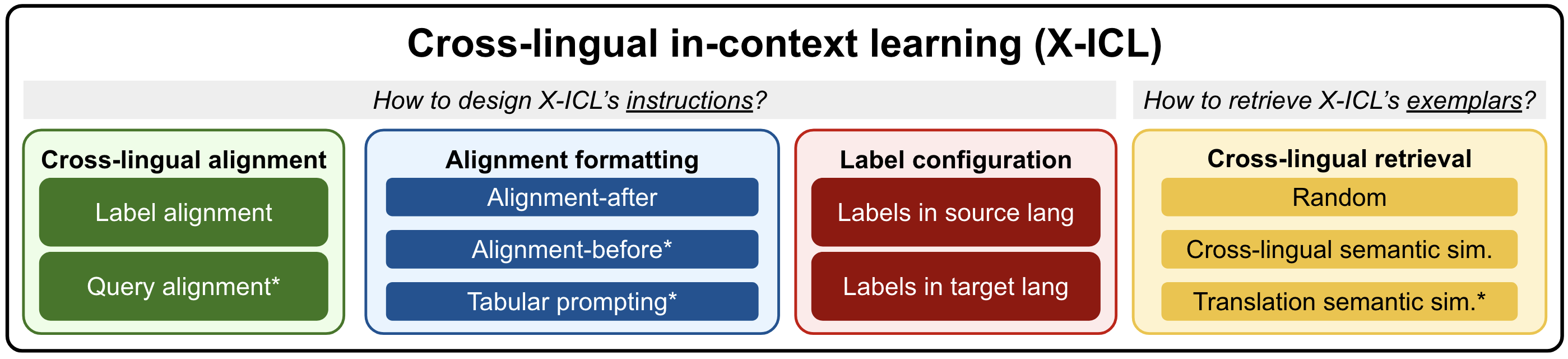}}
    \caption{Framework of cross-lingual in-context learning (X-ICL) methods analyzed in our work. We explore various cross-lingual retrieval methods with different kinds of cross-lingual prompting strategies. $^*$Novel approaches.}
    \vspace{-6pt}
    \label{fig:x-icl-framework}
\end{figure*}

\paragraph{Cross-Lingual In-Context Learning (X-ICL)}

\citet{winata2021language} were among the first to explore the potential of few-shot X-ICL. Using 4 high-resource languages, this work shows that pre-trained LMs significantly outperform random prediction in cross-lingual tasks and produce better results compared to smaller fine-tuned baselines.~\citet{winata-etal-2022-cross} expand this study to unseen languages and find that taking the X-ICL contexts from a mixture of random source languages is surprisingly more effective compared to linguistically similar and geographically similar languages. Expanding the investigation to different aspects of cross-lingual transfers in X-ICL, \citet{lin2022fewshot} explores the use of various languages for instructions and exemplars. 
They find that 
% source language exemplars significantly enhance cross-lingual zero-shot performance in the target language. Moreover, 
incorporating English instructions notably improves zero-shot performance across multiple languages.
% However, these positive effects do not add up.

In a related vein,~\citet{tanwar2023multilingual} analyze the effect of cross-lingual prompt design for X-ICL across 3 text classification tasks using 44 different cross-lingual pairs. Their findings emphasize the limitations of random exemplar selection and propose the use of semantic-based exemplar retrieval and label alignment\footnotemark[1] for superior X-ICL performance. Notably, their findings diverge from our results (\S\ref{sec:alignment}), which contend that label alignment does not provide benefits for X-ICL.

% - Winata (Check Intro)

% - Lin (Check Intro)~\cite{lin2022fewshot}

% - Shi (Check Intro)~\cite{shi2023language}

% - Zhang (Check Intro)

% - Tanwar (Check Intro)~\cite{tanwar2023multilingual}

% - Similarity to RAG (???)

\subsection{LLMs on Low-Resource Languages}

% While the generalization skills of LLMs enable LLMs to utilize knowledge transfer to perform multiple tasks in unseen languages, previous works show that 

% Previous studies showing that LMs trained on multiple languages in general benefit from positive transfer between learning related languages, and this is especially beneficial for understanding low-resource languages. However, due to English having higher magnitudes in size compared to the other languages, transferring models trained on resource-rich languages (e.g., English) to other languages has been actively studied in multilingual NLP.

Rigorous evaluations have been proposed to investigate how LLMs perform on low-resource languages.
% In their study,~\citet{cahyawijaya2023nusacrowd} present NusaNLU and NusaNLG, zero-shot benchmarks covering 19 Indonesian indigenous languages.
According to~\cite{cahyawijaya2023nusacrowd}, while multilingual LLMs typically exhibit positive transfer learning among related languages, these models perform notably better for mid- and high-resource (e.g., Indonesian and English) compared to low-resource languages (e.g., other 18 Indonesian indigenous languages). This implies a challenge in the generalization capability of existing multilingual LLMs to low-resource languages. This is further evidenced by~\cite{cahyawijaya2023nusawrites}, which extends the exploration to 12 low-resource and extremely low-resource languages, where both existing zero-shot prompting LLMs and fine-tuned pre-trained LMs struggle to outperform classical machine learning baselines, which is indicative of LLMs' limited ability to generalize to extremely low-resource languages that are significantly distinct from those encountered during their training. Similar observations have been reported by \citet{asai2023buffet,bang2023multitask,adilazuarda2024lingualchemy} for lower-resource languages. 
% \citet{asai2023buffet} additionally note larger performance drops in models employing English in-context learning, i.e., utilizing English instructions and demonstrations in target languages, on lower-resource languages compared to on high-resource languages.
Furthermore, another line of work emphasizes the challenges faced by multilingual LLMs in understanding~\cite{zhang2023multilingual,adilazuarda2023obscure} and generating~\cite{yong2023prompting} code-switching, a real use case and nuance of multilingualism exhibited by human speakers.

\section{Methods}

% Exemplars DB ==> SS == Exs => PF + AI => X-ICL Prompt
Figure~\ref{fig:x-icl-framework} presents the general framework of X-ICL, comprising: 1) cross-lingual in-context alignment (\S\ref{sec:cross-lingual-alignment}), 2) cross-lingual alignment formatting and 3) label configuration as parts of cross-lingual prompting (\S\ref{sec:cross-lingual-prompting}), as well as 4) cross-lingual retrieval (\S\ref{sec:cross-lingual-retrieval}). We assess variations of these X-ICL aspects to understand their effectiveness on different language resource levels.

% There are multiple approaches to performing X-ICL. These approaches mainly rely on two factors: 1) how to determine the exemplars in context and 2) how to design the instructions. In this work, we describe two semantic similarity methods, i.e., a) cross-lingual and b) translation, to retrieve the exemplars in the source language. As for the instruction design, we categorize the existing methods into: a) X-ICL without alignment, b) X-ICL with alignment, and c) X-ICL with alignment and formatting. All approaches are explained in detail in the following subsections.

\begin{figure*}[!t]
    \centering
    \resizebox{0.86\linewidth}{!}{
     \includegraphics[width=\textwidth]{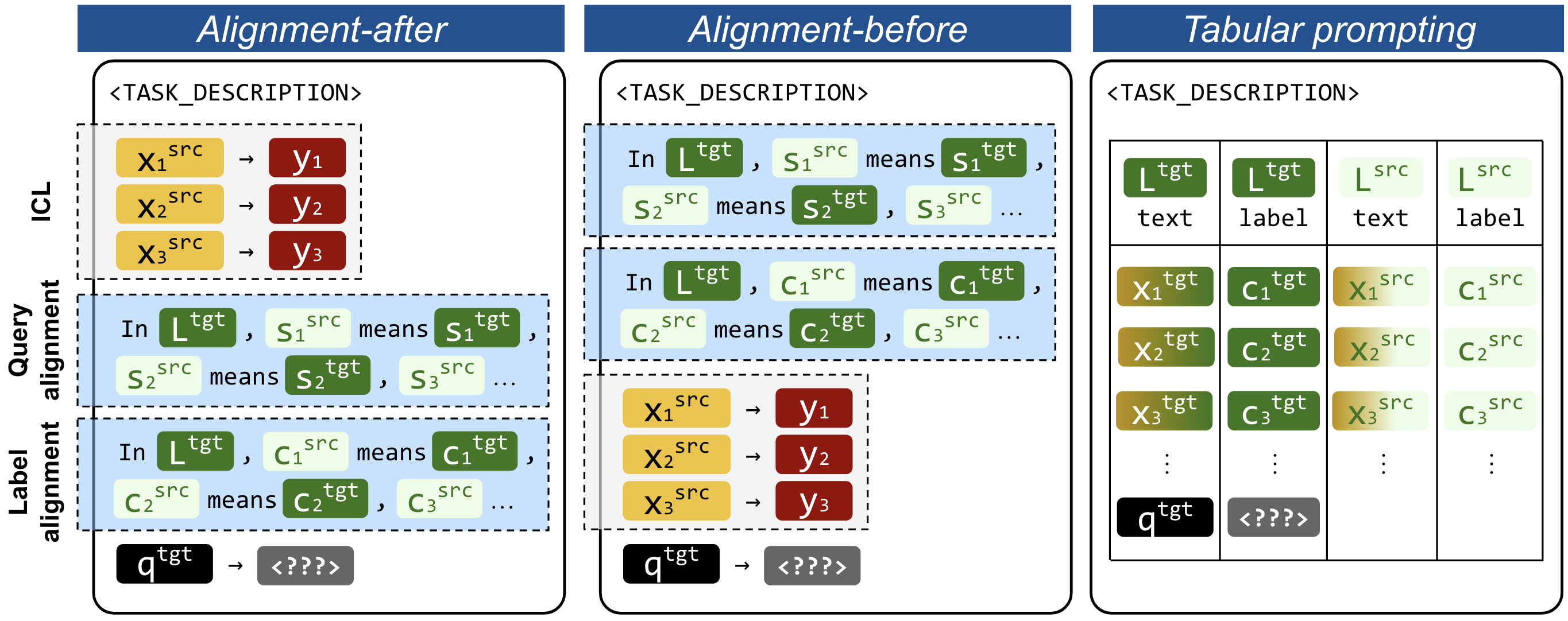}}
    \caption{We explore 3 different alignment formats for X-ICL prompting, i.e., alignment-after, alignment-before, and tabular-prompting. From left to right, the prompt format has a higher degree of format consistency.}
    \label{fig:prompt-format}
\end{figure*}

\subsection{Cross-Lingual Alignment}
\label{sec:cross-lingual-alignment}

Prior works showcase the benefit of cross-lingual in-context learning with random exemplars which can improve the zero-shot performance of LLMs on downstream tasks~\cite{winata-etal-2021-language,shi2023language,asai2023buffet}. More recently, \citet{tanwar2023multilingual} introduce cross-lingual in-context alignment that injects a label aligner to the prompt in between the in-context exemplars and the input query. The label aligner provides the translation of the source label set $C^{src}=\{c^{src}_1,c^{src}_2,\dots,c^{src}_k\}$ to the target label set $C^{tgt}=\{c^{tgt}_1,c^{tgt}_2,\dots,c^{tgt}_k\}$. For instance, given a target language $L^{tgt}$, the label aligner prompt is formatted as follow: \texttt{``In \textcolor{olive}{$L^{tgt}$}, \textcolor{red}{$c^{src}_1$} means \textcolor{blue}{$c^{tgt}_1$}, \textcolor{red}{$c^{src}_2$} means \textcolor{blue}{$c^{tgt}_2$},$\dots$, and \textcolor{red}{$c^{src}_k$} means \textcolor{blue}{$c^{tgt}_k$}''}. This allows the model to align labels between source and target languages. We call this method \textbf{in-context label alignment}.

As opposed to in-context label alignment, we explore another approach, dubbed \textbf{in-context query alignment}, which provides alignment of input distribution by providing the translation of sentences similar to the query while keeping the label set as is. To do so, we utilize the parallel exemplar dataset $D^{para}=\{(s^{src}_1, s^{tgt}_1),(s^{src}_2, s^{tgt}_2),\dots,(s^{src}_m, s^{tgt}_m)\}$, where $(s^{src}_i, s^{tgt}_i)$ respectively denotes to a pair of parallel source and target sentences, and select the top-k most similar parallel pair by maximizing the monolingual similarity between the $q^{tgt}$ with $s^{tgt}_i$. Given a target language $L^{tgt}$, the parallel pairs are then formatted into an input alignment prompt, i.e., \texttt{``In \textcolor{olive}{$L^{tgt}$}, \textcolor{red}{$s^{src}_1$} means \textcolor{blue}{$s^{tgt}_1$}, \textcolor{red}{$s^{src}_2$} means \textcolor{blue}{$s^{tgt}_2$}, $\dots$, and \textcolor{red}{$s^{src}_k$} means \textcolor{blue}{$s^{tgt}_k$}''}. We show the example query for in-context label alignment and in-context query alignment in Appendix~\ref{app:alignment}.

\subsection{Cross-Lingual Prompting}
\label{sec:cross-lingual-prompting}

Although cross-lingual in-context alignment has shown improvements as reported in~\cite{tanwar2023multilingual}, it introduces distortions to certain aspects of the Bayesian inference framework~\cite{xie2022icl,min2022rethinking} underlying in-context learning. Notably, this method compromises the formatting consistency and output distribution of the prompt. While the label aligner is expected to align the output distribution between the source and target labels, it is merely an idealistic assumption, which might not be the case in real cases.
% case that might not happen perfectly in some cases.
To better align with the Bayesian inference framework, we explore two cross-lingual prompt adjustments, i.e., alignment formatting and label configuration.

\paragraph{Alignment Formatting}
\label{sec:alignment-formatting}

Existing X-ICL with alignment approach~\cite{tanwar2023multilingual} places the alignment between ICL exemplars and the input query (dubbed as \textbf{alignment-after}). We argue that such abrupt changes in the prompt format might cause performance degradation. To improve the format consistency, we also explore two prompt formats for X-ICL: 1) \textbf{alignment-before} and 2) \textbf{tabular-prompting}. Alignment-before simply swaps the alignment text with the cross-lingual exemplars. This avoids the abrupt format change between the exemplars and the query such that the neighboring text span is more format-consistent. Tabular-prompting formats the prompt in the form of a table with multiple columns, which allows a consistent prompt format, but at the same time requires either a labeled parallel corpus or disrupting the input-output mapping through incorrect labeling~\cite{min2022rethinking}. The depiction of the prompt formats and the X-ICL alignment is in Figure~\ref{fig:prompt-format}.

\paragraph{Label Configuration}
\label{sec:label-configuration}

To improve the output distribution consistency, we explore alternatives of using either \textbf{source-only labels} and \textbf{target-only labels} as opposed to \textbf{in-context label alignment} which shifts the language of the labels from the source language in the exemplars to the target language in $q^{tgt}$. These alternatives serve as a comparison to measure the effectiveness of in-context label alignment. In this study, we focus on English as the source language.
% as it yields better performance.
Appendix~\ref{app:source-lang} analyzes the use of other closely related source languages.

\subsection{Cross-Lingual Retrieval}
\label{sec:cross-lingual-retrieval}

Another way to improve X-ICL performance is by improving the exemplar retrieval quality. Given an input query $q^{tgt}$ and a source language exemplar dataset $D^{src}=\{(e^{src}_1, y^{src}_1),(e^{src}_2, y^{src}_2),\dots,(e^{src}_n, y^{src}_n)\}$, where $e^{src}_1$ and $y^{src}_1$ respectively denote the input and label of the exemplar, the goal of cross-lingual retrieval is to retrieve one or more labeled exemplars $(e^{src}_i, y^{src}_i)$ semantically relevant to $q^{tgt}$. Most prior works in X-ICL~\cite{winata-etal-2021-multilingual,asai2023buffet,zhang2023multilingual,lin2022shot} incorporate random retrieval, while recently, \citet{tanwar2023multilingual} utilize \textbf{cross-lingual semantic similarity} which significantly improves performance over the random retrieval.

Nevertheless, we argue that this approach might not be optimal in the case of low-resource languages as the semantic representation for these languages might not be well aligned with the high-resource languages (see Appendix~\ref{app:xss}). Thus, we explore \textbf{translation semantic similarity} as an alternative. It performs monolingual semantic similarity on $q^{tgt}$ to obtain a sentence in $L^{tgt}$ from a parallel dataset $D^{para}$, then uses monolingual semantic similarity on its pair in $L^{src}$ to find the high-resource exemplars from $D^{src}$. Although the monolingual semantic similarity between two sentences from a low-resource language is also suboptimal, this problem can be alleviated by incorporating other similarity metrics such as TF-IDF and bag-of-words. We denote ICL method using the translation semantic similarity as \textbf{T-ICL}. We show this analysis in Appendix~\ref{app:xss} along with the depiction of the cross-lingual retrieval methods. 

\begin{table}[t!]
    \centering
    \resizebox{\linewidth}{!}{%
    \begin{tabular}{cccccc}
    % \begin{tabular}{c|c|c|c|c|c}
        \toprule
        \textbf{Dataset} & \textbf{\# Lang} & \begin{tabular}{@{}c@{}}\textbf{\# Unseen}\\\textbf{BLOOM}\end{tabular} & \begin{tabular}{@{}c@{}}\textbf{\# Unseen}\\\textbf{XGLM}\end{tabular} & \begin{tabular}{@{}c@{}}\textbf{\# Lang}\\\textbf{Family}\end{tabular} & \textbf{Region(s)} \\
        \toprule
        NusaTranslation & 6 & 6 & 6 & 1 & Southeast Asia \\
        \midrule
        MasakhaNews & 9 & 4 & 8 & 3 & Africa \\
        \midrule
        AmericasNLI & 10 & 10 & 9 & 8 & South America \\
        \midrule
        \begin{tabular}{@{}c@{}}Tweet Sentiment\\ Multilingual\end{tabular} & 7 & 2 & 0 & 2 & \begin{tabular}{@{}c@{}}Northern Africa,\\Europe, Central Asia\end{tabular} \\
        \bottomrule
    \end{tabular}
    }
    \caption{The datasets and languages under study. Our study covers 25 low-resource languages and 7 relatively higher-resource languages from various regions.}
    \vspace{-6pt}
    \label{tab:languages}
\end{table}

\section{Experimental Settings}

% \subsection{Baselines}
% \label{sec:baselines}

% We compare our methods with 4 additional baselines: 1) \textit{zero-shot prompting} that performs inference without any ICL, 2)
% \textit{monolingual ICL} that performs inference using ICL from the same language as the query, 3) \textit{translate-test} that translates the query and performs inference in the high-resource language, and 4) \textit{translate-test + ICL} that simply combines \textit{translate-test} and \textit{monolingual ICL}. \textit{Zero-shot prompting} serves as a lower bound comparison against other methods, while \textit{monolingual ICL}, \textit{translate-test} and  \textit{translate-test + ICL} serve as comparisons against X-ICL. For all experiments that include translation, we utilize machine translation models from NLLB~\cite{nllb2022nllb}.\footnote{\url{https://huggingface.co/facebook/nllb-200-distilled-1.3B}}

\subsection{Retrieval and In-Context Learning Setup} To calculate the cross-lingual and monolingual semantic similarity, we utilize multilingual sentence transformers~\cite{reimers-2019-sentence-bert,reimers-gurevych-2020-making}.\footnote{As our semantic similarity model, we utilize \texttt{sentence-transformers/stsb-xlm-r-multilingual}} For all ICL experiments, we conduct ICL with 3-shot ICL exemplars. We run our experiments using two LLMs: XGLM-7.5B~\cite{lin2022fewshot} and BLOOM-7B~\cite{scao2022bloom}. To select the prediction label, we take the label that maximizes the marginal probability of the prompt:
\vspace{-4pt}
\begin{align}
    c^{pred} &= \argmax_c P(X^{icl}, X^{align}, q^{tgt},c) \\
     &= f(X^{icl} \oplus X^{align} \oplus q^{tgt} \oplus c)
\end{align}

where $f(.)$ denotes a language model, $\oplus$ denotes the concatenation operator, $X^{icl}$ denotes the ICL exemplars, $X^{align}$ denotes the alignment text, and $c$ denotes the class label taken from the label set.

\begin{table}[!t]
    \centering
    \resizebox{\linewidth}{!}{%
    \begin{tabular}{ccc}
    % \begin{tabular}{c|c|c}
        \toprule
        \textbf{Eval Dataset} & $\boldsymbol{D^{src}}$ & $\boldsymbol{D^{para}}$ \\
        \toprule
        NusaTranslation & \begin{tabular}{@{}c@{}}NusaX-Senti\\\cite{winata2022nusax}\end{tabular} & \begin{tabular}{@{}c@{}}NusaX-MT\\\cite{winata2022nusax}\end{tabular} \\
        \midrule
        MasakhaNews & \begin{tabular}{@{}c@{}}MasakhaNews\\(Eng Train set)\end{tabular} & \begin{tabular}{@{}c@{}}MAFAND\\\cite{adelani-etal-2022-thousand}\end{tabular} \\
        \midrule
        AmericasNLI & \begin{tabular}{@{}c@{}}XNLI (Eng)\\\cite{conneau-etal-2018-xnli}\end{tabular} & \begin{tabular}{@{}c@{}}XNLI (Eng) $\bigoplus$\\AmericasNLI (Dev set)$^\star$\end{tabular} \\
        \midrule
        \begin{tabular}{@{}c@{}}Tweet Sentiment\\ Multilingual\end{tabular} & \begin{tabular}{@{}c@{}}Tweet Sentiment\\ Multilingual (Eng Train set)\end{tabular} & \begin{tabular}{@{}c@{}}Tweet Sentiment\\ Multilingual (Eng MT)$^\dagger$\end{tabular} \\
        \bottomrule
    \end{tabular}
    }
    \caption{The $D^{src}$ and $D^{para}$ for all the evaluation datasets under study.$^\dagger$ Translated to English using NLLB~\cite{nllb2022nllb}.$^\star$ We align the two datasets.}
    \vspace{-6pt}
    \label{tab:dataset}
\end{table}

\subsection{Languages and Datasets}

As shown in Table~\ref{tab:languages}, our study includes 25 low-resource languages from three different regions, i.e., Africa, Americas, and South-East Asia, covering 13 language families. Note that, many of the low-resource languages are unseen to both XGLM and BLOOM, nonetheless, both models might have seen other languages under the same language family group with those low-resource languages, e.g., both models are pre-trained on Indonesian, which falls under the same language family group (i.e., Malayo-Polynesian) to the low-resource languages in Indonesia. We also include 7 relatively higher-resource languages, i.e., Arabic (arb), French (fra), German (deu), Hindi (hin), Italian (ita), Portuguese (por), and Spanish (spa) for comparing the behavior of X-ICL between these relatively higher-resource languages and low-resource languages. Detailed information on all the languages under study is shown in Appendix~\ref{app:languages}.

\begin{figure*}[!t]
    \centering
    \begin{minipage}{.33\linewidth}
        \centering
        \begingroup
        \includegraphics[trim=0 10em 12em 0, width=\linewidth, clip]{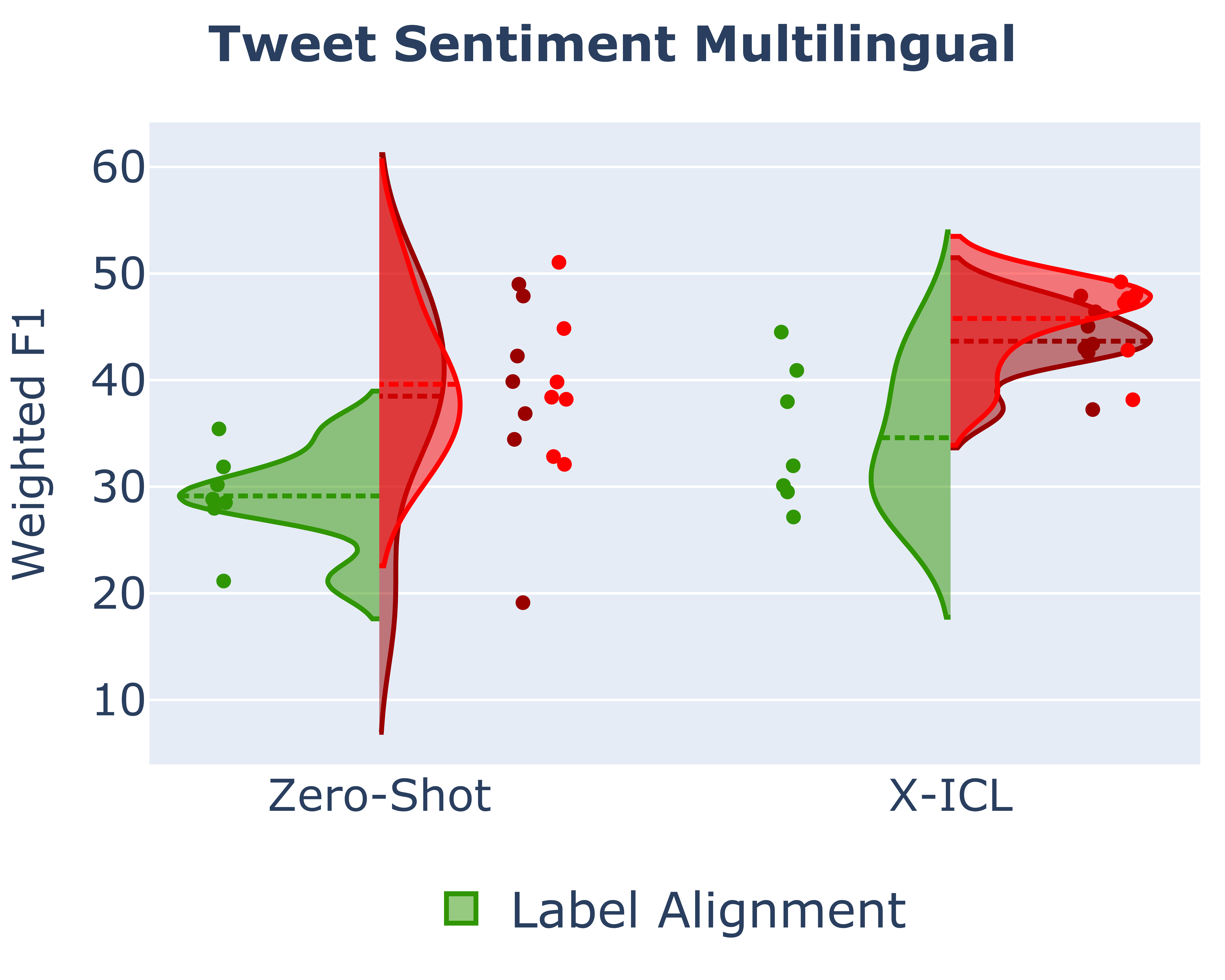}
        \endgroup
    \end{minipage}%
    \hspace{2pt}
    % \hfill\vline\hfill
    \begin{minipage}{.295\linewidth}
        \centering
        \begingroup
        \includegraphics[trim=0 10em 12em 0, width=\linewidth, clip]{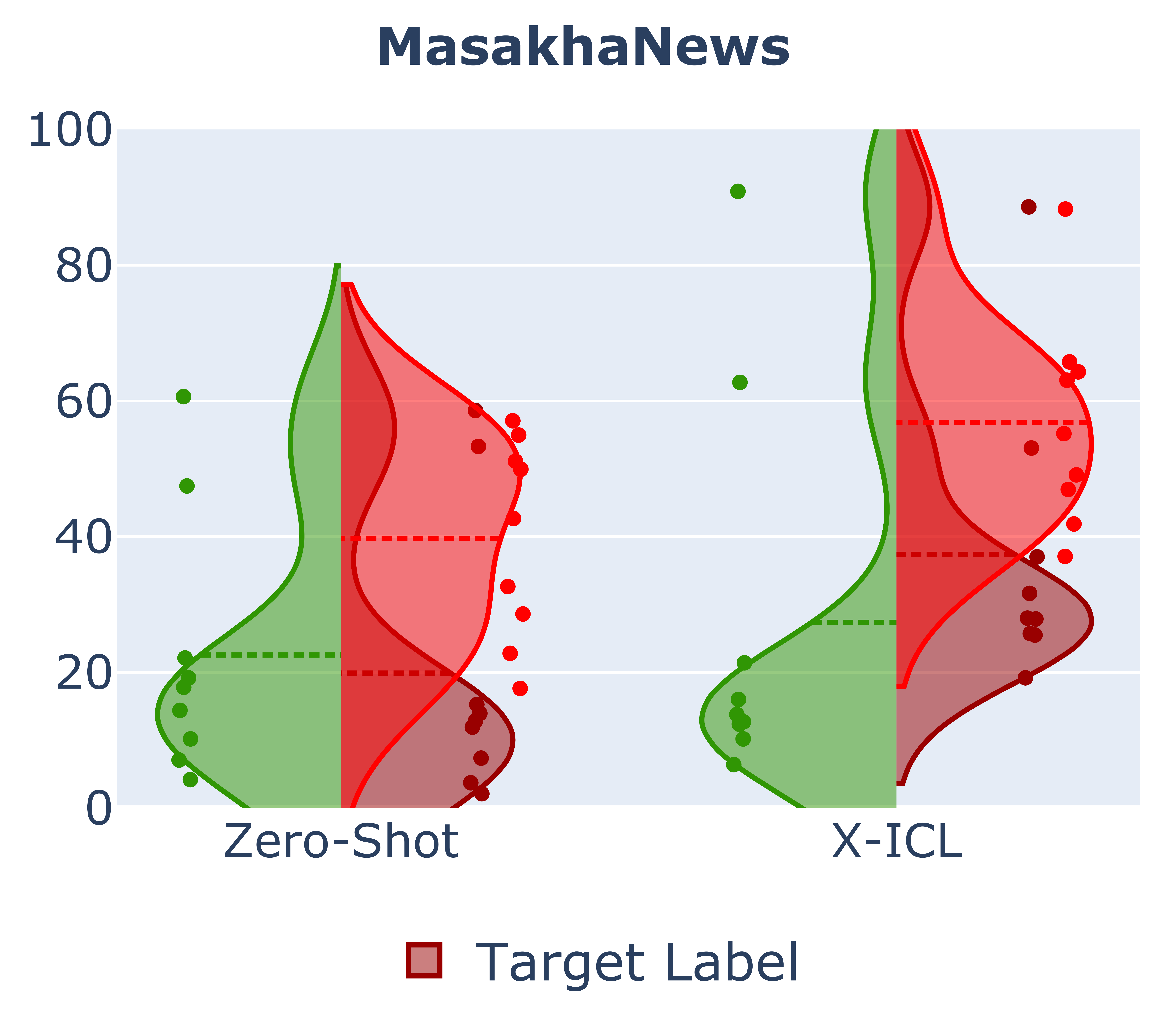}
        \endgroup
    \end{minipage}%
    \hspace{2pt}
    % \hfill\vline\hfill
    \begin{minipage}{.295\linewidth}
        \centering
        \begingroup
        \includegraphics[trim=0 10em 12em 0, width=\linewidth, clip]{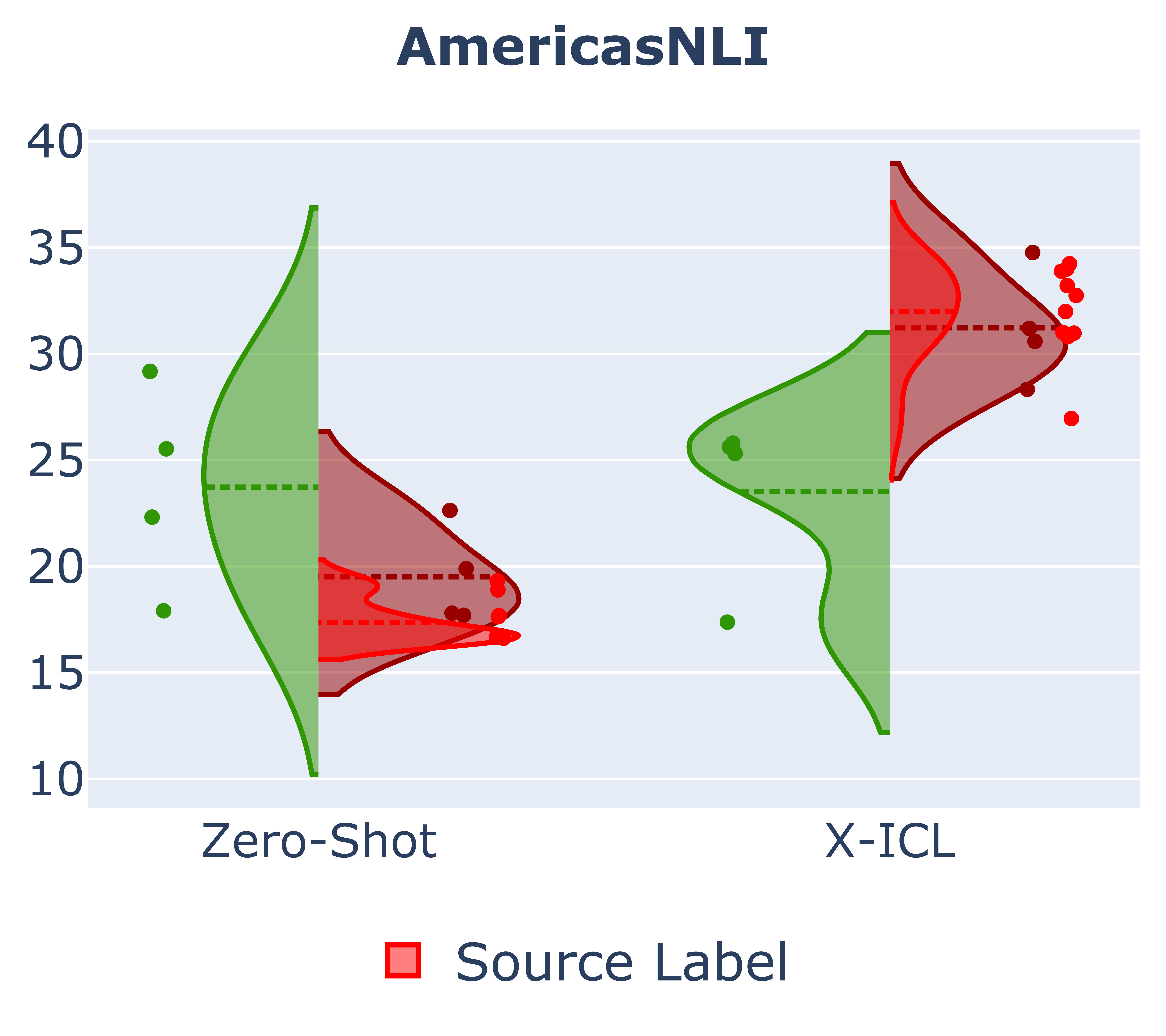}
        \endgroup
    \end{minipage}
    \vspace{-4pt}
    \caption{Performance of XGLM-7.5B with in-context label alignment, target-only label, and source-only label on \textbf{(left)} higher-resource, \textbf{(center)} low-resource African, and \textbf{(right)} low-resource American languages.}
    \vspace{-6pt}
    \label{fig:result_label}
\end{figure*}

All the languages are spread across four different datasets, i.e., MasakhaNews (\textbf{topic classification})~\cite{adelani2023masakhanews}, AmericasNLI (\textbf{natural language inference})~\cite{ebrahimi-etal-2022-americasnli}, NusaTranslation (\textbf{sentiment analysis})~\cite{cahyawijaya2023nusawrites}, and TweetSentimentMultilingual (\textbf{sentiment analysis})~\cite{barbieri-etal-2022-xlmt}. For each dataset, we defined the ICL dataset $D^{src}$ and parallel alignment dataset $D^{para}$ from different dataset subsets or completely different datasets. The details are shown in Table~\ref{tab:dataset}. Note that, we only take languages that are supported in NLLB~\cite{nllb2022nllb}, such that we can compare the performance with machine-translation-based approaches. For the monolingual semantic similarity baselines, we utilize the train and dev sets of the evaluation dataset.

\section{Result and Discussion}
% We break down the results and analysis of our experiments into three sections. 
The per-dataset results of our experiments are shown in Appendix~\ref{app:all_results}. For brevity, we report the analysis mainly for XGLM-7.5B, since we observe that the BLOOM-7B results are similar. We show the results for BLOOM in Appendix~\ref{app:bloom_analysis}.

\subsection{Inferiority of In-Context Label Alignment}
\label{sec:alignment}

\begin{figure}[!t]
    \centering
    \begin{minipage}{\linewidth}
        \centering
        \begingroup
        \includegraphics[trim=0 48em 0 0, width=\linewidth, clip]{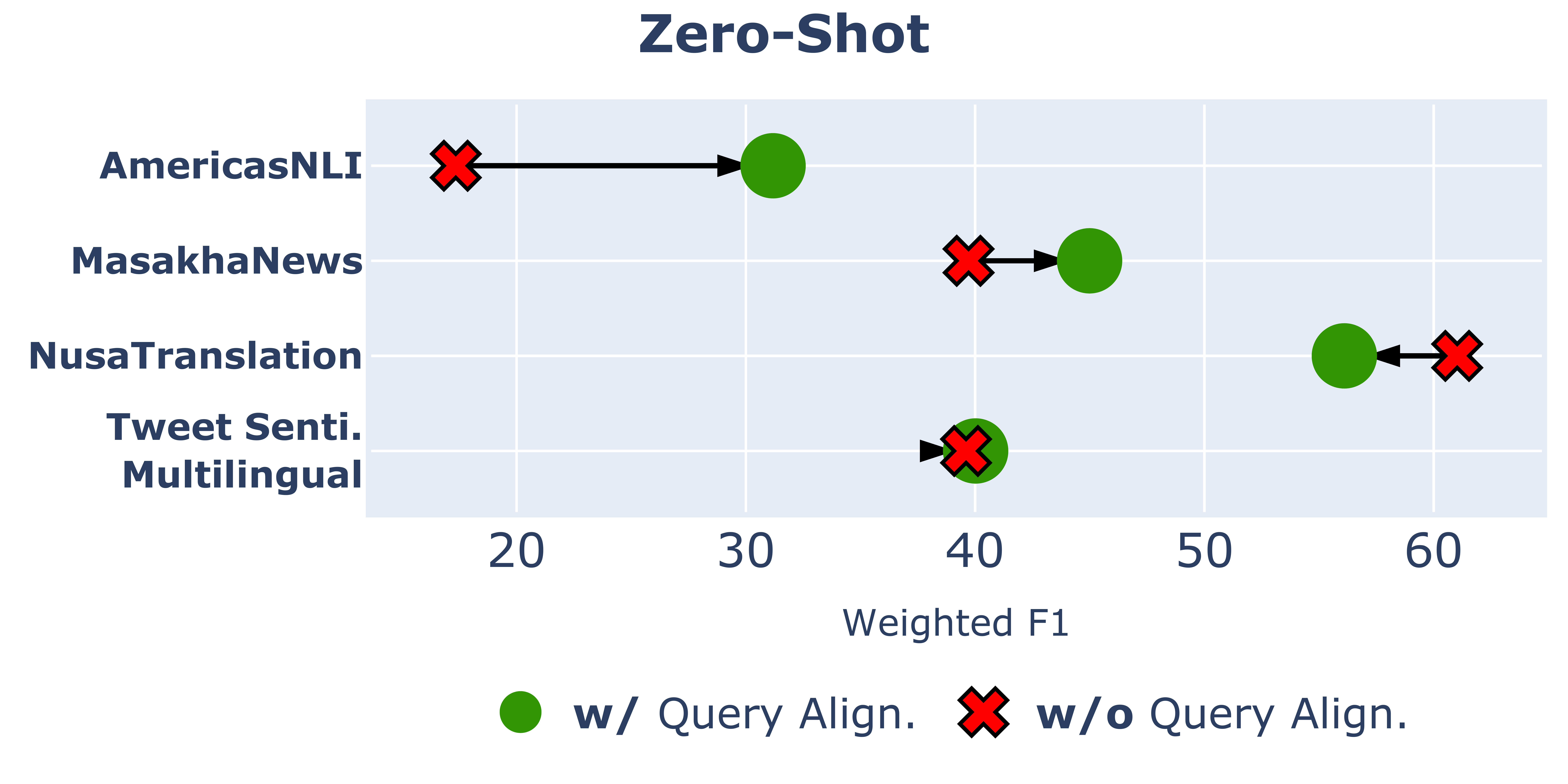}
        \endgroup
    \end{minipage}%
    \hspace{2pt}
    % \hfill\vline\hfill
    \begin{minipage}{\linewidth}
        \centering
        \begingroup
        \includegraphics[trim=0 12em 0 0, width=\linewidth, clip]{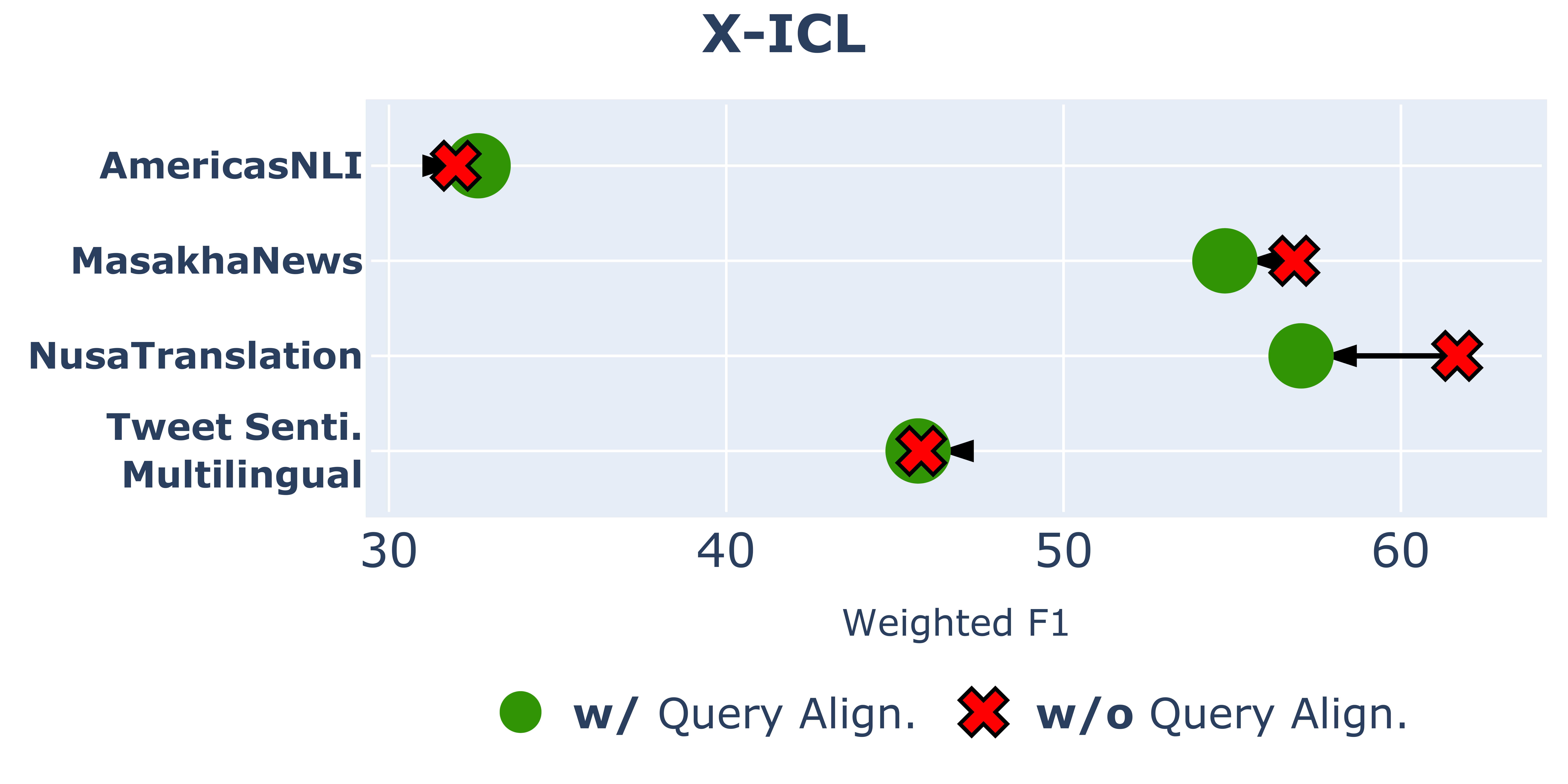}
        \endgroup
    \end{minipage}
    \caption{Performance of XGLM-7.5B with and without query alignment on \textbf{(top)} zero-shot and \textbf{(bottom)} X-ICL settings.}
    \label{fig:result_input}
    \vspace{-6pt}
\end{figure}

% \paragraph{In-Context Label Alignment vs Uniform Label}
% Boxen chart IOA True, False, Target for ZS and X-ICL on Higher resource languages & low-resource languages (per regions)

Figure~\ref{fig:result_label} shows the comparison of \textbf{in-context label alignment} with uniform \textbf{source-only} and \textbf{target-only} labels.
In most languages, in-context label alignment yields lower performance than target-only label, and source-only label yields the best performance. For low-resource African languages,
% , while the performance of \textit{source-only label} remains high, 
the target-only label performs much worse. We conjecture that this is due to the weak representation of these languages, which is less apparent in low-resource Indonesian and American languages because the target labels (see Appendix~\ref{app:language-label}) are similar to higher-resource languages in training.
% , i.e., Indonesian and Spanish, respectively. 
Contrary to \citet{tanwar2023multilingual}, our results highlight the ineffectiveness of \textbf{in-context label alignment} to improve X-ICL on both higher-resource and low-resource languages.

\begin{figure}[!t]
    \centering
    \begin{minipage}{.515\linewidth}
        \centering
        \begingroup
        \includegraphics[trim=0 0 5em 0, width=\linewidth, clip]{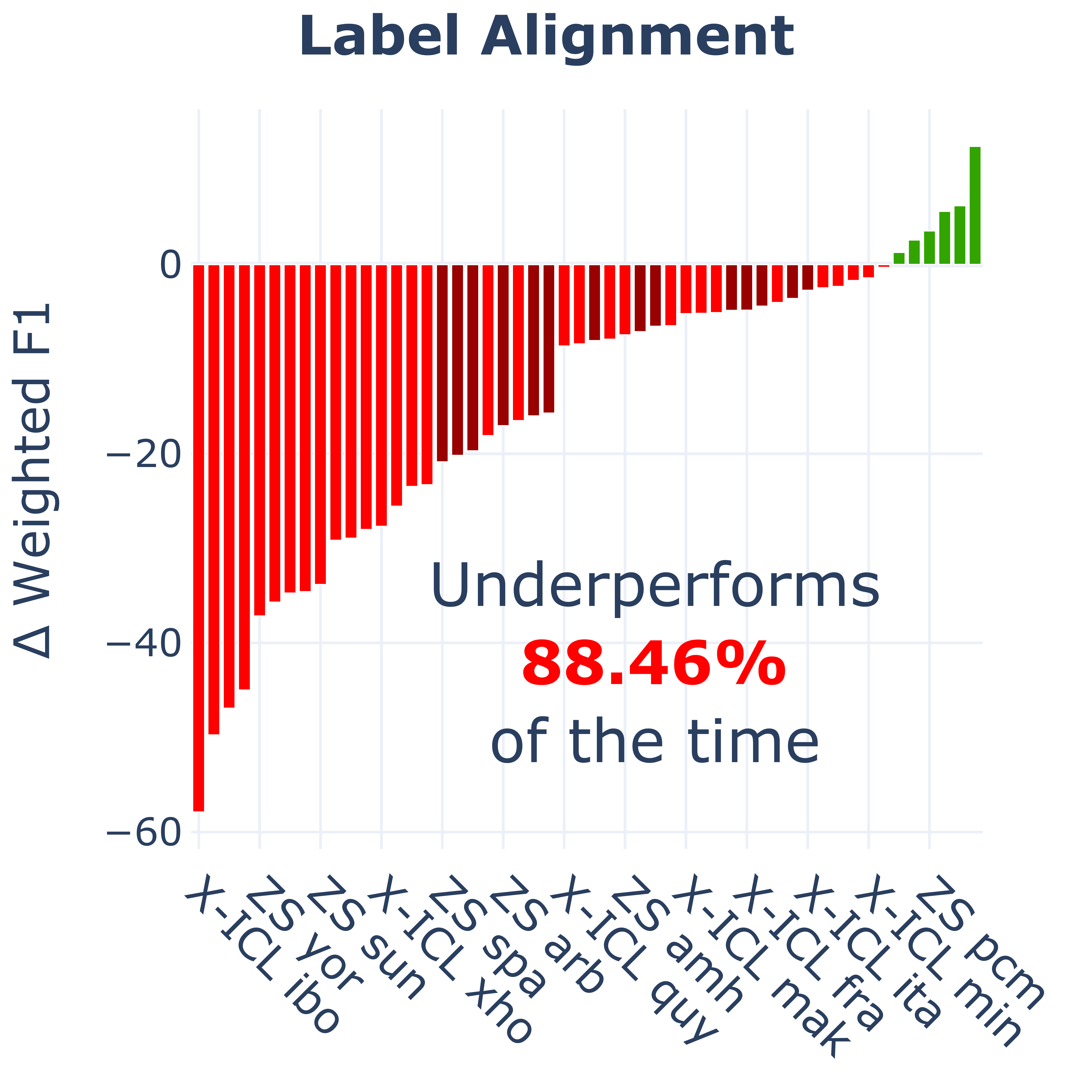}
        \endgroup
    \end{minipage}
    \begin{minipage}{.465\linewidth}
        \centering
        \begingroup
        \includegraphics[trim=27em 0 5em 0, width=\linewidth, clip]{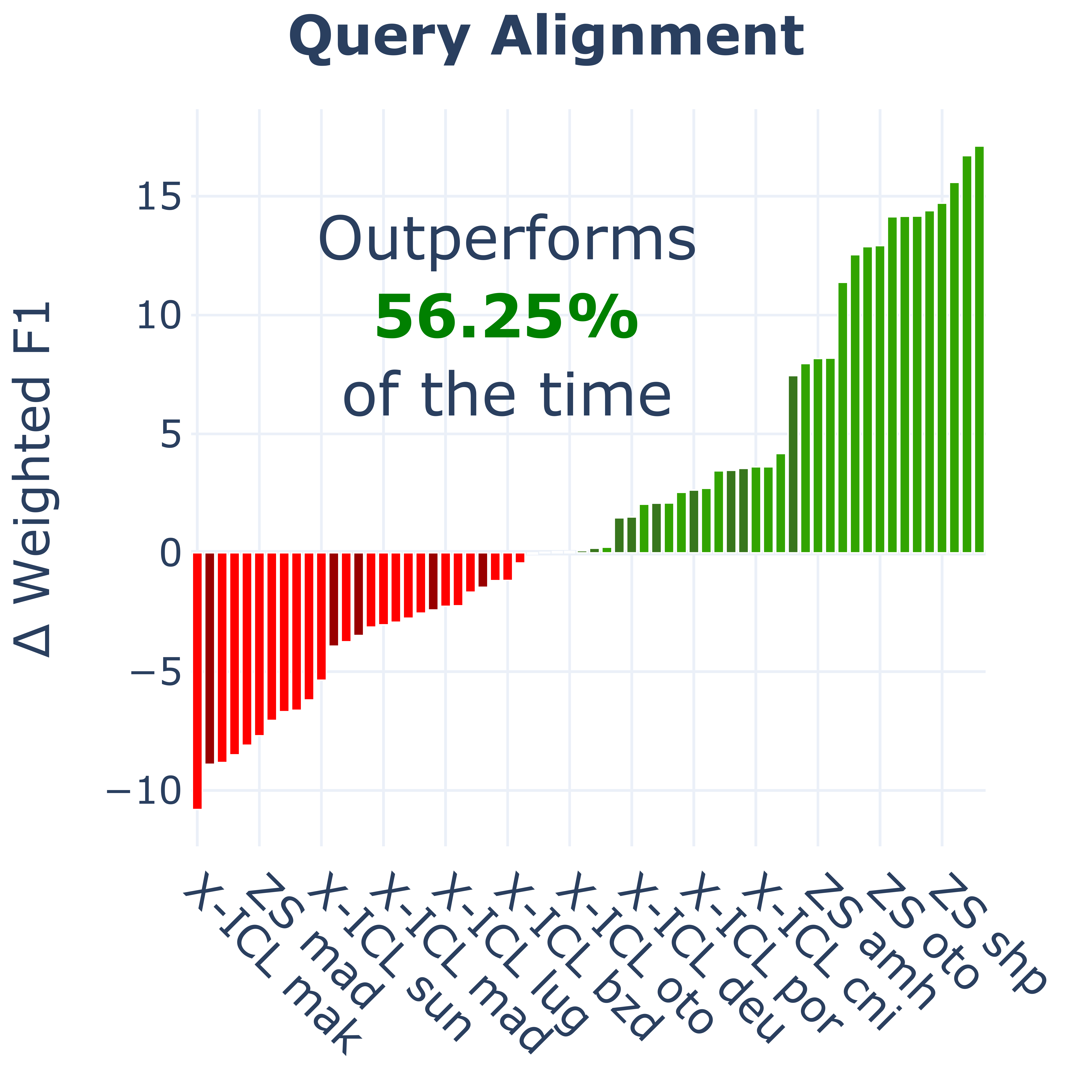}
        \endgroup
    \end{minipage}
    \caption{$\Delta$Weighted F1 of \textbf{(left)} in-context label alignment and \textbf{(right)} in-context query alignment against non-alignment baseline. A score < 0 indicates the in-context alignment degrades the performance.}
    \vspace{-8pt}
    \label{fig:label_vs_input}
\end{figure}

\begin{figure*}[!t]
    \centering
    \begin{minipage}{.205\linewidth}
        \centering
        \begingroup
        \includegraphics[trim=0 2em 0 0, width=\linewidth]{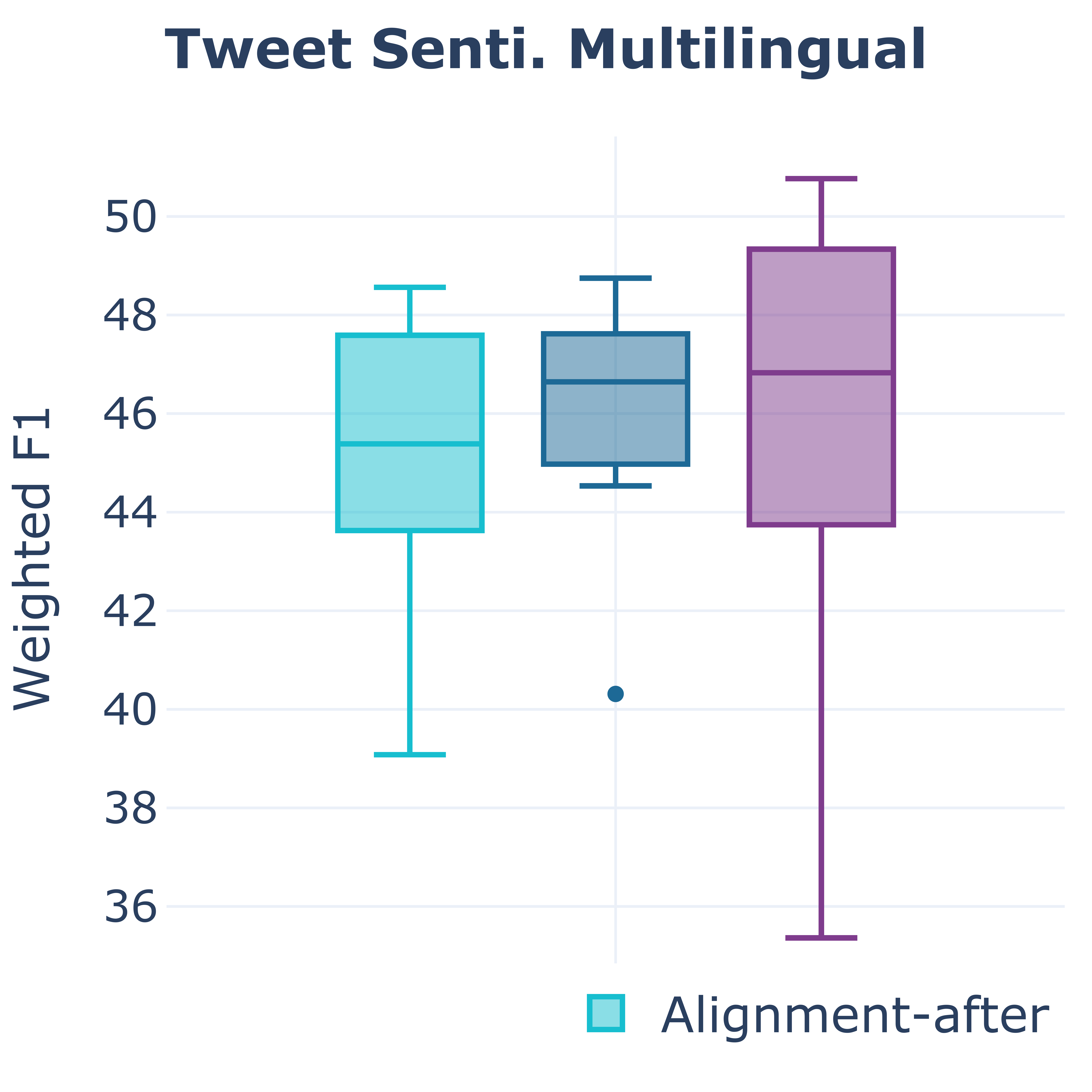}
        \endgroup
    \end{minipage}%
    % \hspace{2pt}
    % \hfill\vline\hfill
    \begin{minipage}{.205\linewidth}
        \centering
        \begingroup
        \includegraphics[trim=0 2em 0 0, width=\linewidth]{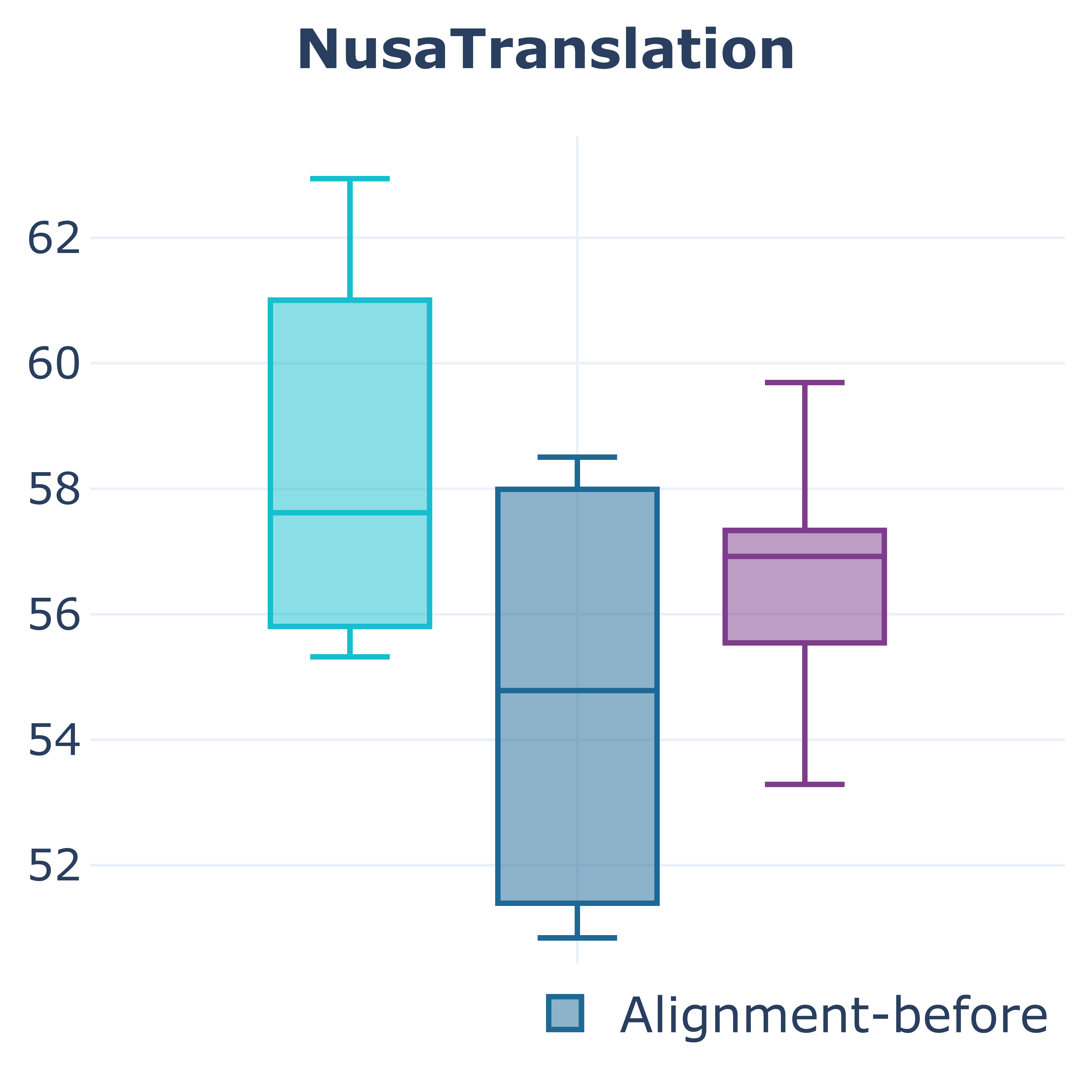}
        \endgroup
    \end{minipage}%
    % \hspace{2pt}
    % \hfill\vline\hfill
    \begin{minipage}{.205\linewidth}
        \centering
        \begingroup
        \includegraphics[trim=0 2em 0 0, width=\linewidth]{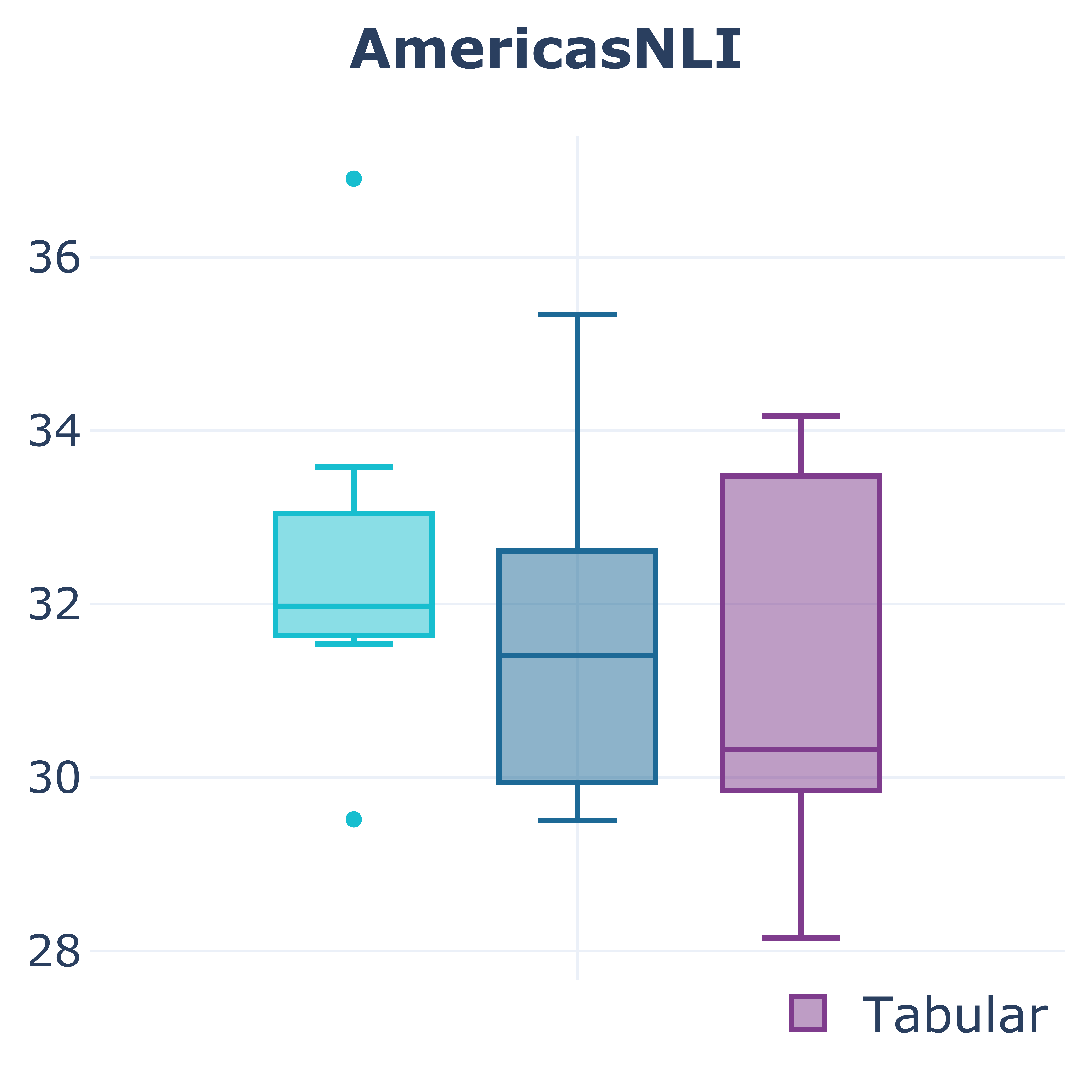}
        \endgroup
    \end{minipage}
    % \hspace{2pt}
    % \hfill\vline\hfill
    \begin{minipage}{.205\linewidth}
        \centering
        \begingroup
        \includegraphics[trim=0 2em 0 0, width=\linewidth]{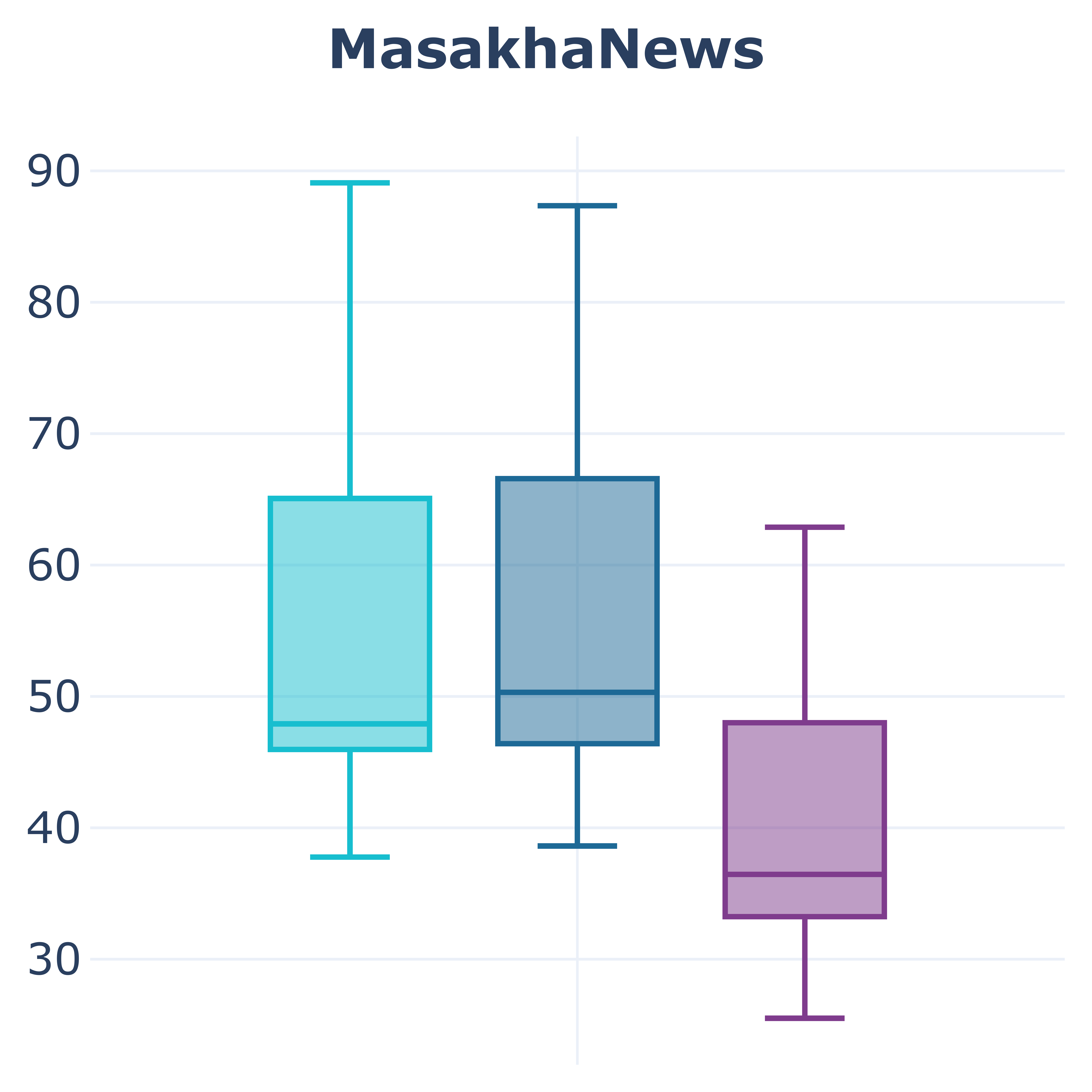}
        \endgroup
    \end{minipage}
    \caption{Performance of XGLM-7.5B with different alignment formats ordered by the degree of formatting consistency on \textbf{(1)} higher-resource languages, \textbf{(2)} low-resource Indonesian languages, \textbf{(3)} low-resource American languages, and \textbf{(4)} low-resource African languages.}
    \vspace{-9pt}
    \label{fig:result_format}
\end{figure*}

\subsection{In-Context Query Alignment}
% Boxen chart IIA True/False for ZS and X-ICL on Higher resource languages & low-resource languages (per regions)

We introduce in-context query alignment as an alternative to in-context label alignment in \S\ref{sec:cross-lingual-alignment}. As shown in Figure~\ref{fig:result_input}, \textbf{in-context query alignment} yields similar performance with the baseline (i.e., without query alignment) on higher-resource languages while improving zero-shot performance on low-resource languages. Nonetheless, the improvement is rather marginal in the X-ICL setting on low-resource languages. In this case, we conclude that in-context query alignment can be used as an alternative to X-ICL, which is favorable when there is no available X-ICL corpus for the particular task. With the recent development of large multilingual parallel corpora, such as Bloom Library~\cite{leong-etal-2022-bloom}, WikiMatrix~\cite{schwenk-etal-2021-wikimatrix}, CC-Aligned~\cite{chaudhary-etal-2019-low,el-kishky-etal-2020-ccaligned}, FLORES-200~\cite{nllb2022nllb}, and  GATITOS~\cite{jones2023bilex}, in-context query alignment can also be a perfect complement to X-ICL for improving LLMs understanding on thousands of languages.

\paragraph{Label Alignment vs Query Alignment} To investigate how well in-context alignments can affect the understanding of all the languages under study,
% within different inference settings (zero-shot / X-ICL)
we analyze their effectiveness by comparing them with the corresponding non-alignment baseline. As shown in Figure~\ref{fig:label_vs_input}, in-context label alignment only improves the performance at $\sim$11.54\% of the time with an improvement of $\sim$5\% weighted F1, while the rest 88.46\% experiments are decreased by $\sim$20\% weighted F1. In-context query alignment, on the other hand, increases the performance 56.25\% of the time with an improvement of $\sim$10\% weighted F1, while the rest 43.75\% of the time experiences a reduction of $\sim$5\% weighted F1. 
Our results suggest that \textbf{in-context query alignment} is superior to \textbf{in-context label alignment},
% and it can substitute or complement X-ICL, particularly for low-resource languages.
and it improves LLMs' understanding of low-resource languages in the absence of X-ICL task-specific data, which leads to performance gain.

\subsection{Why Query Alignment Performs Better}

In regards to the in-context alignment, we can first simplify the effect of alignment into the following two possibilities: 1) when the alignment is successful (upper bound) and when no alignment is done (lower bound). When The LLM successfully aligns the query in the source language to the target language, query alignment will enable the LLM to understand the query in the target as well as in the source languages, the LLM will reach a performance similar to monolingual ICL, which is the upper bound performance. While in label alignment,  when the LLM successfully aligns the label in the source to the target languages, the LLM understands the label semantics, but there is no guidance on how to interpret the query in the target language. In this case, the upper bound is equivalent to performing X-ICL which generally performs slightly worse than monolingual high-resource language ICL which is reflected in our result in \S\ref{sec:x-icl}.

When the LLM completely fails to align the label in the source to the target languages, in the query alignment, the LLM performs a regular X-ICL, which is similar to the best case of the label alignment. While in the label alignment, the LLM performs X-ICL with a shifted label space. The harmful effect of ICL with a shifted label space has been extensively studied in~\cite{min2022rethinking}, which results in severe performance degradation. With regards to the two possibilities, the upper-bound and lower-bound of in-context query alignment are better than label alignment, thus query alignment outperforms label alignment on average. In a more realistic scenario, there is also another factor where the alignment text becomes the noise that will shift the output prediction of the LLMs. As the noise factor happens for both query and label alignment, we can assume the same effect of noise for both methods and omit this factor into account, resulting in the same conclusion.

% \begin{figure*}[!t]
%     \centering
%     \begin{minipage}{.29\linewidth}
%         \centering
%         \begingroup
%         \includegraphics[trim=0 2em 0 0, width=\linewidth]{images/format_consistency/tweetsentimulti_xglm-7.5B.pdf}
%         \endgroup
%     \end{minipage}%
%     \hspace{4pt}
%     \begin{minipage}{.29\linewidth}
%         \centering
%         \begingroup
%         \includegraphics[trim=0 2em 0 0, width=\linewidth]{images/format_consistency/masakhanews_xglm-7.5B.pdf}
%         \endgroup
%     \end{minipage}%
%     \hspace{4pt}
%     \begin{minipage}{.29\linewidth}
%         \centering
%         \begingroup
%         \includegraphics[trim=0 2em 0 0, width=\linewidth]{images/format_consistency/americasnli_xglm-7.5B.pdf}
%         \endgroup
%     \end{minipage}%
%     \caption{Performance of XGLM-7.5B with different alignment formats ordered by the degree of formatting consistency on \textbf{(left)} higher-resource languages, \textbf{(center)} low-resource African languages, and \textbf{(right)} low-resource American languages.}
%     \vspace{-4pt}
%     \label{fig:result_format}
% \end{figure*}

\subsection{Effect of Format Consistency}
% IOA/IIA/IOA+IIA Alignment Before, IOA/IIA/IOA+IIA Alignment After, ITC on X-ICL on Higher resource languages & low-resource languages (per regions)

\begin{figure*}[!t]
    \centering
    \begin{minipage}{.23\linewidth}
        \centering
        \begingroup
        \includegraphics[trim=0 2em 0 0, width=\linewidth]{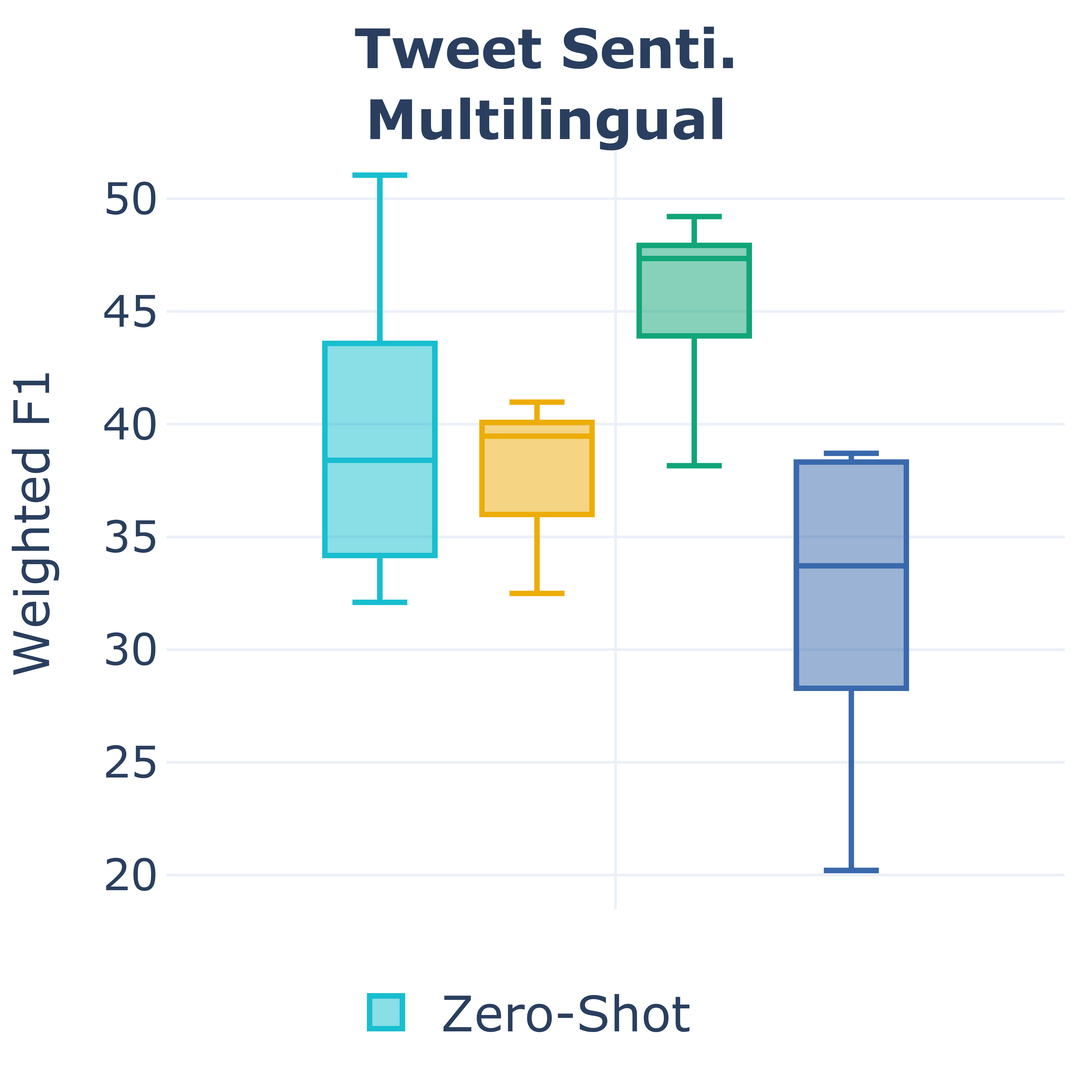}
        \endgroup
    \end{minipage}%
    % \hspace{2pt}
    % \hfill\vline\hfill
    \begin{minipage}{.23\linewidth}
        \centering
        \begingroup
        \includegraphics[trim=0 2em 0 0, width=\linewidth]{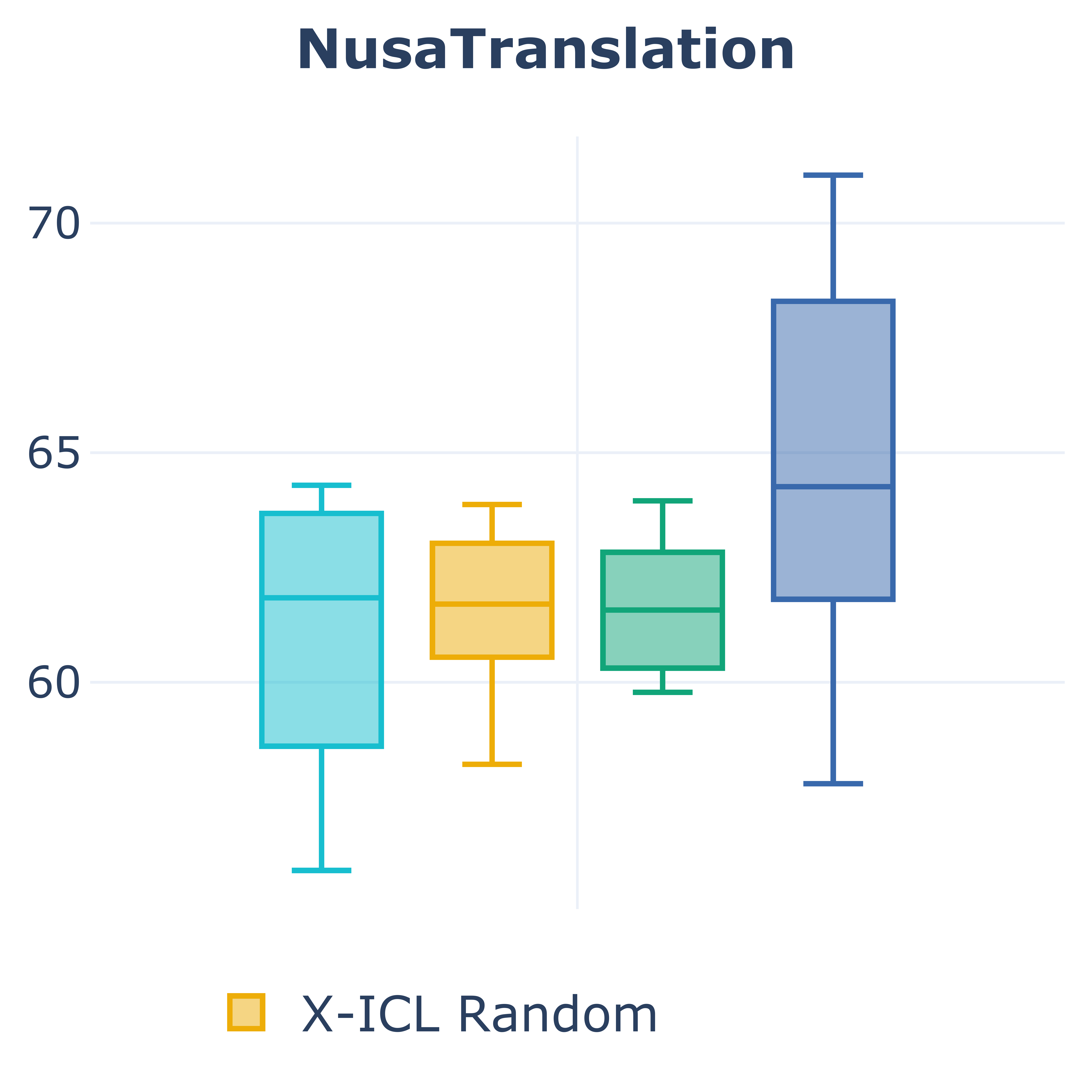}
        \endgroup
    \end{minipage}%
    % \hspace{2pt}
    % \hfill\vline\hfill
    \begin{minipage}{.23\linewidth}
        \centering
        \begingroup
        \includegraphics[trim=0 2em 0 0, width=\linewidth]{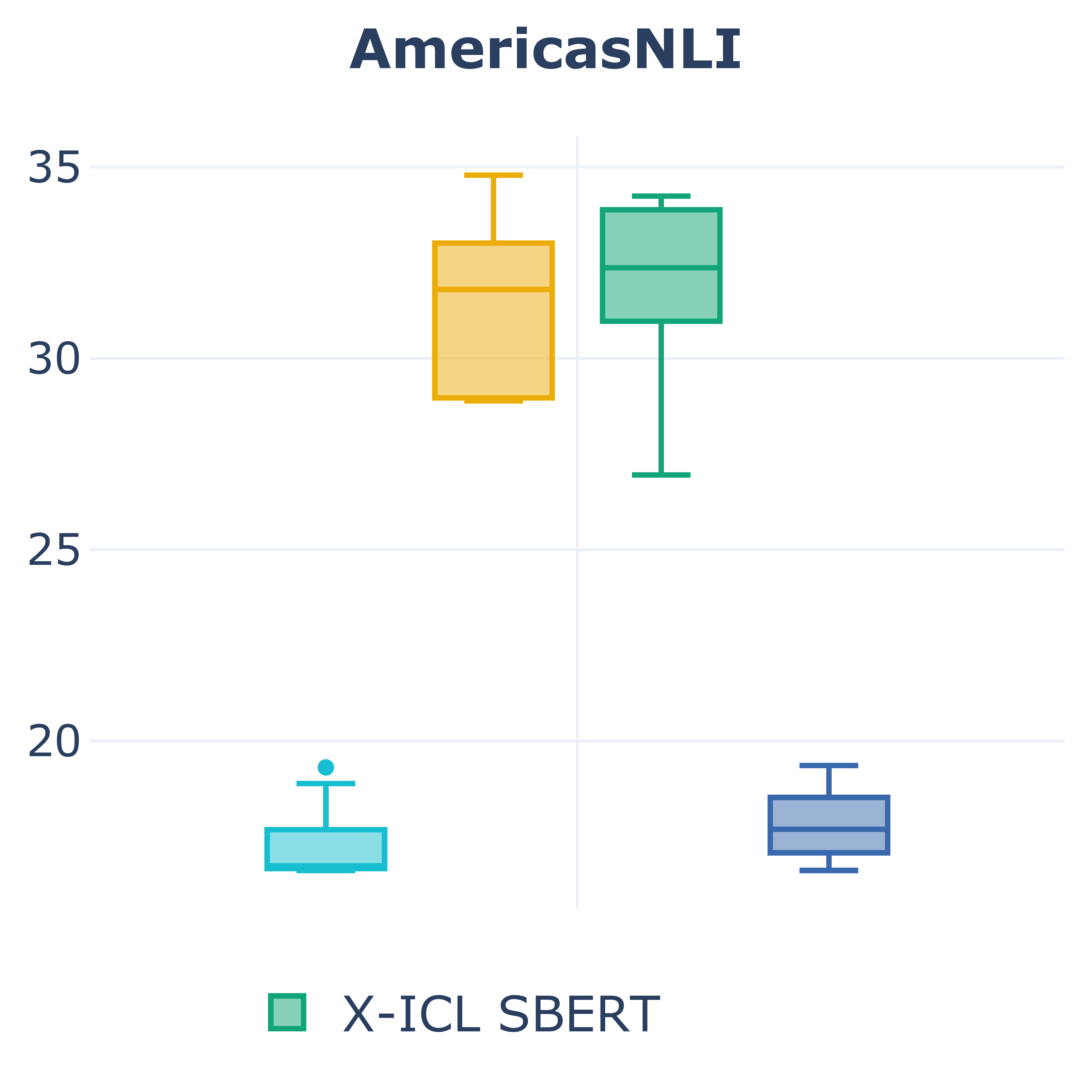}
        \endgroup
    \end{minipage}
    % \hspace{2pt}
    % \hfill\vline\hfill
    \begin{minipage}{.225\linewidth}
        \centering
        \begingroup
        \includegraphics[trim=0 2em 0 0, width=\linewidth]{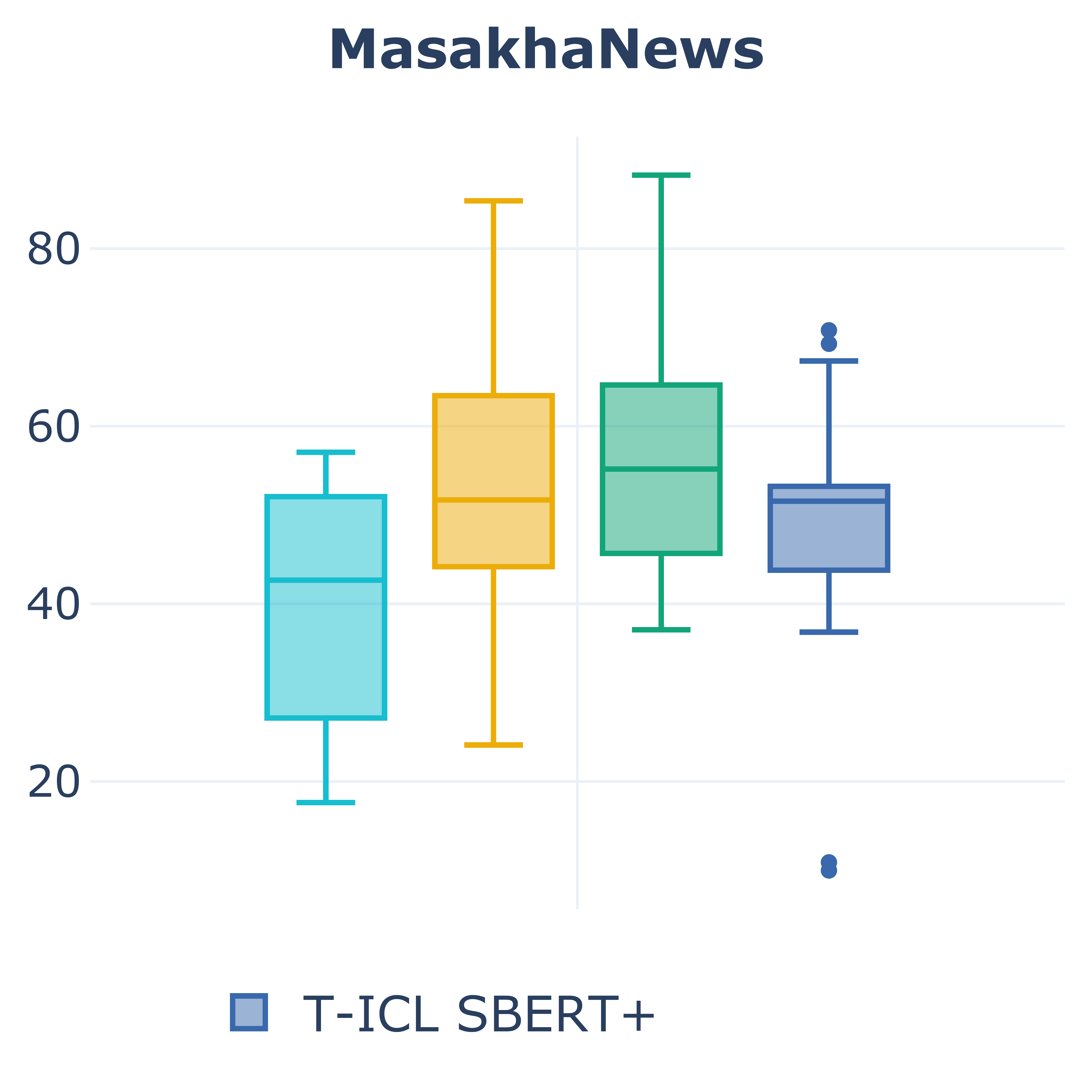}
        \endgroup
    \end{minipage}
    \caption{Performance of XGLM-7.5B with different in-context learning retrievals covering monolingual, cross-lingual, translation semantic similarity (T-ICL) on \textbf{(1)} higher-resource languages, \textbf{(2)} low-resource Indonesian languages, \textbf{(3)} low-resource American languages, and \textbf{(4)} low-resource African languages. Random and SBERT denotes random and semantic-similarity-based exemplar selection, respectively.}
    \vspace{-9pt}
    \label{fig:result_xss}
\end{figure*}

We explore three types of prompting with various degrees of formatting consistency (\S\ref{sec:cross-lingual-prompting}). As shown in Figure~\ref{fig:result_format}, for higher-resource languages, formatting consistency correlates to a slight improvement in the downstream performance for both XGLM and BLOOM (see Appendix~\ref{app:bloom_analysis}) models. Meanwhile, for low-resource languages, the trend for both models is unclear. We conjecture that increasing the format consistency can improve the downstream task performance on well-represented languages. For low-resource languages, increasing the format consistency will not improve the model understanding. Increasing the representation through X-ICL and query alignment would be a better alternative to improve the low-resource language understanding ability of LLMs.

\subsection{Importance of Cross-Lingual Retrieval}

\paragraph{Cross-Lingual Semantic Similarity}
% Different X-ICL & XPresso w/ IOA/IIA/IOA+IIA on Higher resource languages & low-resource languages (per regions)

We compare the effectiveness of cross-lingual semantic similarity to monolingual and translation semantic similarity for retrieving ICL and X-ICL exemplars. Based on Figure~\ref{fig:result_xss}, X-ICL and ICL with cross-lingual and monolingual semantic-similarity-based retrieval, respectively, perform better than zero-shot prompting, suggesting the effectiveness of these approaches for improving the task understanding of LLMs. In addition, we show that translation semantic similarity performs almost on par with the zero-shot baseline. We hypothesize that this problem is attributed to the error propagation of the pipelined nature of the translation semantic similarity system and the limited coverage of parallel exemplars in $D^{para}$, showing the benefit of using direct cross-lingual semantic similarity retrieval over translation-based retrieval. Furthermore, the performance of cross-lingual semantic similarity is similar to or slightly lower than the monolingual semantic similarity approach. Hence, cross-lingual semantic similarity retrieval is important in the case where the corpus for performing monolingual ICL on a particular task is not available.

\begin{figure*}[!t]
    \centering
    \includegraphics[trim=0 0.7em 0 0, width=0.46\linewidth, clip]{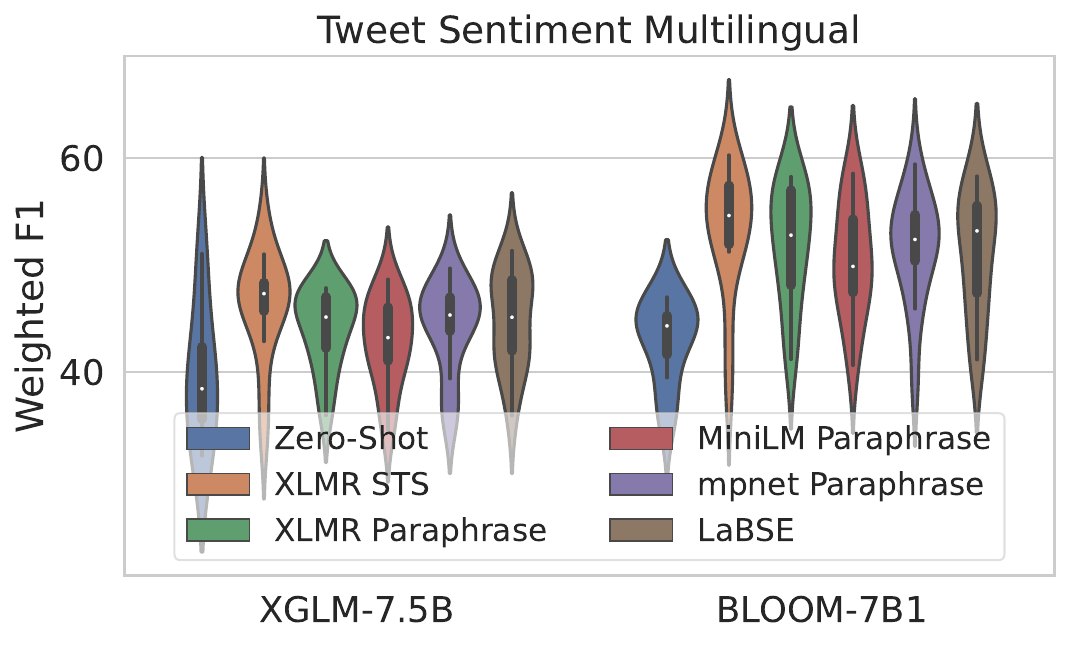}
    \includegraphics[trim=2.5em 0.7em 0 0, width=0.45\linewidth, clip]{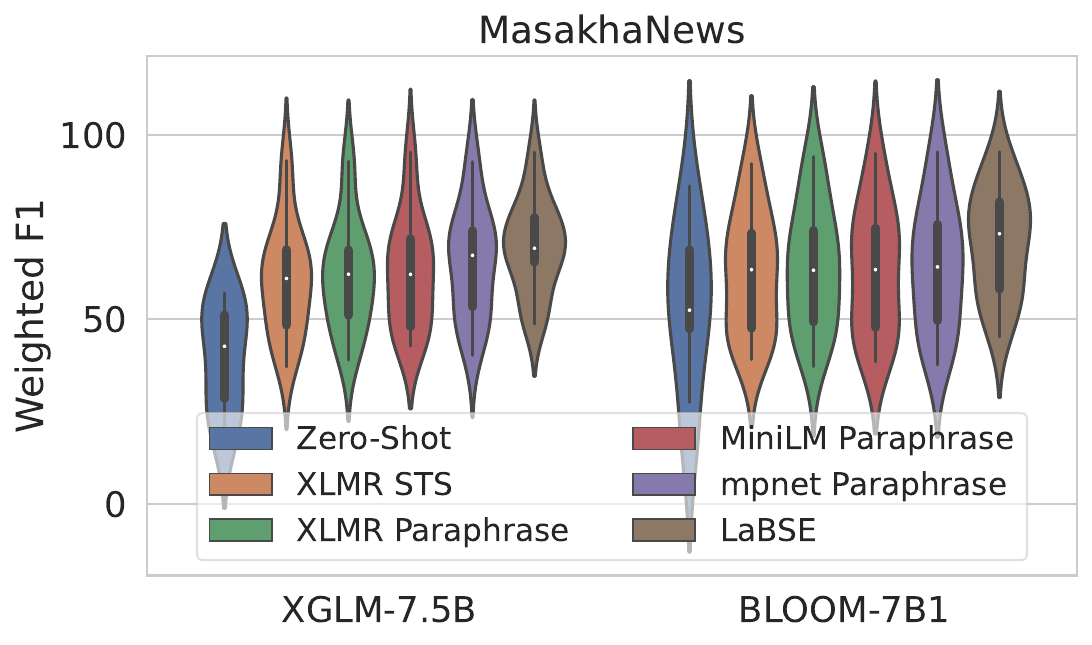}

    % \begin{minipage}{.33\linewidth}
    %     \centering
    %     \begingroup
    %     \includegraphics[width=\linewidth]{images/sbert/tweetsentimulti.pdf}
    %     \endgroup
    % \end{minipage}%
    % % \hspace{2pt}
    % % \hfill\vline\hfill
    % \begin{minipage}{.33\linewidth}
    %     \centering
    %     \begingroup
    %     \includegraphics[width=\linewidth]{images/sbert/masakhanews.pdf}
    %     \endgroup
    % \end{minipage}%
    % % \hspace{2pt}
    % % \hfill\vline\hfill
    % \begin{minipage}{.33\linewidth}
    %     \centering
    %     \begingroup
    %     \includegraphics[width=\linewidth]{images/sbert/americasnli.pdf}
    %     \endgroup
    % \end{minipage}
    \caption{Performance of LLMs with different semantic similarity models on \textbf{(left)} higher-resource languages and \textbf{(right)} low-resource African languages.}
    \vspace{-10pt}
    \label{fig:result_ssm}
\end{figure*}

\paragraph{Variations of Semantic Similarity Models}
% Different Sentence BERT models on Higher resource languages & low-resource languages (per regions)

We further compare the effectiveness of varying the cross-lingual semantic similarity models for cross-lingual retrieval. As shown in Figure~\ref{fig:result_ssm}, all cross-lingual semantic similarity models outperform the zero-shot baseline. Interestingly, despite the reported inferiority of STS-tuned models over paraphrasing-tuned models and LaBSE in prior works~\cite{reimers-2019-sentence-bert,reimers-gurevych-2020-making,feng2022labse}, our results showcase otherwise. On average, XLMR STS performs on par with other models, gaining a better performance on high-resource languages while getting a worse performance on low-resource languages. We find that, depending on the language under study, the choice of cross-lingual semantic similarity models can play a huge role in the downstream performance of X-ICL.

\begin{figure*}[!t]
    \centering
    \includegraphics[trim=0 0 0 0,width=0.48\linewidth,clip]{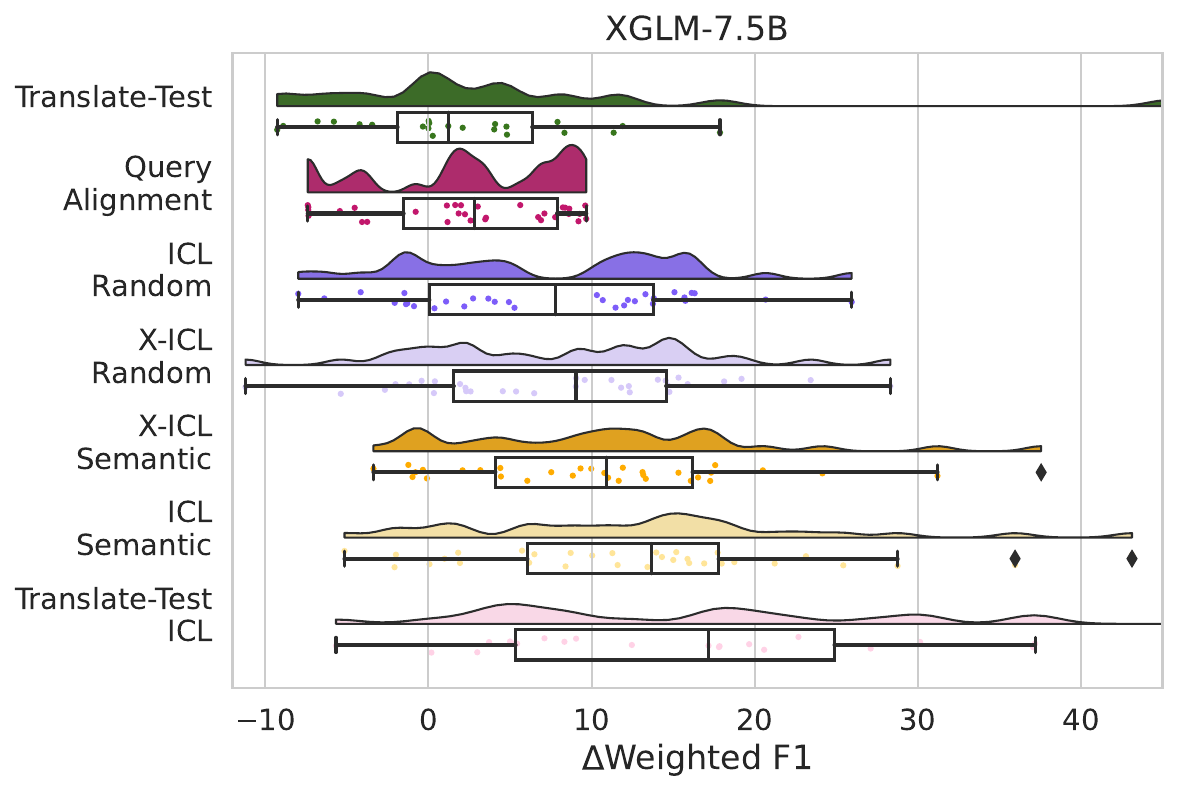}
    \hspace{4pt}
    \includegraphics[trim=0 0 0 0,width=0.48\linewidth,clip]{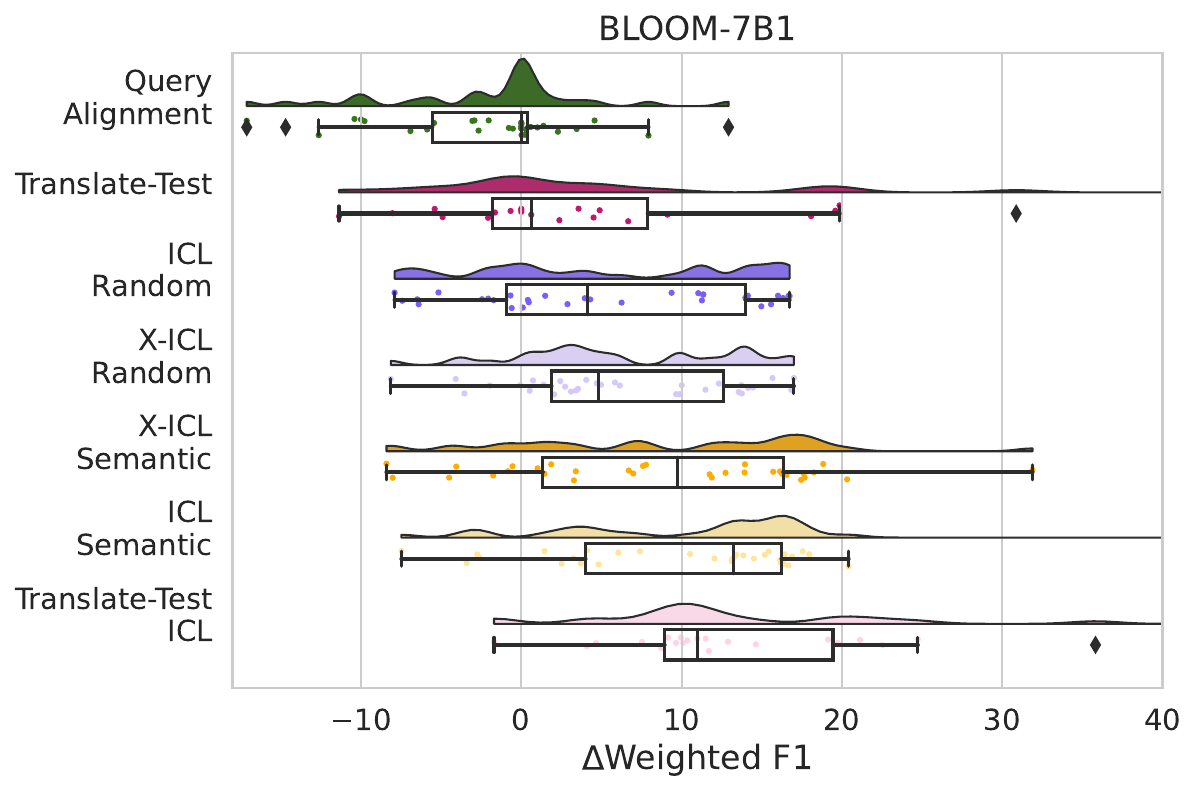}
    % \begin{minipage}{.33\linewidth}
    %     \centering
    %     \begingroup
    %     \includegraphics[width=\linewidth]{images/xicl_ok/tweetsentimulti.pdf}
    %     \endgroup
    % \end{minipage}%
    % % \hspace{2pt}
    % % \hfill\vline\hfill
    % \begin{minipage}{.33\linewidth}
    %     \centering
    %     \begingroup
    %     \includegraphics[width=\linewidth]{images/xicl_ok/masakhanews.pdf}
    %     \endgroup
    % \end{minipage}%
    % % \hspace{2pt}
    % % \hfill\vline\hfill
    % \begin{minipage}{.33\linewidth}
    %     \centering
    %     \begingroup
    %     \includegraphics[width=\linewidth]{images/xicl_ok/nusatranslation.pdf}
    %     \endgroup
    % \end{minipage}
    % \vspace{-6pt}adilazuarda2024lingualchemy
    \caption{Gain/Loss of various test-time adaptation methods for low-resource languages using \textbf{(top)} XGLM-7.5B and \textbf{(bottom)} BLOOM-7B1 backbones.}
    \vspace{-10pt}
    \label{fig:result_ok_xicl}
\end{figure*}

% \section{Discussion}

\subsection{Is X-ICL Effective for low-resource Languages?}
% Boxen chart (ZS, ZS-T, M-ICL, X-ICL, T-ICL) on Higher resource languages & low-resource languages (per regions)
\label{sec:x-icl}

To analyze the effectiveness of X-ICL in low-resource languages, we compare X-ICL with other inference approaches. Specifically, we compare X-ICL with 3 other baselines: 1) monolingual \textbf{ICL} that performs inference using ICL from the same language as the query, 2) \textbf{translate-test} that translates the query and performs zero-shot inference in a high-resource language, i.e., English, and 3) \textbf{translate-test ICL} that simply combines \textbf{translate-test} and monolingual \textbf{ICL}. We measure the $\Delta$Weighted F1 against a simple \textbf{zero-shot prompting} over all languages under study. For all experiments that include translation, we utilize MT models from NLLB~\cite{nllb2022nllb}.\footnote{\url{https://huggingface.co/facebook/nllb-200-distilled-1.3B}}

Based on our experiment results shown in Figure~\ref{fig:result_ok_xicl}, the \textbf{translate-test} slightly improves the performance from the zero-shot baseline in BLOOM and XGLM, while \textbf{in-context query alignment} only improves zero-shot performance on XGLM. This indicates that alignment information only offers a limited benefit to improving LLMs' understanding. Additionally, all ICL approaches improve the performance over zero-shot prompting in most cases.
All approaches with similarity-based retrieval, i.e., \textbf{ICL Semantic} and \textbf{X-ICL Semantic} achieve higher scores than random retrievals, i.e., \textbf{ICL Random} and \textbf{X-ICL Random}, showing the importance of semantic similarity for exemplar retrievals. 
Interestingly, \textbf{X-ICL Semantic} yields a similar performance to \textbf{ICL Semantic}, which utilizes the target language exemplars. This indicates X-ICL can be a good alternative for low-resource languages as the available data in the specified low-resource language are commonly very limited.
% This shows that learning a few related languages during pre-training, although with only a tiny fraction of data, can improve the understanding of LLMs on the unseen languages, e.g., learning Indonesian helps to understand low-resource languages spoken across Indonesia.
% ~\footnote{there is also a possibility that the unseen languages are not truly unseen due to the misclassification from the language identification tools used for tagging the pretraining data.}
Above all, \textbf{translate-test ICL} yields the highest improvement amongst all methods, but this only happens when the machine translation quality is above a certain quality standard. We ablate the effect of machine translation quality to the \textbf{translate-test ICL} performance on Appendix~\ref{app:machine-translation-effect}.
% A similar observation is also observed in prior multilingual research works on encoder-only models~\cite{conneau-etal-2018-xnli,conneau2020unsupervised,ruder-etal-2021-xtreme}. 

To conclude, we offer the following suggestions to improve the low-resource language performance during inference: 1) When tackling low-resource languages, it is best to have a high-quality translation system accompanied by a source language task-specific data for \textbf{translate-test ICL}; 2) When there is no machine translation (MT) system for the specified language, it is best to use either \textbf{ICL} or \textbf{X-ICL} depending on the corpus availability; 3) When there is an MT system, but no task-specific data, \textbf{translate-test} is still the best option; and 4) When there is no high-quality MT system nor task-specific data, the best way is to use a parallel data to utilize \textbf{in-context query-alignment}.

% despite being outperformed by translate-test ICL baselines, X-ICL is still relevant, especially in the case where there is no monolingual corpus for the particular task and no machine translation system for the specified target language.

% \subsection{Key Insights and Beyond}
% \dummy{\lipsum[3-4]}

\section{Conclusion}

% We conduct an extensive exploration of X-ICL with LLMs focusing on low-resource languages. Our work sheds light on various aspects of X-ICL with LLMs. Our findings on in-context alignment showcase the failure of in-context label alignment, and we introduce a more effective alternative, namely in-context query alignment. Attempts to improve X-ICL via cross-lingual formatting consistency demonstrate a marginal impact on low-resource languages. Our study on cross-lingual retrieval approaches highlights the importance of cross-lingual semantic similarity models in X-ICL. Lastly, we assess the effectiveness of X-ICL on low-resource languages. Our finding suggests that despite being underperformed by monolingual and translate-test ICL baselines, X-ICL is still relevant, especially in the case where there is no monolingual corpus for the particular task and no machine translation system is available for the target language, which is prevalent in scenarios involving extremely low-resource language(s).

We systematically investigate the application of X-ICL with LLMs, focusing on low-resource languages. Our comprehensive analysis sheds light on multiple facets of X-ICL with LLMs. Our examination of in-context alignment reveals the limitation of label alignment, thus we suggest a more effective alternative: query alignment. Efforts to enhance X-ICL via formatting consistency only exhibit a marginal impact on low-resource languages. Our exploration of exemplar retrieval approaches underscores the significance of employing cross-lingual semantic similarity in X-ICL. Lastly, we analyze the effectiveness of X-ICL in the context of low-resource languages. Despite being outperformed by translate-test ICL, X-ICL remains relevant, especially when there is no MT model available for the target language---a circumstance prevalent in low-resource language scenarios.

\section*{Acknowledgements}

This work is partially funded by the PF20-43679 Hong Kong PhD Fellowship Scheme, Research Grant Council, Hong Kong; the Hong Kong Fellowship Scheme by the Hong Kong Research Grants Council (RGC); and the National Research Foundation, Singapore under its AI Singapore Programme.

\section*{Ethics Statement}

% Our X-ICL investigation highlights the contrasting experimental results obtained from cross-lingual label alignment with a previous work~\cite{tanwar2023multilingual} and highlights the positive impact obtained from cross-lingual query alignment, which could help the models understand the target language. the significance of cross-lingual semantic similarity models in X-ICL. Our work emphasizes the continued relevance of X-ICL in scenarios where no monolingual corpus or machine translation system is available, particularly in profoundly low-resource language contexts.

Our exploration of ICL and X-ICL methods addresses the linguistic data gap in low-resource languages. In scenarios where no monolingual corpus or machine translation system exists, our work underscores the significance of ICL and X-ICL as a viable solution. By investigating the limitations of in-context label alignment and proposing a more effective in-context query alignment approach, we aim to enhance the applicability of ICL and X-ICL on low-resource languages. This research is motivated by the need to provide computational solutions for languages lacking adequate linguistic resources for LM training. Our findings emphasize that ICL and X-ICL are useful in scenarios where alternative resources are absent, promoting linguistic diversity and inclusivity in the development of language technologies. All the datasets used in our experiments follow the license and term of use of the datasets.

\section*{Limitation}

\paragraph{Limited Coverage of low-resource Languages} We put our best effort into collecting datasets from various low-resource languages and, in the end, we ended up with the three low-resource datasets, i.e, MasakhaNews~\cite{adelani2023masakhanews}, NusaTranslation~\cite{cahyawijaya2023nusawrites}, and AmericasNLI~\cite{ebrahimi2022americasnli}, which suits our cases as these languages have parallel datasets which correspond to one or more high-resource languages and have large enough high-quality labeled datasets for both ICL and evaluation purposes. Furthermore, our study covers broad enough linguistics aspects of multilingual and cross-lingual within these three datasets, including various linguistics distances with the source languages --- from Nigerian Pidgin (pcm) to obscure regional languages such as Batak (btw), Hausa (hau), and Guarani (grn) ---, broad linguistic and geographic diversity ---the low-resource languages under study covers >10 language families from three different continents ---, and the incorporation of different scripts between source and target languages --- in the case of Amharic as a low-resource language and Arabic as a higher-resource language ---.  We leave the study of other and broader scales of low-resource languages for future work.

\paragraph{Choice of Multilingual High-Resource Language Datasets} Prior work on X-ICL with alignment~\cite{tanwar2023multilingual} conduct their study on Multilingual Amazon Review Corpus (MARC)~\cite{keung-etal-2020-multilingual}, Cross-language Sentiment (CLS)~\cite{prettenhofer-stein-2010-cross}, and HatEval~\cite{basile-etal-2019-semeval}. We considered using these datasets as our high-resource languages dataset. Nonetheless, we found that both MARC and CLS datasets are no longer available~\footnote{We checked the MARC dataset from the Hugging Face hub URL (\url{https://huggingface.co/datasets/amazon_reviews_multi}) and the original Amazon Web Service S3 Bucket (\url{https://s3.console.aws.amazon.com/s3/buckets/amazon-reviews-ml}). While for the CLS dataset, we checked the original dataset link in the paper (\url{http://www.webis.de/research/corpora/webis-cls-10/}).}, leaving us with only HatEval dataset. Since HatEval only covers English and Spanish, we do not incorporate it in our study. Instead, we incorporate the TweetSentimentMultilingual dataset~\cite{barbieri-etal-2022-xlmt} which covers 7 relatively high-resource languages in our study. We leave the exploration of other high-resource languages to future work.

\paragraph{Task Coverage} Given the nature of low-resource languages, there are only a handful of datasets available as downstream tasks. We suggest future works to explore the generalization of our approach to a broader task coverage, especially on datasets that cover more culturally relevant nuances of the corresponding low-resource language~\cite{aji-etal-2022-one,kabra-etal-2023-multi,lovenia2024seacrowd}.

\paragraph{Exploration on Other LLMs} We conduct our experiments with a single RTX3090 (24GB) GPU. Due to the large cost of inference and limited computation budget, we do not experiment on larger multilingual LLMs such as Falcon~\cite{falcon40b} and MPT~\cite{MosaicML2023Introducing}. Nonetheless, exploration on incorporating ICL and X-ICL with random exemplar retrieval with larger LLMs and the scaling effect on low-resource settings, such as low-resource languages and code-switching, have been discussed in prior works~\cite{zhang2023multilingual,asai2023buffet,cahyawijaya2024cendol}. We hypothesize that the scaling behavior of our work will follow the same trend.

\bibliography{anthology,custom}

\begin{thebibliography}{71}
\expandafter\ifx\csname natexlab\endcsname\relax\def\natexlab#1{#1}\fi

\bibitem[{Adelani et~al.(2022{\natexlab{a}})Adelani, Alabi, Fan, Kreutzer, Shen, Reid, Ruiter, Klakow, Nabende, Chang, Gwadabe, Sackey, Dossou, Emezue, Leong, Beukman, Muhammad, Jarso, Yousuf, Niyongabo~Rubungo, Hacheme, Wairagala, Nasir, Ajibade, Ajayi, Gitau, Abbott, Ahmed, Ochieng, Aremu, Ogayo, Mukiibi, Ouoba~Kabore, Kalipe, Mbaye, Tapo, Memdjokam~Koagne, Munkoh-Buabeng, Wagner, Abdulmumin, Awokoya, Buzaaba, Sibanda, Bukula, and Manthalu}]{adelani-etal-2022-thousand}
David Adelani, Jesujoba Alabi, Angela Fan, Julia Kreutzer, Xiaoyu Shen, Machel Reid, Dana Ruiter, Dietrich Klakow, Peter Nabende, Ernie Chang, Tajuddeen Gwadabe, Freshia Sackey, Bonaventure F.~P. Dossou, Chris Emezue, Colin Leong, Michael Beukman, Shamsuddeen Muhammad, Guyo Jarso, Oreen Yousuf, Andre Niyongabo~Rubungo, Gilles Hacheme, Eric~Peter Wairagala, Muhammad~Umair Nasir, Benjamin Ajibade, Tunde Ajayi, Yvonne Gitau, Jade Abbott, Mohamed Ahmed, Millicent Ochieng, Anuoluwapo Aremu, Perez Ogayo, Jonathan Mukiibi, Fatoumata Ouoba~Kabore, Godson Kalipe, Derguene Mbaye, Allahsera~Auguste Tapo, Victoire Memdjokam~Koagne, Edwin Munkoh-Buabeng, Valencia Wagner, Idris Abdulmumin, Ayodele Awokoya, Happy Buzaaba, Blessing Sibanda, Andiswa Bukula, and Sam Manthalu. 2022{\natexlab{a}}.
\newblock \href {https://doi.org/10.18653/v1/2022.naacl-main.223} {A few thousand translations go a long way! leveraging pre-trained models for {A}frican news translation}.
\newblock In \emph{Proceedings of the 2022 Conference of the North American Chapter of the Association for Computational Linguistics: Human Language Technologies}, pages 3053--3070, Seattle, United States. Association for Computational Linguistics.

\bibitem[{Adelani et~al.(2022{\natexlab{b}})Adelani, Neubig, Ruder, Rijhwani, Beukman, Palen-Michel, Lignos, Alabi, Muhammad, Nabende, Dione, Bukula, Mabuya, Dossou, Sibanda, Buzaaba, Mukiibi, Kalipe, Mbaye, Taylor, Kabore, Emezue, Aremu, Ogayo, Gitau, Munkoh-Buabeng, Memdjokam~Koagne, Tapo, Macucwa, Marivate, Elvis, Gwadabe, Adewumi, Ahia, Nakatumba-Nabende, Mokono, Ezeani, Chukwuneke, Oluwaseun~Adeyemi, Hacheme, Abdulmumin, Ogundepo, Yousuf, Moteu, and Klakow}]{adelani2022masakhaner2}
David Adelani, Graham Neubig, Sebastian Ruder, Shruti Rijhwani, Michael Beukman, Chester Palen-Michel, Constantine Lignos, Jesujoba Alabi, Shamsuddeen Muhammad, Peter Nabende, Cheikh M.~Bamba Dione, Andiswa Bukula, Rooweither Mabuya, Bonaventure F.~P. Dossou, Blessing Sibanda, Happy Buzaaba, Jonathan Mukiibi, Godson Kalipe, Derguene Mbaye, Amelia Taylor, Fatoumata Kabore, Chris~Chinenye Emezue, Anuoluwapo Aremu, Perez Ogayo, Catherine Gitau, Edwin Munkoh-Buabeng, Victoire Memdjokam~Koagne, Allahsera~Auguste Tapo, Tebogo Macucwa, Vukosi Marivate, Mboning~Tchiaze Elvis, Tajuddeen Gwadabe, Tosin Adewumi, Orevaoghene Ahia, Joyce Nakatumba-Nabende, Neo~Lerato Mokono, Ignatius Ezeani, Chiamaka Chukwuneke, Mofetoluwa Oluwaseun~Adeyemi, Gilles~Quentin Hacheme, Idris Abdulmumin, Odunayo Ogundepo, Oreen Yousuf, Tatiana Moteu, and Dietrich Klakow. 2022{\natexlab{b}}.
\newblock \href {https://aclanthology.org/2022.emnlp-main.298} {{M}asakha{NER} 2.0: {A}frica-centric transfer learning for named entity recognition}.
\newblock In \emph{Proceedings of the 2022 Conference on Empirical Methods in Natural Language Processing}, pages 4488--4508, Abu Dhabi, United Arab Emirates. Association for Computational Linguistics.

\bibitem[{Adelani et~al.(2021)Adelani, Abbott, Neubig, D’souza, Kreutzer, Lignos, Palen-Michel, Buzaaba, Rijhwani, Ruder, Mayhew, Azime, Muhammad, Emezue, Nakatumba-Nabende, Ogayo, Anuoluwapo, Gitau, Mbaye, Alabi, Yimam, Gwadabe, Ezeani, Niyongabo, Mukiibi, Otiende, Orife, David, Ngom, Adewumi, Rayson, Adeyemi, Muriuki, Anebi, Chukwuneke, Odu, Wairagala, Oyerinde, Siro, Bateesa, Oloyede, Wambui, Akinode, Nabagereka, Katusiime, Awokoya, MBOUP, Gebreyohannes, Tilaye, Nwaike, Wolde, Faye, Sibanda, Ahia, Dossou, Ogueji, DIOP, Diallo, Akinfaderin, Marengereke, and Osei}]{adelani2021masakhaner}
David~Ifeoluwa Adelani, Jade Abbott, Graham Neubig, Daniel D’souza, Julia Kreutzer, Constantine Lignos, Chester Palen-Michel, Happy Buzaaba, Shruti Rijhwani, Sebastian Ruder, Stephen Mayhew, Israel~Abebe Azime, Shamsuddeen~H. Muhammad, Chris~Chinenye Emezue, Joyce Nakatumba-Nabende, Perez Ogayo, Aremu Anuoluwapo, Catherine Gitau, Derguene Mbaye, Jesujoba Alabi, Seid~Muhie Yimam, Tajuddeen~Rabiu Gwadabe, Ignatius Ezeani, Rubungo~Andre Niyongabo, Jonathan Mukiibi, Verrah Otiende, Iroro Orife, Davis David, Samba Ngom, Tosin Adewumi, Paul Rayson, Mofetoluwa Adeyemi, Gerald Muriuki, Emmanuel Anebi, Chiamaka Chukwuneke, Nkiruka Odu, Eric~Peter Wairagala, Samuel Oyerinde, Clemencia Siro, Tobius~Saul Bateesa, Temilola Oloyede, Yvonne Wambui, Victor Akinode, Deborah Nabagereka, Maurice Katusiime, Ayodele Awokoya, Mouhamadane MBOUP, Dibora Gebreyohannes, Henok Tilaye, Kelechi Nwaike, Degaga Wolde, Abdoulaye Faye, Blessing Sibanda, Orevaoghene Ahia, Bonaventure F.~P. Dossou, Kelechi Ogueji, Thierno~Ibrahima DIOP,
  Abdoulaye Diallo, Adewale Akinfaderin, Tendai Marengereke, and Salomey Osei. 2021.
\newblock \href {https://doi.org/10.1162/tacl_a_00416} {{MasakhaNER: Named Entity Recognition for African Languages}}.
\newblock \emph{Transactions of the Association for Computational Linguistics}, 9:1116--1131.

\bibitem[{Adelani et~al.(2023)Adelani, Masiak, Azime, Alabi, Tonja, Mwase, Ogundepo, Dossou, Oladipo, Nixdorf, Emezue, sana~al azzawi, Sibanda, David, Ndolela, Mukiibi, Ajayi, Moteu, Odhiambo, Owodunni, Obiefuna, Mohamed, Muhammad, Ababu, Salahudeen, Yigezu, Gwadabe, Abdulmumin, Taye, Awoyomi, Shode, Adelani, Abdulganiyu, Omotayo, Adeeko, Afolabi, Aremu, Samuel, Siro, Kimotho, Ogbu, Mbonu, Chukwuneke, Fanijo, Ojo, Awosan, Kebede, Sakayo, Nyatsine, Sidume, Yousuf, Oduwole, Tshinu, Kimanuka, Diko, Nxakama, Nigusse, Johar, Mohamed, Hassan, Mehamed, Ngabire, Jules, Ssenkungu, and Stenetorp}]{adelani2023masakhanews}
David~Ifeoluwa Adelani, Marek Masiak, Israel~Abebe Azime, Jesujoba Alabi, Atnafu~Lambebo Tonja, Christine Mwase, Odunayo Ogundepo, Bonaventure F.~P. Dossou, Akintunde Oladipo, Doreen Nixdorf, Chris~Chinenye Emezue, sana~al azzawi, Blessing Sibanda, Davis David, Lolwethu Ndolela, Jonathan Mukiibi, Tunde Ajayi, Tatiana Moteu, Brian Odhiambo, Abraham Owodunni, Nnaemeka Obiefuna, Muhidin Mohamed, Shamsuddeen~Hassan Muhammad, Teshome~Mulugeta Ababu, Saheed~Abdullahi Salahudeen, Mesay~Gemeda Yigezu, Tajuddeen Gwadabe, Idris Abdulmumin, Mahlet Taye, Oluwabusayo Awoyomi, Iyanuoluwa Shode, Tolulope Adelani, Habiba Abdulganiyu, Abdul-Hakeem Omotayo, Adetola Adeeko, Abeeb Afolabi, Anuoluwapo Aremu, Olanrewaju Samuel, Clemencia Siro, Wangari Kimotho, Onyekachi Ogbu, Chinedu Mbonu, Chiamaka Chukwuneke, Samuel Fanijo, Jessica Ojo, Oyinkansola Awosan, Tadesse Kebede, Toadoum~Sari Sakayo, Pamela Nyatsine, Freedmore Sidume, Oreen Yousuf, Mardiyyah Oduwole, Tshinu Tshinu, Ussen Kimanuka, Thina Diko, Siyanda Nxakama, Sinodos
  Nigusse, Abdulmejid Johar, Shafie Mohamed, Fuad~Mire Hassan, Moges~Ahmed Mehamed, Evrard Ngabire, Jules Jules, Ivan Ssenkungu, and Pontus Stenetorp. 2023.
\newblock \href {http://arxiv.org/abs/2304.09972} {Masakhanews: News topic classification for african languages}.

\bibitem[{Adilazuarda et~al.(2024)Adilazuarda, Cahyawijaya, Aji, Winata, and Purwarianti}]{adilazuarda2024lingualchemy}
Muhammad~Farid Adilazuarda, Samuel Cahyawijaya, Alham~Fikri Aji, Genta~Indra Winata, and Ayu Purwarianti. 2024.
\newblock \href {http://arxiv.org/abs/2401.06034} {Lingualchemy: Fusing typological and geographical elements for unseen language generalization}.

\bibitem[{Adilazuarda et~al.(2023)Adilazuarda, Cahyawijaya, and Purwarianti}]{adilazuarda2023obscure}
Muhammad~Farid Adilazuarda, Samuel Cahyawijaya, and Ayu Purwarianti. 2023.
\newblock \href {http://arxiv.org/abs/2311.12375} {The obscure limitation of modular multilingual language models}.

\bibitem[{Aji et~al.(2022{\natexlab{a}})Aji, Winata, Koto, Cahyawijaya, Romadhony, Mahendra, Kurniawan, Moeljadi, Prasojo, Baldwin, Lau, and Ruder}]{aji2022one}
Alham~Fikri Aji, Genta~Indra Winata, Fajri Koto, Samuel Cahyawijaya, Ade Romadhony, Rahmad Mahendra, Kemal Kurniawan, David Moeljadi, Radityo~Eko Prasojo, Timothy Baldwin, Jey~Han Lau, and Sebastian Ruder. 2022{\natexlab{a}}.
\newblock \href {https://doi.org/10.18653/v1/2022.acl-long.500} {One country, 700+ languages: {NLP} challenges for underrepresented languages and dialects in {I}ndonesia}.
\newblock In \emph{Proceedings of the 60th Annual Meeting of the Association for Computational Linguistics (Volume 1: Long Papers)}, pages 7226--7249, Dublin, Ireland. Association for Computational Linguistics.

\bibitem[{Aji et~al.(2022{\natexlab{b}})Aji, Winata, Koto, Cahyawijaya, Romadhony, Mahendra, Kurniawan, Moeljadi, Prasojo, Baldwin, Lau, and Ruder}]{aji-etal-2022-one}
Alham~Fikri Aji, Genta~Indra Winata, Fajri Koto, Samuel Cahyawijaya, Ade Romadhony, Rahmad Mahendra, Kemal Kurniawan, David Moeljadi, Radityo~Eko Prasojo, Timothy Baldwin, Jey~Han Lau, and Sebastian Ruder. 2022{\natexlab{b}}.
\newblock \href {https://doi.org/10.18653/v1/2022.acl-long.500} {One country, 700+ languages: {NLP} challenges for underrepresented languages and dialects in {I}ndonesia}.
\newblock In \emph{Proceedings of the 60th Annual Meeting of the Association for Computational Linguistics (Volume 1: Long Papers)}, pages 7226--7249, Dublin, Ireland. Association for Computational Linguistics.

\bibitem[{Almazrouei et~al.(2023)Almazrouei, Alobeidli, Alshamsi, Cappelli, Cojocaru, Debbah, Goffinet, Heslow, Launay, Malartic, Noune, Pannier, and Penedo}]{falcon40b}
Ebtesam Almazrouei, Hamza Alobeidli, Abdulaziz Alshamsi, Alessandro Cappelli, Ruxandra Cojocaru, Merouane Debbah, Etienne Goffinet, Daniel Heslow, Julien Launay, Quentin Malartic, Badreddine Noune, Baptiste Pannier, and Guilherme Penedo. 2023.
\newblock {Falcon-40B}: an open large language model with state-of-the-art performance.

\bibitem[{Asai et~al.(2023)Asai, Kudugunta, Yu, Blevins, Gonen, Reid, Tsvetkov, Ruder, and Hajishirzi}]{asai2023buffet}
Akari Asai, Sneha Kudugunta, Xinyan~Velocity Yu, Terra Blevins, Hila Gonen, Machel Reid, Yulia Tsvetkov, Sebastian Ruder, and Hannaneh Hajishirzi. 2023.
\newblock \href {http://arxiv.org/abs/2305.14857} {Buffet: Benchmarking large language models for few-shot cross-lingual transfer}.

\bibitem[{Bandarkar et~al.(2023)Bandarkar, Liang, Muller, Artetxe, Shukla, Husa, Goyal, Krishnan, Zettlemoyer, and Khabsa}]{bandarkar2023belebele}
Lucas Bandarkar, Davis Liang, Benjamin Muller, Mikel Artetxe, Satya~Narayan Shukla, Donald Husa, Naman Goyal, Abhinandan Krishnan, Luke Zettlemoyer, and Madian Khabsa. 2023.
\newblock The belebele benchmark: a parallel reading comprehension dataset in 122 language variants.
\newblock \emph{arXiv preprint arXiv:2308.16884}.

\bibitem[{Bang et~al.(2023)Bang, Cahyawijaya, Lee, Dai, Su, Wilie, Lovenia, Ji, Yu, Chung, Do, Xu, and Fung}]{bang2023multitask}
Yejin Bang, Samuel Cahyawijaya, Nayeon Lee, Wenliang Dai, Dan Su, Bryan Wilie, Holy Lovenia, Ziwei Ji, Tiezheng Yu, Willy Chung, Quyet~V. Do, Yan Xu, and Pascale Fung. 2023.
\newblock \href {http://arxiv.org/abs/2302.04023} {A multitask, multilingual, multimodal evaluation of chatgpt on reasoning, hallucination, and interactivity}.

\bibitem[{Barbieri et~al.(2022)Barbieri, Espinosa~Anke, and Camacho-Collados}]{barbieri-etal-2022-xlmt}
Francesco Barbieri, Luis Espinosa~Anke, and Jose Camacho-Collados. 2022.
\newblock \href {https://aclanthology.org/2022.lrec-1.27} {{XLM}-{T}: Multilingual language models in {T}witter for sentiment analysis and beyond}.
\newblock In \emph{Proceedings of the Thirteenth Language Resources and Evaluation Conference}, pages 258--266, Marseille, France. European Language Resources Association.

\bibitem[{Basile et~al.(2019)Basile, Bosco, Fersini, Nozza, Patti, Rangel~Pardo, Rosso, and Sanguinetti}]{basile-etal-2019-semeval}
Valerio Basile, Cristina Bosco, Elisabetta Fersini, Debora Nozza, Viviana Patti, Francisco~Manuel Rangel~Pardo, Paolo Rosso, and Manuela Sanguinetti. 2019.
\newblock \href {https://doi.org/10.18653/v1/S19-2007} {{S}em{E}val-2019 task 5: Multilingual detection of hate speech against immigrants and women in {T}witter}.
\newblock In \emph{Proceedings of the 13th International Workshop on Semantic Evaluation}, pages 54--63, Minneapolis, Minnesota, USA. Association for Computational Linguistics.

\bibitem[{Brown et~al.(2020{\natexlab{a}})Brown, Mann, Ryder, Subbiah, Kaplan, Dhariwal, Neelakantan, Shyam, Sastry, Askell, Agarwal, Herbert-Voss, Krueger, Henighan, Child, Ramesh, Ziegler, Wu, Winter, Hesse, Chen, Sigler, Litwin, Gray, Chess, Clark, Berner, McCandlish, Radford, Sutskever, and Amodei}]{brown2020gpt3}
Tom Brown, Benjamin Mann, Nick Ryder, Melanie Subbiah, Jared~D Kaplan, Prafulla Dhariwal, Arvind Neelakantan, Pranav Shyam, Girish Sastry, Amanda Askell, Sandhini Agarwal, Ariel Herbert-Voss, Gretchen Krueger, Tom Henighan, Rewon Child, Aditya Ramesh, Daniel Ziegler, Jeffrey Wu, Clemens Winter, Chris Hesse, Mark Chen, Eric Sigler, Mateusz Litwin, Scott Gray, Benjamin Chess, Jack Clark, Christopher Berner, Sam McCandlish, Alec Radford, Ilya Sutskever, and Dario Amodei. 2020{\natexlab{a}}.
\newblock \href {https://proceedings.neurips.cc/paper_files/paper/2020/file/1457c0d6bfcb4967418bfb8ac142f64a-Paper.pdf} {Language models are few-shot learners}.
\newblock In \emph{Advances in Neural Information Processing Systems}, volume~33, pages 1877--1901. Curran Associates, Inc.

\bibitem[{Brown et~al.(2020{\natexlab{b}})Brown, Mann, Ryder, Subbiah, Kaplan, Dhariwal, Neelakantan, Shyam, Sastry, Askell et~al.}]{brown2020language}
Tom Brown, Benjamin Mann, Nick Ryder, Melanie Subbiah, Jared~D Kaplan, Prafulla Dhariwal, Arvind Neelakantan, Pranav Shyam, Girish Sastry, Amanda Askell, et~al. 2020{\natexlab{b}}.
\newblock Language models are few-shot learners.
\newblock \emph{Advances in neural information processing systems}, 33:1877--1901.

\bibitem[{Cahyawijaya et~al.(2023{\natexlab{a}})Cahyawijaya, Lovenia, Aji, Winata, Wilie, Mahendra, Wibisono, Romadhony, Vincentio, Koto, Santoso, Moeljadi, Wirawan, Hudi, Parmonangan, Alfina, Wicaksono, Putra, Rahmadani, Oenang, Septiandri, Jaya, Dhole, Suryani, Putri, Su, Stevens, Nityasya, Adilazuarda, Ignatius, Diandaru, Yu, Ghifari, Dai, Xu, Damapuspita, Tho, Karo, Fatyanosa, Ji, Fung, Neubig, Baldwin, Ruder, Sujaini, Sakti, and Purwarianti}]{cahyawijaya2023nusacrowd}
Samuel Cahyawijaya, Holy Lovenia, Alham~Fikri Aji, Genta~Indra Winata, Bryan Wilie, Rahmad Mahendra, Christian Wibisono, Ade Romadhony, Karissa Vincentio, Fajri Koto, Jennifer Santoso, David Moeljadi, Cahya Wirawan, Frederikus Hudi, Ivan~Halim Parmonangan, Ika Alfina, Muhammad~Satrio Wicaksono, Ilham~Firdausi Putra, Samsul Rahmadani, Yulianti Oenang, Ali~Akbar Septiandri, James Jaya, Kaustubh~D. Dhole, Arie~Ardiyanti Suryani, Rifki~Afina Putri, Dan Su, Keith Stevens, Made~Nindyatama Nityasya, Muhammad~Farid Adilazuarda, Ryan Ignatius, Ryandito Diandaru, Tiezheng Yu, Vito Ghifari, Wenliang Dai, Yan Xu, Dyah Damapuspita, Cuk Tho, Ichwanul Muslim~Karo Karo, Tirana~Noor Fatyanosa, Ziwei Ji, Pascale Fung, Graham Neubig, Timothy Baldwin, Sebastian Ruder, Herry Sujaini, Sakriani Sakti, and Ayu Purwarianti. 2023{\natexlab{a}}.
\newblock \href {http://arxiv.org/abs/2212.09648} {Nusacrowd: Open source initiative for indonesian nlp resources}.

\bibitem[{Cahyawijaya et~al.(2023{\natexlab{b}})Cahyawijaya, Lovenia, Koto, Adhista, Dave, Oktavianti, Akbar, Lee, Shadieq, Cenggoro, linuwih, Wilie, Muridan, Winata, Moeljadi, Aji, Purwarianti, and Fung}]{cahyawijaya2023nusawrites}
Samuel Cahyawijaya, Holy Lovenia, Fajri Koto, Dea Adhista, Emmanuel Dave, Sarah Oktavianti, Salsabil Akbar, Jhonson Lee, Nuur Shadieq, Tjeng~Wawan Cenggoro, hanung linuwih, Bryan Wilie, Galih Muridan, Genta Winata, David Moeljadi, Alham~Fikri Aji, Ayu Purwarianti, and Pascale Fung. 2023{\natexlab{b}}.
\newblock \href {https://aclanthology.org/2023.ijcnlp-long.60} {Nusawrites: Constructing high-quality corpora for underrepresented and extremely low-resource languages}.

\bibitem[{Cahyawijaya et~al.(2024)Cahyawijaya, Lovenia, Koto, Putri, Dave, Lee, Shadieq, Cenggoro, Akbar, Mahendra, Putri, Wilie, Winata, Aji, Purwarianti, and Fung}]{cahyawijaya2024cendol}
Samuel Cahyawijaya, Holy Lovenia, Fajri Koto, Rifki~Afina Putri, Emmanuel Dave, Jhonson Lee, Nuur Shadieq, Wawan Cenggoro, Salsabil~Maulana Akbar, Muhammad~Ihza Mahendra, Dea~Annisayanti Putri, Bryan Wilie, Genta~Indra Winata, Alham~Fikri Aji, Ayu Purwarianti, and Pascale Fung. 2024.
\newblock \href {http://arxiv.org/abs/2404.06138} {Cendol: Open instruction-tuned generative large language models for indonesian languages}.

\bibitem[{Cahyawijaya et~al.(2023{\natexlab{c}})Cahyawijaya, Lovenia, Yu, Chung, and Fung}]{cahyawijaya2023instructalign}
Samuel Cahyawijaya, Holy Lovenia, Tiezheng Yu, Willy Chung, and Pascale Fung. 2023{\natexlab{c}}.
\newblock \href {http://arxiv.org/abs/2305.13627} {Instruct-align: Teaching novel languages with to llms through alignment-based cross-lingual instruction}.

\bibitem[{Cahyawijaya et~al.(2021)Cahyawijaya, Winata, Wilie, Vincentio, Li, Kuncoro, Ruder, Lim, Bahar, Khodra, Purwarianti, and Fung}]{cahyawijaya2021indonlg}
Samuel Cahyawijaya, Genta~Indra Winata, Bryan Wilie, Karissa Vincentio, Xiaohong Li, Adhiguna Kuncoro, Sebastian Ruder, Zhi~Yuan Lim, Syafri Bahar, Masayu Khodra, Ayu Purwarianti, and Pascale Fung. 2021.
\newblock \href {https://doi.org/10.18653/v1/2021.emnlp-main.699} {{I}ndo{NLG}: Benchmark and resources for evaluating {I}ndonesian natural language generation}.
\newblock In \emph{Proceedings of the 2021 Conference on Empirical Methods in Natural Language Processing}, pages 8875--8898, Online and Punta Cana, Dominican Republic. Association for Computational Linguistics.

\bibitem[{Chaudhary et~al.(2019)Chaudhary, Tang, Guzm{\'a}n, Schwenk, and Koehn}]{chaudhary-etal-2019-low}
Vishrav Chaudhary, Yuqing Tang, Francisco Guzm{\'a}n, Holger Schwenk, and Philipp Koehn. 2019.
\newblock \href {https://doi.org/10.18653/v1/W19-5435} {Low-resource corpus filtering using multilingual sentence embeddings}.
\newblock In \emph{Proceedings of the Fourth Conference on Machine Translation (Volume 3: Shared Task Papers, Day 2)}, pages 261--266, Florence, Italy. Association for Computational Linguistics.

\bibitem[{Chaudhry et~al.(2019)Chaudhry, Rohrbach, Elhoseiny, Ajanthan, Dokania, Torr, and Ranzato}]{chaudhry2019tiny}
Arslan Chaudhry, Marcus Rohrbach, Mohamed Elhoseiny, Thalaiyasingam Ajanthan, Puneet~K. Dokania, Philip H.~S. Torr, and Marc'Aurelio Ranzato. 2019.
\newblock \href {http://arxiv.org/abs/1902.10486} {On tiny episodic memories in continual learning}.

\bibitem[{Chowdhery et~al.(2022)Chowdhery, Narang, Devlin, Bosma, Mishra, Roberts, Barham, Chung, Sutton, Gehrmann, Schuh, Shi, Tsvyashchenko, Maynez, Rao, Barnes, Tay, Shazeer, Prabhakaran, Reif, Du, Hutchinson, Pope, Bradbury, Austin, Isard, Gur-Ari, Yin, Duke, Levskaya, Ghemawat, Dev, Michalewski, Garcia, Misra, Robinson, Fedus, Zhou, Ippolito, Luan, Lim, Zoph, Spiridonov, Sepassi, Dohan, Agrawal, Omernick, Dai, Pillai, Pellat, Lewkowycz, Moreira, Child, Polozov, Lee, Zhou, Wang, Saeta, Diaz, Firat, Catasta, Wei, Meier-Hellstern, Eck, Dean, Petrov, and Fiedel}]{chowdhery2022palm}
Aakanksha Chowdhery, Sharan Narang, Jacob Devlin, Maarten Bosma, Gaurav Mishra, Adam Roberts, Paul Barham, Hyung~Won Chung, Charles Sutton, Sebastian Gehrmann, Parker Schuh, Kensen Shi, Sasha Tsvyashchenko, Joshua Maynez, Abhishek Rao, Parker Barnes, Yi~Tay, Noam Shazeer, Vinodkumar Prabhakaran, Emily Reif, Nan Du, Ben Hutchinson, Reiner Pope, James Bradbury, Jacob Austin, Michael Isard, Guy Gur-Ari, Pengcheng Yin, Toju Duke, Anselm Levskaya, Sanjay Ghemawat, Sunipa Dev, Henryk Michalewski, Xavier Garcia, Vedant Misra, Kevin Robinson, Liam Fedus, Denny Zhou, Daphne Ippolito, David Luan, Hyeontaek Lim, Barret Zoph, Alexander Spiridonov, Ryan Sepassi, David Dohan, Shivani Agrawal, Mark Omernick, Andrew~M. Dai, Thanumalayan~Sankaranarayana Pillai, Marie Pellat, Aitor Lewkowycz, Erica Moreira, Rewon Child, Oleksandr Polozov, Katherine Lee, Zongwei Zhou, Xuezhi Wang, Brennan Saeta, Mark Diaz, Orhan Firat, Michele Catasta, Jason Wei, Kathy Meier-Hellstern, Douglas Eck, Jeff Dean, Slav Petrov, and Noah Fiedel. 2022.
\newblock \href {http://arxiv.org/abs/2204.02311} {Palm: Scaling language modeling with pathways}.

\bibitem[{Conneau et~al.(2018)Conneau, Rinott, Lample, Williams, Bowman, Schwenk, and Stoyanov}]{conneau-etal-2018-xnli}
Alexis Conneau, Ruty Rinott, Guillaume Lample, Adina Williams, Samuel Bowman, Holger Schwenk, and Veselin Stoyanov. 2018.
\newblock \href {https://doi.org/10.18653/v1/D18-1269} {{XNLI}: Evaluating cross-lingual sentence representations}.
\newblock In \emph{Proceedings of the 2018 Conference on Empirical Methods in Natural Language Processing}, pages 2475--2485, Brussels, Belgium. Association for Computational Linguistics.

\bibitem[{Ebrahimi et~al.(2022{\natexlab{a}})Ebrahimi, Mager, Oncevay, Chaudhary, Chiruzzo, Fan, Ortega, Ramos, Rios, Meza~Ruiz, Gim{\'e}nez-Lugo, Mager, Neubig, Palmer, Coto-Solano, Vu, and Kann}]{ebrahimi2022americasnli}
Abteen Ebrahimi, Manuel Mager, Arturo Oncevay, Vishrav Chaudhary, Luis Chiruzzo, Angela Fan, John Ortega, Ricardo Ramos, Annette Rios, Ivan~Vladimir Meza~Ruiz, Gustavo Gim{\'e}nez-Lugo, Elisabeth Mager, Graham Neubig, Alexis Palmer, Rolando Coto-Solano, Thang Vu, and Katharina Kann. 2022{\natexlab{a}}.
\newblock \href {https://doi.org/10.18653/v1/2022.acl-long.435} {{A}mericas{NLI}: Evaluating zero-shot natural language understanding of pretrained multilingual models in truly low-resource languages}.
\newblock In \emph{Proceedings of the 60th Annual Meeting of the Association for Computational Linguistics (Volume 1: Long Papers)}, pages 6279--6299, Dublin, Ireland. Association for Computational Linguistics.

\bibitem[{Ebrahimi et~al.(2022{\natexlab{b}})Ebrahimi, Mager, Oncevay, Chaudhary, Chiruzzo, Fan, Ortega, Ramos, Rios, Meza~Ruiz, Gim{\'e}nez-Lugo, Mager, Neubig, Palmer, Coto-Solano, Vu, and Kann}]{ebrahimi-etal-2022-americasnli}
Abteen Ebrahimi, Manuel Mager, Arturo Oncevay, Vishrav Chaudhary, Luis Chiruzzo, Angela Fan, John Ortega, Ricardo Ramos, Annette Rios, Ivan~Vladimir Meza~Ruiz, Gustavo Gim{\'e}nez-Lugo, Elisabeth Mager, Graham Neubig, Alexis Palmer, Rolando Coto-Solano, Thang Vu, and Katharina Kann. 2022{\natexlab{b}}.
\newblock \href {https://doi.org/10.18653/v1/2022.acl-long.435} {{A}mericas{NLI}: Evaluating zero-shot natural language understanding of pretrained multilingual models in truly low-resource languages}.
\newblock In \emph{Proceedings of the 60th Annual Meeting of the Association for Computational Linguistics (Volume 1: Long Papers)}, pages 6279--6299, Dublin, Ireland. Association for Computational Linguistics.

\bibitem[{El-Kishky et~al.(2020)El-Kishky, Chaudhary, Guzm{\'a}n, and Koehn}]{el-kishky-etal-2020-ccaligned}
Ahmed El-Kishky, Vishrav Chaudhary, Francisco Guzm{\'a}n, and Philipp Koehn. 2020.
\newblock \href {https://doi.org/10.18653/v1/2020.emnlp-main.480} {{CCA}ligned: A massive collection of cross-lingual web-document pairs}.
\newblock In \emph{Proceedings of the 2020 Conference on Empirical Methods in Natural Language Processing (EMNLP)}, pages 5960--5969, Online. Association for Computational Linguistics.

\bibitem[{Feng et~al.(2022)Feng, Yang, Cer, Arivazhagan, and Wang}]{feng2022labse}
Fangxiaoyu Feng, Yinfei Yang, Daniel Cer, Naveen Arivazhagan, and Wei Wang. 2022.
\newblock \href {https://doi.org/10.18653/v1/2022.acl-long.62} {Language-agnostic {BERT} sentence embedding}.
\newblock In \emph{Proceedings of the 60th Annual Meeting of the Association for Computational Linguistics (Volume 1: Long Papers)}, pages 878--891, Dublin, Ireland. Association for Computational Linguistics.

\bibitem[{French(1993)}]{robert1993catastrophic}
Robert~M. French. 1993.
\newblock Catastrophic interference in connectionist networks: Can it be predicted, can it be prevented?
\newblock In \emph{Proceedings of the 6th International Conference on Neural Information Processing Systems}, NIPS'93, page 1176–1177, San Francisco, CA, USA. Morgan Kaufmann Publishers Inc.

\bibitem[{Goyal et~al.(2022)Goyal, Gao, Chaudhary, Chen, Wenzek, Ju, Krishnan, Ranzato, Guzm{\'a}n, and Fan}]{goyal-etal-2022-flores}
Naman Goyal, Cynthia Gao, Vishrav Chaudhary, Peng-Jen Chen, Guillaume Wenzek, Da~Ju, Sanjana Krishnan, Marc{'}Aurelio Ranzato, Francisco Guzm{\'a}n, and Angela Fan. 2022.
\newblock \href {https://doi.org/10.1162/tacl_a_00474} {The {F}lores-101 evaluation benchmark for low-resource and multilingual machine translation}.
\newblock \emph{Transactions of the Association for Computational Linguistics}, 10:522--538.

\bibitem[{Jin et~al.(2023)Jin, Ren, Preotiuc-Pietro, and Cheng}]{jin2023dataless}
Xisen Jin, Xiang Ren, Daniel Preotiuc-Pietro, and Pengxiang Cheng. 2023.
\newblock \href {https://openreview.net/forum?id=FCnohuR6AnM} {Dataless knowledge fusion by merging weights of language models}.
\newblock In \emph{The Eleventh International Conference on Learning Representations}.

\bibitem[{Jones et~al.(2023)Jones, Caswell, Saxena, and Firat}]{jones2023bilex}
Alex Jones, Isaac Caswell, Ishank Saxena, and Orhan Firat. 2023.
\newblock \href {http://arxiv.org/abs/2303.15265} {Bilex rx: Lexical data augmentation for massively multilingual machine translation}.

\bibitem[{Kabra et~al.(2023)Kabra, Liu, Khanuja, Aji, Winata, Cahyawijaya, Aremu, Ogayo, and Neubig}]{kabra-etal-2023-multi}
Anubha Kabra, Emmy Liu, Simran Khanuja, Alham~Fikri Aji, Genta Winata, Samuel Cahyawijaya, Anuoluwapo Aremu, Perez Ogayo, and Graham Neubig. 2023.
\newblock \href {https://doi.org/10.18653/v1/2023.findings-acl.525} {Multi-lingual and multi-cultural figurative language understanding}.
\newblock In \emph{Findings of the Association for Computational Linguistics: ACL 2023}, pages 8269--8284, Toronto, Canada. Association for Computational Linguistics.

\bibitem[{Kakwani et~al.(2020)Kakwani, Kunchukuttan, Golla, N.C., Bhattacharyya, Khapra, and Kumar}]{kakwani2020indicnlpsuite}
Divyanshu Kakwani, Anoop Kunchukuttan, Satish Golla, Gokul N.C., Avik Bhattacharyya, Mitesh~M. Khapra, and Pratyush Kumar. 2020.
\newblock \href {https://doi.org/10.18653/v1/2020.findings-emnlp.445} {{I}ndic{NLPS}uite: Monolingual corpora, evaluation benchmarks and pre-trained multilingual language models for {I}ndian languages}.
\newblock In \emph{Findings of the Association for Computational Linguistics: EMNLP 2020}, pages 4948--4961, Online. Association for Computational Linguistics.

\bibitem[{Keung et~al.(2020)Keung, Lu, Szarvas, and Smith}]{keung-etal-2020-multilingual}
Phillip Keung, Yichao Lu, Gy{\"o}rgy Szarvas, and Noah~A. Smith. 2020.
\newblock \href {https://doi.org/10.18653/v1/2020.emnlp-main.369} {The multilingual {A}mazon reviews corpus}.
\newblock In \emph{Proceedings of the 2020 Conference on Empirical Methods in Natural Language Processing (EMNLP)}, pages 4563--4568, Online. Association for Computational Linguistics.

\bibitem[{Kojima et~al.(2022)Kojima, Gu, Reid, Matsuo, and Iwasawa}]{kojima2022large}
Takeshi Kojima, Shixiang~Shane Gu, Machel Reid, Yutaka Matsuo, and Yusuke Iwasawa. 2022.
\newblock \href {https://openreview.net/forum?id=6p3AuaHAFiN} {Large language models are zero-shot reasoners}.
\newblock In \emph{ICML 2022 Workshop on Knowledge Retrieval and Language Models}.

\bibitem[{Koto et~al.(2023)Koto, Aisyah, Li, and Baldwin}]{koto-etal-2023-large}
Fajri Koto, Nurul Aisyah, Haonan Li, and Timothy Baldwin. 2023.
\newblock \href {https://doi.org/10.18653/v1/2023.emnlp-main.760} {Large language models only pass primary school exams in {I}ndonesia: A comprehensive test on {I}ndo{MMLU}}.
\newblock In \emph{Proceedings of the 2023 Conference on Empirical Methods in Natural Language Processing}, pages 12359--12374, Singapore. Association for Computational Linguistics.

\bibitem[{Kumar et~al.(2022)Kumar, Shrotriya, Sahu, Mishra, Dabre, Puduppully, Kunchukuttan, Khapra, and Kumar}]{kumar2022indicnlg}
Aman Kumar, Himani Shrotriya, Prachi Sahu, Amogh Mishra, Raj Dabre, Ratish Puduppully, Anoop Kunchukuttan, Mitesh~M. Khapra, and Pratyush Kumar. 2022.
\newblock \href {https://aclanthology.org/2022.emnlp-main.360} {{I}ndic{NLG} benchmark: Multilingual datasets for diverse {NLG} tasks in {I}ndic languages}.
\newblock In \emph{Proceedings of the 2022 Conference on Empirical Methods in Natural Language Processing}, pages 5363--5394, Abu Dhabi, United Arab Emirates. Association for Computational Linguistics.

\bibitem[{Leong et~al.(2022)Leong, Nemecek, Mansdorfer, Filighera, Owodunni, and Whitenack}]{leong-etal-2022-bloom}
Colin Leong, Joshua Nemecek, Jacob Mansdorfer, Anna Filighera, Abraham Owodunni, and Daniel Whitenack. 2022.
\newblock \href {https://doi.org/10.18653/v1/2022.emnlp-main.590} {Bloom library: Multimodal datasets in 300+ languages for a variety of downstream tasks}.
\newblock In \emph{Proceedings of the 2022 Conference on Empirical Methods in Natural Language Processing}, pages 8608--8621, Abu Dhabi, United Arab Emirates. Association for Computational Linguistics.

\bibitem[{Liang et~al.(2023)Liang, Bommasani, Lee, Tsipras, Soylu, Yasunaga, Zhang, Narayanan, Wu, Kumar, Newman, Yuan, Yan, Zhang, Cosgrove, Manning, Re, Acosta-Navas, Hudson, Zelikman, Durmus, Ladhak, Rong, Ren, Yao, WANG, Santhanam, Orr, Zheng, Yuksekgonul, Suzgun, Kim, Guha, Chatterji, Khattab, Henderson, Huang, Chi, Xie, Santurkar, Ganguli, Hashimoto, Icard, Zhang, Chaudhary, Wang, Li, Mai, Zhang, and Koreeda}]{liang2023holistic}
Percy Liang, Rishi Bommasani, Tony Lee, Dimitris Tsipras, Dilara Soylu, Michihiro Yasunaga, Yian Zhang, Deepak Narayanan, Yuhuai Wu, Ananya Kumar, Benjamin Newman, Binhang Yuan, Bobby Yan, Ce~Zhang, Christian~Alexander Cosgrove, Christopher~D Manning, Christopher Re, Diana Acosta-Navas, Drew~Arad Hudson, Eric Zelikman, Esin Durmus, Faisal Ladhak, Frieda Rong, Hongyu Ren, Huaxiu Yao, Jue WANG, Keshav Santhanam, Laurel Orr, Lucia Zheng, Mert Yuksekgonul, Mirac Suzgun, Nathan Kim, Neel Guha, Niladri~S. Chatterji, Omar Khattab, Peter Henderson, Qian Huang, Ryan~Andrew Chi, Sang~Michael Xie, Shibani Santurkar, Surya Ganguli, Tatsunori Hashimoto, Thomas Icard, Tianyi Zhang, Vishrav Chaudhary, William Wang, Xuechen Li, Yifan Mai, Yuhui Zhang, and Yuta Koreeda. 2023.
\newblock \href {https://openreview.net/forum?id=iO4LZibEqW} {Holistic evaluation of language models}.
\newblock \emph{Transactions on Machine Learning Research}.
\newblock Featured Certification, Expert Certification.

\bibitem[{Lin et~al.(2022{\natexlab{a}})Lin, Mihaylov, Artetxe, Wang, Chen, Simig, Ott, Goyal, Bhosale, Du, Pasunuru, Shleifer, Koura, Chaudhary, O{'}Horo, Wang, Zettlemoyer, Kozareva, Diab, Stoyanov, and Li}]{lin2022shot}
Xi~Victoria Lin, Todor Mihaylov, Mikel Artetxe, Tianlu Wang, Shuohui Chen, Daniel Simig, Myle Ott, Naman Goyal, Shruti Bhosale, Jingfei Du, Ramakanth Pasunuru, Sam Shleifer, Punit~Singh Koura, Vishrav Chaudhary, Brian O{'}Horo, Jeff Wang, Luke Zettlemoyer, Zornitsa Kozareva, Mona Diab, Veselin Stoyanov, and Xian Li. 2022{\natexlab{a}}.
\newblock \href {https://doi.org/10.18653/v1/2022.emnlp-main.616} {Few-shot learning with multilingual generative language models}.
\newblock In \emph{Proceedings of the 2022 Conference on Empirical Methods in Natural Language Processing}, pages 9019--9052, Abu Dhabi, United Arab Emirates. Association for Computational Linguistics.

\bibitem[{Lin et~al.(2022{\natexlab{b}})Lin, Mihaylov, Artetxe, Wang, Chen, Simig, Ott, Goyal, Bhosale, Du, Pasunuru, Shleifer, Koura, Chaudhary, O'Horo, Wang, Zettlemoyer, Kozareva, Diab, Stoyanov, and Li}]{lin2022fewshot}
Xi~Victoria Lin, Todor Mihaylov, Mikel Artetxe, Tianlu Wang, Shuohui Chen, Daniel Simig, Myle Ott, Naman Goyal, Shruti Bhosale, Jingfei Du, Ramakanth Pasunuru, Sam Shleifer, Punit~Singh Koura, Vishrav Chaudhary, Brian O'Horo, Jeff Wang, Luke Zettlemoyer, Zornitsa Kozareva, Mona Diab, Veselin Stoyanov, and Xian Li. 2022{\natexlab{b}}.
\newblock \href {http://arxiv.org/abs/2112.10668} {Few-shot learning with multilingual language models}.

\bibitem[{Lovenia et~al.(2023)Lovenia, Dai, Cahyawijaya, Ji, and Fung}]{lovenia2023negative}
Holy Lovenia, Wenliang Dai, Samuel Cahyawijaya, Ziwei Ji, and Pascale Fung. 2023.
\newblock Negative object presence evaluation (nope) to measure object hallucination in vision-language models.
\newblock \emph{arXiv preprint arXiv:2310.05338}.

\bibitem[{Lovenia et~al.(2024)Lovenia, Mahendra, Akbar, Miranda, Santoso, Aco, Fadhilah, Mansurov, Imperial, Kampman, Moniz, Habibi, Hudi, Montalan, Ignatius, Lopo, Nixon, Karlsson, Jaya, Diandaru, Gao, Amadeus, Wang, Cruz, Whitehouse, Parmonangan, Khelli, Zhang, Susanto, Ryanda, Hermawan, Velasco, Kautsar, Hendria, Moslem, Flynn, Adilazuarda, Li, Lee, Damanhuri, Sun, Qorib, Djanibekov, Leong, Do, Muennighoff, Pansuwan, Putra, Xu, Tai, Purwarianti, Ruder, Tjhi, Limkonchotiwat, Aji, Keh, Winata, Zhang, Koto, Yong, and Cahyawijaya}]{lovenia2024seacrowd}
Holy Lovenia, Rahmad Mahendra, Salsabil~Maulana Akbar, Lester James~V. Miranda, Jennifer Santoso, Elyanah Aco, Akhdan Fadhilah, Jonibek Mansurov, Joseph~Marvin Imperial, Onno~P. Kampman, Joel Ruben~Antony Moniz, Muhammad Ravi~Shulthan Habibi, Frederikus Hudi, Railey Montalan, Ryan Ignatius, Joanito~Agili Lopo, William Nixon, Börje~F. Karlsson, James Jaya, Ryandito Diandaru, Yuze Gao, Patrick Amadeus, Bin Wang, Jan Christian~Blaise Cruz, Chenxi Whitehouse, Ivan~Halim Parmonangan, Maria Khelli, Wenyu Zhang, Lucky Susanto, Reynard~Adha Ryanda, Sonny~Lazuardi Hermawan, Dan~John Velasco, Muhammad Dehan~Al Kautsar, Willy~Fitra Hendria, Yasmin Moslem, Noah Flynn, Muhammad~Farid Adilazuarda, Haochen Li, Johanes Lee, R.~Damanhuri, Shuo Sun, Muhammad~Reza Qorib, Amirbek Djanibekov, Wei~Qi Leong, Quyet~V. Do, Niklas Muennighoff, Tanrada Pansuwan, Ilham~Firdausi Putra, Yan Xu, Ngee~Chia Tai, Ayu Purwarianti, Sebastian Ruder, William Tjhi, Peerat Limkonchotiwat, Alham~Fikri Aji, Sedrick Keh, Genta~Indra Winata, Ruochen
  Zhang, Fajri Koto, Zheng-Xin Yong, and Samuel Cahyawijaya. 2024.
\newblock \href {http://arxiv.org/abs/2406.10118} {Seacrowd: A multilingual multimodal data hub and benchmark suite for southeast asian languages}.

\bibitem[{Min et~al.(2022)Min, Lyu, Holtzman, Artetxe, Lewis, Hajishirzi, and Zettlemoyer}]{min2022rethinking}
Sewon Min, Xinxi Lyu, Ari Holtzman, Mikel Artetxe, Mike Lewis, Hannaneh Hajishirzi, and Luke Zettlemoyer. 2022.
\newblock \href {https://doi.org/10.18653/v1/2022.emnlp-main.759} {Rethinking the role of demonstrations: What makes in-context learning work?}
\newblock In \emph{Proceedings of the 2022 Conference on Empirical Methods in Natural Language Processing}, pages 11048--11064, Abu Dhabi, United Arab Emirates. Association for Computational Linguistics.

\bibitem[{Prettenhofer and Stein(2010)}]{prettenhofer-stein-2010-cross}
Peter Prettenhofer and Benno Stein. 2010.
\newblock \href {https://aclanthology.org/P10-1114} {Cross-language text classification using structural correspondence learning}.
\newblock In \emph{Proceedings of the 48th Annual Meeting of the Association for Computational Linguistics}, pages 1118--1127, Uppsala, Sweden. Association for Computational Linguistics.

\bibitem[{Rae et~al.(2022)Rae, Borgeaud, Cai, Millican, Hoffmann, Song, Aslanides, Henderson, Ring, Young, Rutherford, Hennigan, Menick, Cassirer, Powell, van~den Driessche, Hendricks, Rauh, Huang, Glaese, Welbl, Dathathri, Huang, Uesato, Mellor, Higgins, Creswell, McAleese, Wu, Elsen, Jayakumar, Buchatskaya, Budden, Sutherland, Simonyan, Paganini, Sifre, Martens, Li, Kuncoro, Nematzadeh, Gribovskaya, Donato, Lazaridou, Mensch, Lespiau, Tsimpoukelli, Grigorev, Fritz, Sottiaux, Pajarskas, Pohlen, Gong, Toyama, de~Masson~d'Autume, Li, Terzi, Mikulik, Babuschkin, Clark, de~Las~Casas, Guy, Jones, Bradbury, Johnson, Hechtman, Weidinger, Gabriel, Isaac, Lockhart, Osindero, Rimell, Dyer, Vinyals, Ayoub, Stanway, Bennett, Hassabis, Kavukcuoglu, and Irving}]{rae2022scaling}
Jack~W. Rae, Sebastian Borgeaud, Trevor Cai, Katie Millican, Jordan Hoffmann, Francis Song, John Aslanides, Sarah Henderson, Roman Ring, Susannah Young, Eliza Rutherford, Tom Hennigan, Jacob Menick, Albin Cassirer, Richard Powell, George van~den Driessche, Lisa~Anne Hendricks, Maribeth Rauh, Po-Sen Huang, Amelia Glaese, Johannes Welbl, Sumanth Dathathri, Saffron Huang, Jonathan Uesato, John Mellor, Irina Higgins, Antonia Creswell, Nat McAleese, Amy Wu, Erich Elsen, Siddhant Jayakumar, Elena Buchatskaya, David Budden, Esme Sutherland, Karen Simonyan, Michela Paganini, Laurent Sifre, Lena Martens, Xiang~Lorraine Li, Adhiguna Kuncoro, Aida Nematzadeh, Elena Gribovskaya, Domenic Donato, Angeliki Lazaridou, Arthur Mensch, Jean-Baptiste Lespiau, Maria Tsimpoukelli, Nikolai Grigorev, Doug Fritz, Thibault Sottiaux, Mantas Pajarskas, Toby Pohlen, Zhitao Gong, Daniel Toyama, Cyprien de~Masson~d'Autume, Yujia Li, Tayfun Terzi, Vladimir Mikulik, Igor Babuschkin, Aidan Clark, Diego de~Las~Casas, Aurelia Guy, Chris Jones,
  James Bradbury, Matthew Johnson, Blake Hechtman, Laura Weidinger, Iason Gabriel, William Isaac, Ed~Lockhart, Simon Osindero, Laura Rimell, Chris Dyer, Oriol Vinyals, Kareem Ayoub, Jeff Stanway, Lorrayne Bennett, Demis Hassabis, Koray Kavukcuoglu, and Geoffrey Irving. 2022.
\newblock \href {http://arxiv.org/abs/2112.11446} {Scaling language models: Methods, analysis \& insights from training gopher}.

\bibitem[{Reimers and Gurevych(2019)}]{reimers-2019-sentence-bert}
Nils Reimers and Iryna Gurevych. 2019.
\newblock \href {https://arxiv.org/abs/1908.10084} {Sentence-bert: Sentence embeddings using siamese bert-networks}.
\newblock In \emph{Proceedings of the 2019 Conference on Empirical Methods in Natural Language Processing}. Association for Computational Linguistics.

\bibitem[{Reimers and Gurevych(2020)}]{reimers-gurevych-2020-making}
Nils Reimers and Iryna Gurevych. 2020.
\newblock \href {https://doi.org/10.18653/v1/2020.emnlp-main.365} {Making monolingual sentence embeddings multilingual using knowledge distillation}.
\newblock In \emph{Proceedings of the 2020 Conference on Empirical Methods in Natural Language Processing (EMNLP)}, pages 4512--4525, Online. Association for Computational Linguistics.

\bibitem[{Rolnick et~al.(2019)Rolnick, Ahuja, Schwarz, Lillicrap, and Wayne}]{david2019experiencereplay}
David Rolnick, Arun Ahuja, Jonathan Schwarz, Timothy Lillicrap, and Gregory Wayne. 2019.
\newblock \href {https://proceedings.neurips.cc/paper_files/paper/2019/file/fa7cdfad1a5aaf8370ebeda47a1ff1c3-Paper.pdf} {Experience replay for continual learning}.
\newblock In \emph{Advances in Neural Information Processing Systems}, volume~32. Curran Associates, Inc.

\bibitem[{Scao et~al.(2022)Scao, Fan, Akiki, Pavlick, Ili{\'c}, Hesslow, Castagn{\'e}, Luccioni, Yvon, Gall{\'e} et~al.}]{scao2022bloom}
Teven~Le Scao, Angela Fan, Christopher Akiki, Ellie Pavlick, Suzana Ili{\'c}, Daniel Hesslow, Roman Castagn{\'e}, Alexandra~Sasha Luccioni, Fran{\c{c}}ois Yvon, Matthias Gall{\'e}, et~al. 2022.
\newblock Bloom: A 176b-parameter open-access multilingual language model.
\newblock \emph{arXiv preprint arXiv:2211.05100}.

\bibitem[{Schwenk et~al.(2021)Schwenk, Chaudhary, Sun, Gong, and Guzm{\'a}n}]{schwenk-etal-2021-wikimatrix}
Holger Schwenk, Vishrav Chaudhary, Shuo Sun, Hongyu Gong, and Francisco Guzm{\'a}n. 2021.
\newblock \href {https://doi.org/10.18653/v1/2021.eacl-main.115} {{W}iki{M}atrix: Mining 135{M} parallel sentences in 1620 language pairs from {W}ikipedia}.
\newblock In \emph{Proceedings of the 16th Conference of the European Chapter of the Association for Computational Linguistics: Main Volume}, pages 1351--1361, Online. Association for Computational Linguistics.

\bibitem[{Shi et~al.(2023)Shi, Suzgun, Freitag, Wang, Srivats, Vosoughi, Chung, Tay, Ruder, Zhou, Das, and Wei}]{shi2023language}
Freda Shi, Mirac Suzgun, Markus Freitag, Xuezhi Wang, Suraj Srivats, Soroush Vosoughi, Hyung~Won Chung, Yi~Tay, Sebastian Ruder, Denny Zhou, Dipanjan Das, and Jason Wei. 2023.
\newblock \href {https://openreview.net/forum?id=fR3wGCk-IXp} {Language models are multilingual chain-of-thought reasoners}.
\newblock In \emph{The Eleventh International Conference on Learning Representations}.

\bibitem[{Smith et~al.(2022)Smith, Patwary, Norick, LeGresley, Rajbhandari, Casper, Liu, Prabhumoye, Zerveas, Korthikanti, Zhang, Child, Aminabadi, Bernauer, Song, Shoeybi, He, Houston, Tiwary, and Catanzaro}]{smith2022megatron}
Shaden Smith, Mostofa Patwary, Brandon Norick, Patrick LeGresley, Samyam Rajbhandari, Jared Casper, Zhun Liu, Shrimai Prabhumoye, George Zerveas, Vijay Korthikanti, Elton Zhang, Rewon Child, Reza~Yazdani Aminabadi, Julie Bernauer, Xia Song, Mohammad Shoeybi, Yuxiong He, Michael Houston, Saurabh Tiwary, and Bryan Catanzaro. 2022.
\newblock \href {http://arxiv.org/abs/2201.11990} {Using deepspeed and megatron to train megatron-turing nlg 530b, a large-scale generative language model}.

\bibitem[{Srivastava et~al.(2023)Srivastava, Rastogi, Rao, Shoeb, Abid, Fisch, Brown, Santoro, Gupta, Garriga-Alonso, Kluska, Lewkowycz, Agarwal, Power, Ray, Warstadt, Kocurek, Safaya, Tazarv, Xiang, Parrish, Nie, Hussain, Askell, Dsouza, Slone, Rahane, Iyer, Andreassen, Madotto, Santilli, Stuhlm{\"u}ller, Dai, La, Lampinen, Zou, Jiang, Chen, Vuong, Gupta, Gottardi, Norelli, Venkatesh, Gholamidavoodi, Tabassum, Menezes, Kirubarajan, Mullokandov, Sabharwal, Herrick, Efrat, Erdem, Karaka{\c{s}}, Roberts, Loe, Zoph, Bojanowski, {\"O}zyurt, Hedayatnia, Neyshabur, Inden, Stein, Ekmekci, Lin, Howald, Orinion, Diao, Dour, Stinson, Argueta, Ferri, Singh, Rathkopf, Meng, Baral, Wu, Callison-Burch, Waites, Voigt, Manning, Potts, Ramirez, Rivera, Siro, Raffel, Ashcraft, Garbacea, Sileo, Garrette, Hendrycks, Kilman, Roth, Freeman, Khashabi, Levy, Gonz{\'a}lez, Perszyk, Hernandez, Chen, Ippolito, Gilboa, Dohan, Drakard, Jurgens, Datta, Ganguli, Emelin, Kleyko, Yuret, Chen, Tam, Hupkes, Misra, Buzan, Mollo, Yang, Lee,
  Schrader, Shutova, Cubuk, Segal, Hagerman, Barnes, Donoway, Pavlick, Rodol{\`a}, Lam, Chu, Tang, Erdem, Chang, Chi, Dyer, Jerzak, Kim, Manyasi, Zheltonozhskii, Xia, Siar, Mart{\'\i}nez-Plumed, Happ{\'e}, Chollet, Rong, Mishra, Winata, de~Melo, Kruszewski, Parascandolo, Mariani, Wang, Jaimovitch-Lopez, Betz, Gur-Ari, Galijasevic, Kim, Rashkin, Hajishirzi, Mehta, Bogar, Shevlin, Schuetze, Yakura, Zhang, Wong, Ng, Noble, Jumelet, Geissinger, Kernion, Hilton, Lee, Fisac, Simon, Koppel, Zheng, Zou, Kocon, Thompson, Wingfield, Kaplan, Radom, Sohl-Dickstein, Phang, Wei, Yosinski, Novikova, Bosscher, Marsh, Kim, Taal, Engel, Alabi, Xu, Song, Tang, Waweru, Burden, Miller, Balis, Batchelder, Berant, Frohberg, Rozen, Hernandez-Orallo, Boudeman, Guerr, Jones, Tenenbaum, Rule, Chua, Kanclerz, Livescu, Krauth, Gopalakrishnan, Ignatyeva, Markert, Dhole, Gimpel, Omondi, Mathewson, Chiafullo, Shkaruta, Shridhar, McDonell, Richardson, Reynolds, Gao, Zhang, Dugan, Qin, Contreras-Ochando, Morency, Moschella, Lam, Noble,
  Schmidt, He, Oliveros-Col{\'o}n, Metz, Senel, Bosma, Sap, Hoeve, Farooqi, Faruqui, Mazeika, Baturan, Marelli, Maru, Ramirez-Quintana, Tolkiehn, Giulianelli, Lewis, Potthast, Leavitt, Hagen, Schubert, Baitemirova, Arnaud, McElrath, Yee, Cohen, Gu, Ivanitskiy, Starritt, Strube, Sw{\k{e}}drowski, Bevilacqua, Yasunaga, Kale, Cain, Xu, Suzgun, Walker, Tiwari, Bansal, Aminnaseri, Geva, Gheini, T, Peng, Chi, Lee, Krakover, Cameron, Roberts, Doiron, Martinez, Nangia, Deckers, Muennighoff, Keskar, Iyer, Constant, Fiedel, Wen, Zhang, Agha, Elbaghdadi, Levy, Evans, Casares, Doshi, Fung, Liang, Vicol, Alipoormolabashi, Liao, Liang, Chang, Eckersley, Htut, Hwang, Mi{\l}kowski, Patil, Pezeshkpour, Oli, Mei, Lyu, Chen, Banjade, Rudolph, Gabriel, Habacker, Risco, Milli{\`e}re, Garg, Barnes, Saurous, Arakawa, Raymaekers, Frank, Sikand, Novak, Sitelew, Bras, Liu, Jacobs, Zhang, Salakhutdinov, Chi, Lee, Stovall, Teehan, Yang, Singh, Mohammad, Anand, Dillavou, Shleifer, Wiseman, Gruetter, Bowman, Schoenholz, Han, Kwatra, Rous,
  Ghazarian, Ghosh, Casey, Bischoff, Gehrmann, Schuster, Sadeghi, Hamdan, Zhou, Srivastava, Shi, Singh, Asaadi, Gu, Pachchigar, Toshniwal, Upadhyay, Debnath, Shakeri, Thormeyer, Melzi, Reddy, Makini, Lee, Torene, Hatwar, Dehaene, Divic, Ermon, Biderman, Lin, Prasad, Piantadosi, Shieber, Misherghi, Kiritchenko, Mishra, Linzen, Schuster, Li, Yu, Ali, Hashimoto, Wu, Desbordes, Rothschild, Phan, Wang, Nkinyili, Schick, Kornev, Tunduny, Gerstenberg, Chang, Neeraj, Khot, Shultz, Shaham, Misra, Demberg, Nyamai, Raunak, Ramasesh, vinay~uday prabhu, Padmakumar, Srikumar, Fedus, Saunders, Zhang, Vossen, Ren, Tong, Zhao, Wu, Shen, Yaghoobzadeh, Lakretz, Song, Bahri, Choi, Yang, Hao, Chen, Belinkov, Hou, Hou, Bai, Seid, Zhao, Wang, Wang, Wang, and Wu}]{srivastava2023beyond}
Aarohi Srivastava, Abhinav Rastogi, Abhishek Rao, Abu Awal~Md Shoeb, Abubakar Abid, Adam Fisch, Adam~R. Brown, Adam Santoro, Aditya Gupta, Adri{\`a} Garriga-Alonso, Agnieszka Kluska, Aitor Lewkowycz, Akshat Agarwal, Alethea Power, Alex Ray, Alex Warstadt, Alexander~W. Kocurek, Ali Safaya, Ali Tazarv, Alice Xiang, Alicia Parrish, Allen Nie, Aman Hussain, Amanda Askell, Amanda Dsouza, Ambrose Slone, Ameet Rahane, Anantharaman~S. Iyer, Anders~Johan Andreassen, Andrea Madotto, Andrea Santilli, Andreas Stuhlm{\"u}ller, Andrew~M. Dai, Andrew La, Andrew Lampinen, Andy Zou, Angela Jiang, Angelica Chen, Anh Vuong, Animesh Gupta, Anna Gottardi, Antonio Norelli, Anu Venkatesh, Arash Gholamidavoodi, Arfa Tabassum, Arul Menezes, Arun Kirubarajan, Asher Mullokandov, Ashish Sabharwal, Austin Herrick, Avia Efrat, Aykut Erdem, Ayla Karaka{\c{s}}, B.~Ryan Roberts, Bao~Sheng Loe, Barret Zoph, Bart{\l}omiej Bojanowski, Batuhan {\"O}zyurt, Behnam Hedayatnia, Behnam Neyshabur, Benjamin Inden, Benno Stein, Berk Ekmekci, Bill~Yuchen
  Lin, Blake Howald, Bryan Orinion, Cameron Diao, Cameron Dour, Catherine Stinson, Cedrick Argueta, Cesar Ferri, Chandan Singh, Charles Rathkopf, Chenlin Meng, Chitta Baral, Chiyu Wu, Chris Callison-Burch, Christopher Waites, Christian Voigt, Christopher~D Manning, Christopher Potts, Cindy Ramirez, Clara~E. Rivera, Clemencia Siro, Colin Raffel, Courtney Ashcraft, Cristina Garbacea, Damien Sileo, Dan Garrette, Dan Hendrycks, Dan Kilman, Dan Roth, C.~Daniel Freeman, Daniel Khashabi, Daniel Levy, Daniel~Mosegu{\'\i} Gonz{\'a}lez, Danielle Perszyk, Danny Hernandez, Danqi Chen, Daphne Ippolito, Dar Gilboa, David Dohan, David Drakard, David Jurgens, Debajyoti Datta, Deep Ganguli, Denis Emelin, Denis Kleyko, Deniz Yuret, Derek Chen, Derek Tam, Dieuwke Hupkes, Diganta Misra, Dilyar Buzan, Dimitri~Coelho Mollo, Diyi Yang, Dong-Ho Lee, Dylan Schrader, Ekaterina Shutova, Ekin~Dogus Cubuk, Elad Segal, Eleanor Hagerman, Elizabeth Barnes, Elizabeth Donoway, Ellie Pavlick, Emanuele Rodol{\`a}, Emma Lam, Eric Chu, Eric Tang,
  Erkut Erdem, Ernie Chang, Ethan~A Chi, Ethan Dyer, Ethan Jerzak, Ethan Kim, Eunice~Engefu Manyasi, Evgenii Zheltonozhskii, Fanyue Xia, Fatemeh Siar, Fernando Mart{\'\i}nez-Plumed, Francesca Happ{\'e}, Francois Chollet, Frieda Rong, Gaurav Mishra, Genta~Indra Winata, Gerard de~Melo, Germ{\'a}n Kruszewski, Giambattista Parascandolo, Giorgio Mariani, Gloria~Xinyue Wang, Gonzalo Jaimovitch-Lopez, Gregor Betz, Guy Gur-Ari, Hana Galijasevic, Hannah Kim, Hannah Rashkin, Hannaneh Hajishirzi, Harsh Mehta, Hayden Bogar, Henry Francis~Anthony Shevlin, Hinrich Schuetze, Hiromu Yakura, Hongming Zhang, Hugh~Mee Wong, Ian Ng, Isaac Noble, Jaap Jumelet, Jack Geissinger, Jackson Kernion, Jacob Hilton, Jaehoon Lee, Jaime~Fern{\'a}ndez Fisac, James~B Simon, James Koppel, James Zheng, James Zou, Jan Kocon, Jana Thompson, Janelle Wingfield, Jared Kaplan, Jarema Radom, Jascha Sohl-Dickstein, Jason Phang, Jason Wei, Jason Yosinski, Jekaterina Novikova, Jelle Bosscher, Jennifer Marsh, Jeremy Kim, Jeroen Taal, Jesse Engel, Jesujoba
  Alabi, Jiacheng Xu, Jiaming Song, Jillian Tang, Joan Waweru, John Burden, John Miller, John~U. Balis, Jonathan Batchelder, Jonathan Berant, J{\"o}rg Frohberg, Jos Rozen, Jose Hernandez-Orallo, Joseph Boudeman, Joseph Guerr, Joseph Jones, Joshua~B. Tenenbaum, Joshua~S. Rule, Joyce Chua, Kamil Kanclerz, Karen Livescu, Karl Krauth, Karthik Gopalakrishnan, Katerina Ignatyeva, Katja Markert, Kaustubh Dhole, Kevin Gimpel, Kevin Omondi, Kory~Wallace Mathewson, Kristen Chiafullo, Ksenia Shkaruta, Kumar Shridhar, Kyle McDonell, Kyle Richardson, Laria Reynolds, Leo Gao, Li~Zhang, Liam Dugan, Lianhui Qin, Lidia Contreras-Ochando, Louis-Philippe Morency, Luca Moschella, Lucas Lam, Lucy Noble, Ludwig Schmidt, Luheng He, Luis Oliveros-Col{\'o}n, Luke Metz, L{\"u}tfi~Kerem Senel, Maarten Bosma, Maarten Sap, Maartje~Ter Hoeve, Maheen Farooqi, Manaal Faruqui, Mantas Mazeika, Marco Baturan, Marco Marelli, Marco Maru, Maria~Jose Ramirez-Quintana, Marie Tolkiehn, Mario Giulianelli, Martha Lewis, Martin Potthast, Matthew~L
  Leavitt, Matthias Hagen, M{\'a}ty{\'a}s Schubert, Medina~Orduna Baitemirova, Melody Arnaud, Melvin McElrath, Michael~Andrew Yee, Michael Cohen, Michael Gu, Michael Ivanitskiy, Michael Starritt, Michael Strube, Micha{\l} Sw{\k{e}}drowski, Michele Bevilacqua, Michihiro Yasunaga, Mihir Kale, Mike Cain, Mimee Xu, Mirac Suzgun, Mitch Walker, Mo~Tiwari, Mohit Bansal, Moin Aminnaseri, Mor Geva, Mozhdeh Gheini, Mukund~Varma T, Nanyun Peng, Nathan~Andrew Chi, Nayeon Lee, Neta Gur-Ari Krakover, Nicholas Cameron, Nicholas Roberts, Nick Doiron, Nicole Martinez, Nikita Nangia, Niklas Deckers, Niklas Muennighoff, Nitish~Shirish Keskar, Niveditha~S. Iyer, Noah Constant, Noah Fiedel, Nuan Wen, Oliver Zhang, Omar Agha, Omar Elbaghdadi, Omer Levy, Owain Evans, Pablo Antonio~Moreno Casares, Parth Doshi, Pascale Fung, Paul~Pu Liang, Paul Vicol, Pegah Alipoormolabashi, Peiyuan Liao, Percy Liang, Peter~W Chang, Peter Eckersley, Phu~Mon Htut, Pinyu Hwang, Piotr Mi{\l}kowski, Piyush Patil, Pouya Pezeshkpour, Priti Oli, Qiaozhu
  Mei, Qing Lyu, Qinlang Chen, Rabin Banjade, Rachel~Etta Rudolph, Raefer Gabriel, Rahel Habacker, Ramon Risco, Rapha{\"e}l Milli{\`e}re, Rhythm Garg, Richard Barnes, Rif~A. Saurous, Riku Arakawa, Robbe Raymaekers, Robert Frank, Rohan Sikand, Roman Novak, Roman Sitelew, Ronan~Le Bras, Rosanne Liu, Rowan Jacobs, Rui Zhang, Russ Salakhutdinov, Ryan~Andrew Chi, Seungjae~Ryan Lee, Ryan Stovall, Ryan Teehan, Rylan Yang, Sahib Singh, Saif~M. Mohammad, Sajant Anand, Sam Dillavou, Sam Shleifer, Sam Wiseman, Samuel Gruetter, Samuel~R. Bowman, Samuel~Stern Schoenholz, Sanghyun Han, Sanjeev Kwatra, Sarah~A. Rous, Sarik Ghazarian, Sayan Ghosh, Sean Casey, Sebastian Bischoff, Sebastian Gehrmann, Sebastian Schuster, Sepideh Sadeghi, Shadi Hamdan, Sharon Zhou, Shashank Srivastava, Sherry Shi, Shikhar Singh, Shima Asaadi, Shixiang~Shane Gu, Shubh Pachchigar, Shubham Toshniwal, Shyam Upadhyay, Shyamolima~Shammie Debnath, Siamak Shakeri, Simon Thormeyer, Simone Melzi, Siva Reddy, Sneha~Priscilla Makini, Soo-Hwan Lee, Spencer
  Torene, Sriharsha Hatwar, Stanislas Dehaene, Stefan Divic, Stefano Ermon, Stella Biderman, Stephanie Lin, Stephen Prasad, Steven Piantadosi, Stuart Shieber, Summer Misherghi, Svetlana Kiritchenko, Swaroop Mishra, Tal Linzen, Tal Schuster, Tao Li, Tao Yu, Tariq Ali, Tatsunori Hashimoto, Te-Lin Wu, Th{\'e}o Desbordes, Theodore Rothschild, Thomas Phan, Tianle Wang, Tiberius Nkinyili, Timo Schick, Timofei Kornev, Titus Tunduny, Tobias Gerstenberg, Trenton Chang, Trishala Neeraj, Tushar Khot, Tyler Shultz, Uri Shaham, Vedant Misra, Vera Demberg, Victoria Nyamai, Vikas Raunak, Vinay~Venkatesh Ramasesh, vinay~uday prabhu, Vishakh Padmakumar, Vivek Srikumar, William Fedus, William Saunders, William Zhang, Wout Vossen, Xiang Ren, Xiaoyu Tong, Xinran Zhao, Xinyi Wu, Xudong Shen, Yadollah Yaghoobzadeh, Yair Lakretz, Yangqiu Song, Yasaman Bahri, Yejin Choi, Yichi Yang, Yiding Hao, Yifu Chen, Yonatan Belinkov, Yu~Hou, Yufang Hou, Yuntao Bai, Zachary Seid, Zhuoye Zhao, Zijian Wang, Zijie~J. Wang, Zirui Wang, and Ziyi Wu.
  2023.
\newblock \href {https://openreview.net/forum?id=uyTL5Bvosj} {Beyond the imitation game: Quantifying and extrapolating the capabilities of language models}.
\newblock \emph{Transactions on Machine Learning Research}.

\bibitem[{Tanwar et~al.(2023)Tanwar, Dutta, Borthakur, and Chakraborty}]{tanwar2023multilingual}
Eshaan Tanwar, Subhabrata Dutta, Manish Borthakur, and Tanmoy Chakraborty. 2023.
\newblock \href {https://doi.org/10.18653/v1/2023.acl-long.346} {Multilingual {LLM}s are better cross-lingual in-context learners with alignment}.
\newblock In \emph{Proceedings of the 61st Annual Meeting of the Association for Computational Linguistics (Volume 1: Long Papers)}, pages 6292--6307, Toronto, Canada. Association for Computational Linguistics.

\bibitem[{Team(2023)}]{MosaicML2023Introducing}
MosaicML~NLP Team. 2023.
\newblock \href {www.mosaicml.com/blog/mpt-7b} {Introducing mpt-7b: A new standard for open-source, commercially usable llms}.
\newblock Accessed: 2023-05-05.

\bibitem[{Team et~al.(2022)Team, Costa-jussà, Cross, Çelebi, Elbayad, Heafield, Heffernan, Kalbassi, Lam, Licht, Maillard, Sun, Wang, Wenzek, Youngblood, Akula, Barrault, Gonzalez, Hansanti, Hoffman, Jarrett, Sadagopan, Rowe, Spruit, Tran, Andrews, Ayan, Bhosale, Edunov, Fan, Gao, Goswami, Guzmán, Koehn, Mourachko, Ropers, Saleem, Schwenk, and Wang}]{nllb2022nllb}
NLLB Team, Marta~R. Costa-jussà, James Cross, Onur Çelebi, Maha Elbayad, Kenneth Heafield, Kevin Heffernan, Elahe Kalbassi, Janice Lam, Daniel Licht, Jean Maillard, Anna Sun, Skyler Wang, Guillaume Wenzek, Al~Youngblood, Bapi Akula, Loic Barrault, Gabriel~Mejia Gonzalez, Prangthip Hansanti, John Hoffman, Semarley Jarrett, Kaushik~Ram Sadagopan, Dirk Rowe, Shannon Spruit, Chau Tran, Pierre Andrews, Necip~Fazil Ayan, Shruti Bhosale, Sergey Edunov, Angela Fan, Cynthia Gao, Vedanuj Goswami, Francisco Guzmán, Philipp Koehn, Alexandre Mourachko, Christophe Ropers, Safiyyah Saleem, Holger Schwenk, and Jeff Wang. 2022.
\newblock No language left behind: Scaling human-centered machine translation.

\bibitem[{Wei et~al.(2022)Wei, Wang, Schuurmans, Bosma, brian ichter, Xia, Chi, Le, and Zhou}]{wei2022chain}
Jason Wei, Xuezhi Wang, Dale Schuurmans, Maarten Bosma, brian ichter, Fei Xia, Ed~H. Chi, Quoc~V Le, and Denny Zhou. 2022.
\newblock \href {https://openreview.net/forum?id=_VjQlMeSB_J} {Chain of thought prompting elicits reasoning in large language models}.
\newblock In \emph{Advances in Neural Information Processing Systems}.

\bibitem[{Wibowo et~al.(2023)Wibowo, Fuadi, Nityasya, Prasojo, and Aji}]{wibowo2023copalid}
Haryo~Akbarianto Wibowo, Erland~Hilman Fuadi, Made~Nindyatama Nityasya, Radityo~Eko Prasojo, and Alham~Fikri Aji. 2023.
\newblock \href {http://arxiv.org/abs/2311.01012} {Copal-id: Indonesian language reasoning with local culture and nuances}.

\bibitem[{Wilie et~al.(2020)Wilie, Vincentio, Winata, Cahyawijaya, Li, Lim, Soleman, Mahendra, Fung, Bahar, and Purwarianti}]{wilie2020indonlu}
Bryan Wilie, Karissa Vincentio, Genta~Indra Winata, Samuel Cahyawijaya, Xiaohong Li, Zhi~Yuan Lim, Sidik Soleman, Rahmad Mahendra, Pascale Fung, Syafri Bahar, and Ayu Purwarianti. 2020.
\newblock \href {https://aclanthology.org/2020.aacl-main.85} {{I}ndo{NLU}: Benchmark and resources for evaluating {I}ndonesian natural language understanding}.
\newblock In \emph{Proceedings of the 1st Conference of the Asia-Pacific Chapter of the Association for Computational Linguistics and the 10th International Joint Conference on Natural Language Processing}, pages 843--857, Suzhou, China. Association for Computational Linguistics.

\bibitem[{Winata et~al.(2022{\natexlab{a}})Winata, Wu, Kulkarni, Solorio, and Preotiuc-Pietro}]{winata-etal-2022-cross}
Genta Winata, Shijie Wu, Mayank Kulkarni, Thamar Solorio, and Daniel Preotiuc-Pietro. 2022{\natexlab{a}}.
\newblock \href {https://aclanthology.org/2022.aacl-main.59} {Cross-lingual few-shot learning on unseen languages}.
\newblock In \emph{Proceedings of the 2nd Conference of the Asia-Pacific Chapter of the Association for Computational Linguistics and the 12th International Joint Conference on Natural Language Processing (Volume 1: Long Papers)}, pages 777--791, Online only. Association for Computational Linguistics.

\bibitem[{Winata et~al.(2022{\natexlab{b}})Winata, Aji, Cahyawijaya, Mahendra, Koto, Romadhony, Kurniawan, Moeljadi, Prasojo, Fung et~al.}]{winata2022nusax}
Genta~Indra Winata, Alham~Fikri Aji, Samuel Cahyawijaya, Rahmad Mahendra, Fajri Koto, Ade Romadhony, Kemal Kurniawan, David Moeljadi, Radityo~Eko Prasojo, Pascale Fung, et~al. 2022{\natexlab{b}}.
\newblock Nusax: Multilingual parallel sentiment dataset for 10 indonesian local languages.
\newblock \emph{arXiv preprint arXiv:2205.15960}.

\bibitem[{Winata et~al.(2021{\natexlab{a}})Winata, Cahyawijaya, Liu, Lin, Madotto, and Fung}]{winata-etal-2021-multilingual}
Genta~Indra Winata, Samuel Cahyawijaya, Zihan Liu, Zhaojiang Lin, Andrea Madotto, and Pascale Fung. 2021{\natexlab{a}}.
\newblock \href {https://doi.org/10.18653/v1/2021.calcs-1.20} {Are multilingual models effective in code-switching?}
\newblock In \emph{Proceedings of the Fifth Workshop on Computational Approaches to Linguistic Code-Switching}, pages 142--153, Online. Association for Computational Linguistics.

\bibitem[{Winata et~al.(2021{\natexlab{b}})Winata, Madotto, Lin, Liu, Yosinski, and Fung}]{winata2021language}
Genta~Indra Winata, Andrea Madotto, Zhaojiang Lin, Rosanne Liu, Jason Yosinski, and Pascale Fung. 2021{\natexlab{b}}.
\newblock Language models are few-shot multilingual learners.
\newblock In \emph{Proceedings of the 1st Workshop on Multilingual Representation Learning}, pages 1--15.

\bibitem[{Winata et~al.(2021{\natexlab{c}})Winata, Madotto, Lin, Liu, Yosinski, and Fung}]{winata-etal-2021-language}
Genta~Indra Winata, Andrea Madotto, Zhaojiang Lin, Rosanne Liu, Jason Yosinski, and Pascale Fung. 2021{\natexlab{c}}.
\newblock \href {https://doi.org/10.18653/v1/2021.mrl-1.1} {Language models are few-shot multilingual learners}.
\newblock In \emph{Proceedings of the 1st Workshop on Multilingual Representation Learning}, pages 1--15, Punta Cana, Dominican Republic. Association for Computational Linguistics.

\bibitem[{Xie et~al.(2022)Xie, Raghunathan, Liang, and Ma}]{xie2022icl}
Sang~Michael Xie, Aditi Raghunathan, Percy Liang, and Tengyu Ma. 2022.
\newblock \href {https://openreview.net/forum?id=RdJVFCHjUMI} {An explanation of in-context learning as implicit bayesian inference}.
\newblock In \emph{International Conference on Learning Representations}.

\bibitem[{Yong et~al.(2022)Yong, Schoelkopf, Muennighoff, Aji, Adelani, Almubarak, Bari, Sutawika, Kasai, Baruwa, Winata, Biderman, Radev, and Nikoulina}]{yong2022bloom1}
Zheng-Xin Yong, Hailey Schoelkopf, Niklas Muennighoff, Alham~Fikri Aji, David~Ifeoluwa Adelani, Khalid Almubarak, M~Saiful Bari, Lintang Sutawika, Jungo Kasai, Ahmed Baruwa, Genta~Indra Winata, Stella Biderman, Dragomir Radev, and Vassilina Nikoulina. 2022.
\newblock \href {http://arxiv.org/abs/2212.09535} {Bloom+1: Adding language support to bloom for zero-shot prompting}.

\bibitem[{Yong et~al.(2023)Yong, Zhang, Zosa~Forde, Wang, Cahyawijaya, Lovenia, Indra~Winata, Sutawika, Blaise~Cruz, Phan et~al.}]{yong2023prompting}
Zheng-Xin Yong, Ruochen Zhang, Jessica Zosa~Forde, Skyler Wang, Samuel Cahyawijaya, Holy Lovenia, Genta Indra~Winata, Lintang Sutawika, Jan~Christian Blaise~Cruz, Long Phan, et~al. 2023.
\newblock Prompting multilingual large language models to generate code-mixed texts: The case of south east asian languages.
\newblock \emph{arXiv preprint arXiv:2303.13592}.

\bibitem[{Zhang et~al.(2023)Zhang, Cahyawijaya, Cruz, and Aji}]{zhang2023multilingual}
Ruochen Zhang, Samuel Cahyawijaya, Jan Christian~Blaise Cruz, and Alham~Fikri Aji. 2023.
\newblock \href {http://arxiv.org/abs/2305.14235} {Multilingual large language models are not (yet) code-switchers}.

\end{thebibliography}
\bibliographystyle{acl_natbib}

\appendix

\onecolumn

\newpage 

\section{Languages Under Study}
\label{app:languages}

We conduct experiments on 32 languages, with 25 low-resource languages and 7 relatively high-resource languages. We provide the detailed list of all the languages under study in Table~\ref{tab:all-languages}.

\begin{table*}[!h]
    \centering
    \resizebox{\linewidth}{!}{
        \begin{tabular}{c|c|c|c|c|c|c|c}
            \toprule
            Language & Language & Dataset & Test & Geographic & Language & \%BLOOM & \%XGLM \\
            Code & Name & Name & Size & Region & Family & Pretraining & Pretraining \\
            \midrule
            btk & Batak & NusaTranslation & 1200 & South-East Asia & Austronesian & - & - \\
            sun & Sundanese & NusaTranslation & 1200 & South-East Asia & Austronesian & - & - \\
            jav & Javanese & NusaTranslation & 1200 & South-East Asia & Austronesian & - & - \\
            mad & Madurese & NusaTranslation & 1200 & South-East Asia & Austronesian & - & - \\
            mak & Makassarese & NusaTranslation & 1200 & South-East Asia & Austronesian & - & - \\
            min & Minangkabau & NusaTranslation & 1200 & South-East Asia & Austronesian & - & - \\
            \midrule
            amh & Amharic & MasakhaNews & 376 & Africa & Afro-Asiatic & - & - \\
            hau & Hausa & MasakhaNews & 637 & Africa & Afro-Asiatic & - & - \\
            ibo & Igbo & MasakhaNews & 390 & Africa & Niger-Congo & 0.00\% & - \\
            lug & Luganda & MasakhaNews & 223 & Africa & Niger-Congo & 0.00\% & - \\            
            pcm & Nigerian Pidgin & MasakhaNews & 305 & Africa & English Creole & - & - \\
            sna & chiShona & MasakhaNews & 369 & Africa & Niger-Congo & - & - \\
            swa & Kiswahili & MasakhaNews & 476 & Africa & Niger-Congo & 0.01\% & 0.25\% \\
            xho & isiXhosa & MasakhaNews & 297 & Africa & Niger-Congo & 0.00\% & - \\
            yor & Yorùbá & MasakhaNews & 411 & Africa & Niger-Congo & 0.01\% & - \\
            \midrule
            aym & Aymara & AmericasNLI & 750 & South America & Aymaran & - & - \\
            bzd & Bribri & AmericasNLI & 750 & South America & Chibchan & - & - \\
            cni & Asháninka & AmericasNLI & 750 & South America & Arawak & - & - \\
            grn & Guaraní & AmericasNLI & 750 & South America & Tupian & - & - \\
            hch & Wixarika & AmericasNLI & 750 & South America & Uto-Aztecan & - & - \\
            nah & Nahuatl & AmericasNLI & 738 & South America & Uto-Aztecan & - & - \\
            oto & Otomí & AmericasNLI & 748 & South America & Oto-Manguean & - & - \\
            quy & Quechua & AmericasNLI & 750 & South America & Quechuan & - & 0.01\% \\
            shp & Shipibo-Konibo & AmericasNLI & 750 & South America & Pano-Tacanan & - & - \\
            tar & Rarámuri & AmericasNLI & 750 & South America & Uto-Aztecan & - & - \\
            \midrule
            arb & Arabic & TweetSentimentMultilingual & 870 & Northern Africa & Afro-Asiatic & 4.64\% & 0.75\% \\
            fra & French & TweetSentimentMultilingual & 870 & Europe & Indo-European & 12.90\% & 3.00\% \\
            deu & German & TweetSentimentMultilingual & 870 & Europe & Indo-European & - & 3.50\% \\
            hin & Hindi & TweetSentimentMultilingual & 870 & Central Asia & Indo-European & 1.53\% & 1.00\% \\
            ita & Italian & TweetSentimentMultilingual & 870 & Europe & Indo-European & - & 1.50\% \\
            por & Portuguese & TweetSentimentMultilingual & 870 & Europe & Indo-European & 4.91\% & 2.25\% \\
            spa & Spanish & TweetSentimentMultilingual & 870 & Europe & Indo-European & 10.85\% & 3.25\% \\
            \bottomrule
        \end{tabular}
    }
    \caption{List of languages under study. "-" denotes the language is not on the pre-training dataset, while 0.00\% denotes a very small percentage (<0.01\%) of the pre-training data is in that language.}
    \label{tab:all-languages}
\end{table*}

\clearpage

\section{Alignment Prompt}
\label{app:alignment}

\begin{figure}[!h]
    \centering
    \resizebox{\linewidth}{!}{    
    \includegraphics{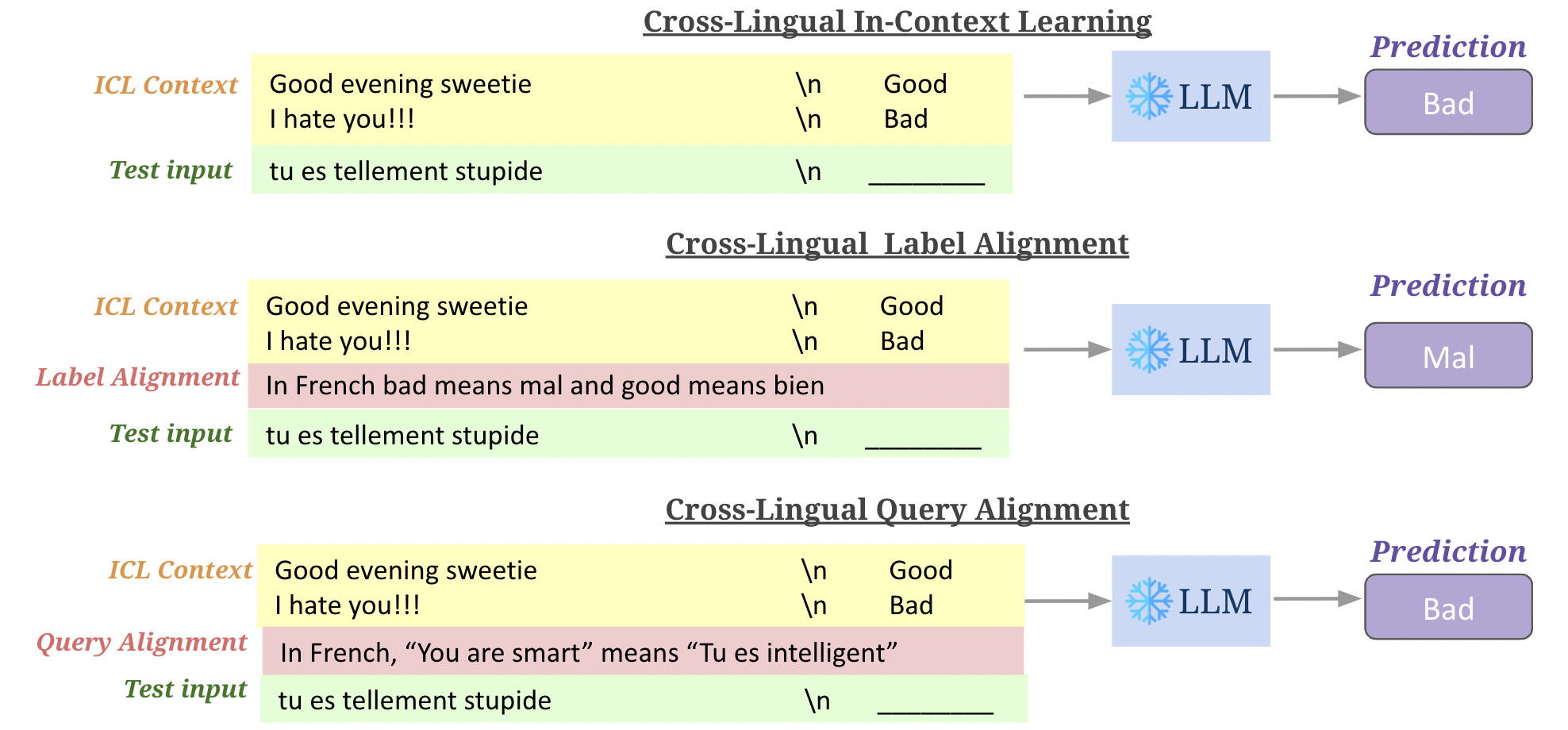}
    }
    \caption{Example prompt for in-context label alignment and in-context query alignment.}
    \label{fig:alignment}
\end{figure}

We showcase the example prompt for cross-lingual in-context learning, in-context label alignment, and in-context query alignment in Figure~\ref{fig:alignment}.

\section{Analysis on Cross-lingual Semantic Similarity}
\label{app:xss}

\begin{figure*}[!h]
    \centering
    \begin{minipage}{.48\linewidth}
        \centering
        \begingroup
        \includegraphics[width=0.8\linewidth]{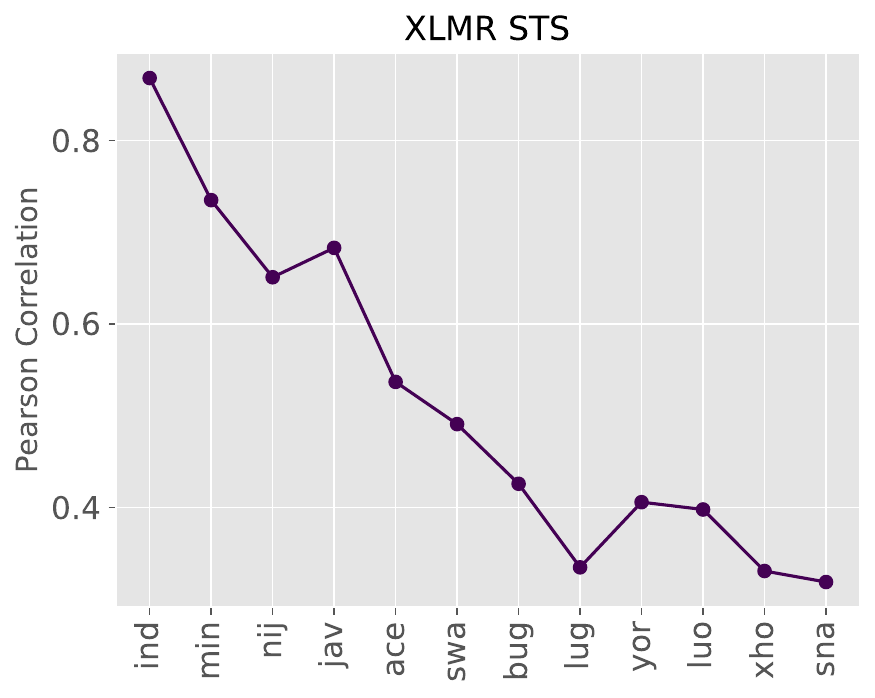}
        \endgroup
    \end{minipage}%
    \hspace{2pt}
    % \hfill\vline\hfill
    \begin{minipage}{.48\linewidth}
        \centering
        \begingroup
        \includegraphics[width=0.81\linewidth]{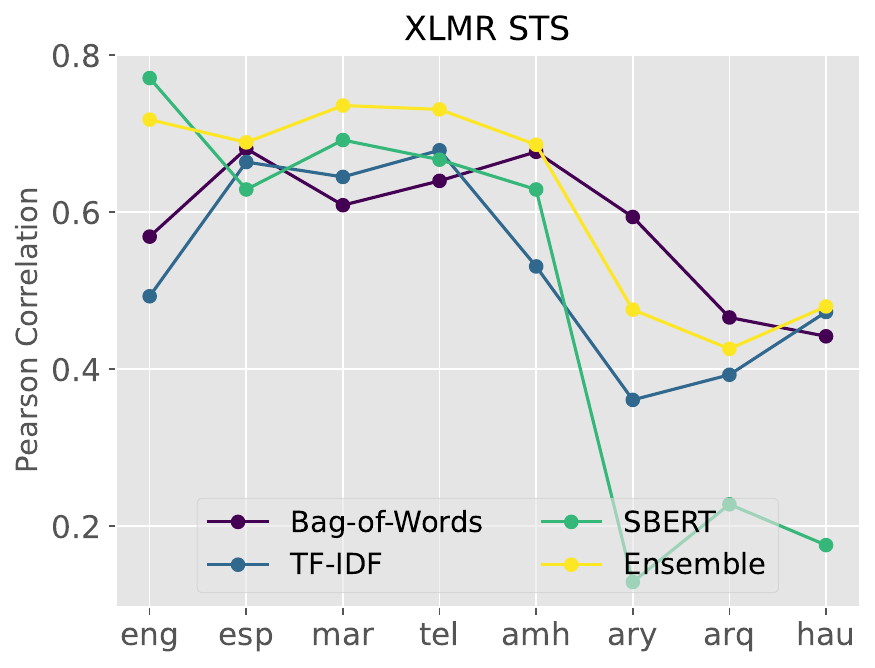}
        \endgroup
    \end{minipage}%
    \caption{\textbf{(top)} Correlation of cross-lingual similarity with the correct label for the XLMR STS model. \textbf{(bottom)} Correlation of monolingual similarity with the correct label for the XLMR STS model.}
    \label{fig:xss-mss}
\end{figure*}

We showcase that the semantic representation for these languages might not be well aligned with the high-resource languages. We construct sentence similarity dataset covering 26 languages by utilizing the translation samples from two machine translation datasets, i.e., MAFAND~\cite{adelani-etal-2022-thousand} and NusaX-MT~\cite{winata2022nusax}. We create a balanced dataset with 50\% positive pairs and 50\% negative pairs over all 26 languages. We measure the cross-lingual semantic similarity performance of using Sentence Transformers~\cite{reimers-2019-sentence-bert}. We also conduct monolingual semantic similarity analysis for various languages using the data from SemEval 2024 Task 12: Textual Semantic Relatedness dataset~\footnote{\url{https://github.com/semantic-textual-relatedness/Semantic_Relatedness_SemEval2024}}. For the monolingual semantic similarity we add additional word frequency features including bag-of-words and TF-IDF to improve the retrieval quality of the semantic similarity model.

\begin{figure*}[!h]
    \centering
    \resizebox{0.95\linewidth}{!}{
     \includegraphics[width=\textwidth]{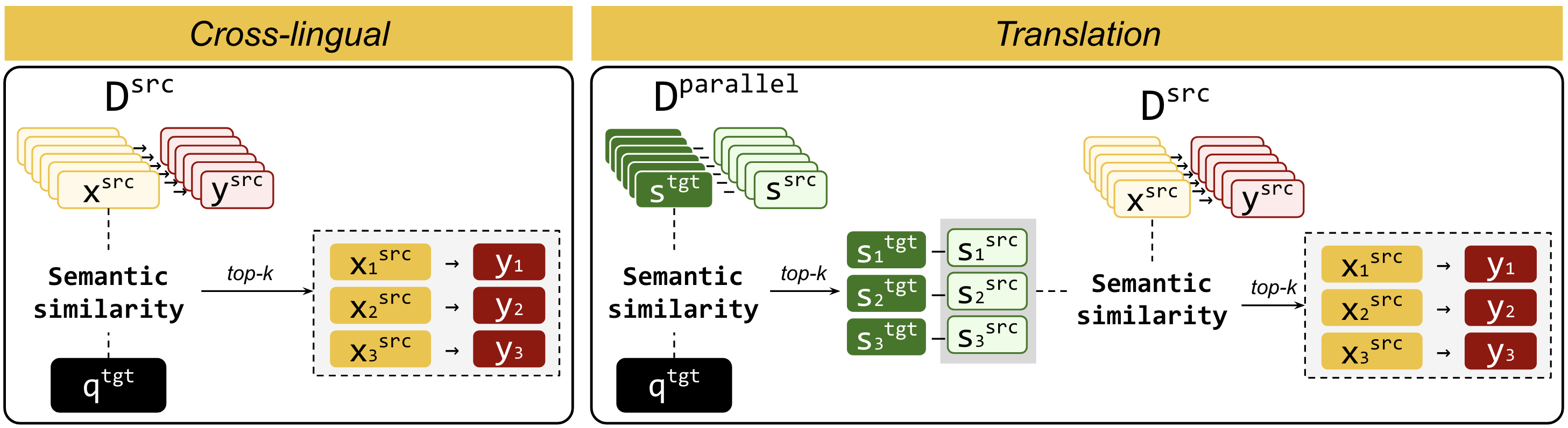}}
    \caption{We explore two semantic similarity methods for cross-lingual exemplar retrieval in X-ICL, i.e., cross-lingual semantic similarity and translation semantic similarity (T-ICL).}
    \label{fig:semantic-similarity}
\end{figure*}

As shown in Figure~\ref{fig:xss-mss}, both monolingual and cross-lingual semantic similarity on more low-resource languages are generally yield a much lower correlation which signifies the limitation of the sentence embedding model to represent the sentences on these languages. Nevertheless, for monolingual semantic similarity, it is possible to improve the similarity on these low-resource languages with minimal trade off on the other language by employing character/word frequency features to support the semantic similarity model. With that in mind, we explore an alternative approach for cross-lingual retrieval by using monolingual semantic similarity and an external parallel corpus. We called this semantic similarity method as \textbf{translation semantic similarity}. The comparison of cross-lingual retrieval using cross-lingual semantic similarity and translation semantic similarity is shown in Figure~\ref{fig:semantic-similarity}.

\begin{table*}[!h]
    \centering
    \resizebox{\linewidth}{!}{
        \begin{tabular}{c|c|c|c|c|c|c|c}
            \toprule
            Language & \multicolumn{7}{c}{Label Set} \\
            \midrule
            eng & business & entertainment & health & politics & religion & sports & technology \\
            % amh & ንግድ & መዝናኛ & ትምህርተ፡ጤና & ፖለቲካ & ሃይማኖት & ስፖርት & ቴክኖዎሎጂ \\
            hau & kasuwanci & nishadi & lafiya & siyasa & addini & wasanni & fasaha \\
            ibo & azumahia & nturundu & ahuike & ndoro ndoro ochichi & okpukpere chi & egwuregwu & teknuzu \\
            lug & bizinensi & okwesanyusa & obulamu & ebyobufuzi & eddiini & ebyemizannyo & tekinolojiya \\
            pcm & business & entertainment & health & politics & religion & sports & technology \\
            sna & business & varaidzo & utano & zvematongerwo enyika & chitendero & mitambo & teknolojia \\
            swa & biashara & burudani & afya & siasa & dini & michezo & teknolojia \\
            xho & ishishini & ukuzonwabisa & impilo & kwezopolitiko & unqulo & ezemidlalo & iteknoloji \\
            yor & iṣowo & Idanilaraya & ilera & oselu & esin & idaraya & ona ero \\
            \bottomrule
        \end{tabular}
    }
    \caption{Label set for each language of the MasakhaNews dataset.}
    \label{tab:masakha-label}
\end{table*}

\newpage
\twocolumn

\section{Language Label}
\label{app:language-label}

We provide the label set in the source and target languages used in all the languages under study in MasakhaNews, NusaTranslation, AmericasNLI, and TweetSentimentMultilingual on Table~\ref{tab:masakha-label},  Table~\ref{tab:nusa-label},  Table~\ref{tab:anli-label},  and Table~\ref{tab:tsm-label}, respectively.

\begin{table}[!h]
    \centering
    \resizebox{\linewidth}{!}{
        \begin{tabular}{c|c|c|c}
            \toprule
            Language & \multicolumn{3}{c}{Label Set} \\
            \midrule
            eng & negative & neutral & positive \\
            \midrule
            % ara & سلبي & حيادي & إيجابي \\
            fra & négatif & neutre & positif \\
            deu & negativ & neutral & positiv \\
            % hin & नकारात्मक & तटस्थ & सकारात्मक \\
            ita & negativo & neutro & positivo \\
            por & negativo & neutro & positivo \\
            spa & negativo & neutral & positivo \\
            \bottomrule
        \end{tabular}
    }
    \caption{Label set for each language in  the TweetSentimentMultilingual dataset.}
    \label{tab:tsm-label}
\end{table}

\begin{table}[!h]
    \centering
    \resizebox{\linewidth}{!}{
        \begin{tabular}{c|c|c|c}
            \toprule
            Language & \multicolumn{3}{c}{Label Set} \\
            \midrule
            eng & negative & neutral & positive \\
            ind & negatif & netral & positif \\
            btk & negatif & netral & positif \\
            sun & negatif & netral & positif \\
            jav & negatif & netral & positif \\
            mad & negatif & netral & positif \\
            mak & negatif & netral & positif \\
            min & negatif & netral & positif \\
            \bottomrule
        \end{tabular}
    }
    \caption{Label set for each language of the NusaTranslation dataset.}
    \label{tab:nusa-label}
\end{table}

\begin{table}[!h]
    \centering
    \resizebox{\linewidth}{!}{
        \begin{tabular}{c|c|c|c}
            \toprule
            Language & \multicolumn{3}{c}{Label Set} \\
            \midrule
            eng & entailment & neutral & contradiction \\
            spa & vinculación & neutral & contradicción \\
            aym & vinculación & niwtrala & contradicción \\
            bzd & - & - & - \\
            cni & - & - & - \\
            grn & vinculación & ñemombyte & contradicción \\
            hch & - & - & - \\
            nah & - & - & - \\
            oto & vinculación & neutral & contradicción \\
            quy & hukllanakuy & chawpi & contradicción \\
            shp & - & - & - \\
            tar & - & - & - \\
            \bottomrule
        \end{tabular}
    }
    \caption{Label set for each language of the AmericasNLI dataset.}
    \label{tab:anli-label}
\end{table}

\newpage
\onecolumn

\section{Effect of Machine Translation Quality to X-ICL}
\label{app:machine-translation-effect}

\begin{table}[!h]
    \centering
    \resizebox{\linewidth}{!}{
        \begin{tabular}{cccccccc}
        \toprule
        \multirow{2}{*}{\textbf{Dataset}} & \textbf{Language} & \textbf{Language} & \textbf{chrF++} & \multicolumn{2}{c}{\textbf{XGLM}} & \multicolumn{2}{c}{\textbf{BLOOMZ}} \\
        \cmidrule(lr){5-6} \cmidrule(lr){7-8}
        & \textbf{Code} & \textbf{Name} & \textbf{(xxx2eng)} & \textbf{Zero-Shot (MT)} & \textbf{ICL (MT)} & \textbf{Zero-Shot (MT)} & \textbf{ICL (MT)} \\
        \midrule
        NusaTranslation & min & Minangkabau & 60.30 & 68.32 & 67.28 & 67.26 & 76.83 \\
        NusaTranslation & sun & Sundanese & 60.7 & 71.58 & 70.78 & 76.31 & 80.53 \\
        NusaTranslation & jav & Javanese & 61.4 & 71.26 & 68.35 & 73.89 & 78.95 \\
        \midrule
        AmericasNLI & aym & Aymara & 31.7 & 16.94 & 34.52 & 16.66 & 35.8 \\
        AmericasNLI & quy & Quechua & 32.7 & 16.66 & 37.24 & 16.66 & 39.19 \\
        AmericasNLI & grn & Guaraní & 47.6 & 16.66 & 34.42 & 16.66 & 37.79 \\
        \midrule
        TweetSentiMulti & spa & Spanish & 58.3 & 42.14 & 45.38 & 45.47 & 55.8 \\
        TweetSentiMulti & ita & Italian & 60.6 & 39.61 & 43.39 & 45.04 & 54.51 \\
        TweetSentiMulti & arb & Arabic & 64.6 & 33.97 & 50.66 & 35.73 & 55.28 \\
        TweetSentiMulti & hin & Hindi & 65. & 32.11 & 40.43 & 35.09 & 45.40 \\
        TweetSentiMulti & deu & German & 66.70 & 36.37 & 45.07 & 42.98 & 51.10 \\
        TweetSentiMulti & fra & French & 67.20 & 36.91 & 41.87 & 40.22 & 55.73 \\
        TweetSentiMulti & por & Portuguese & 70.60 & 39.04 & 45.02 & 42.21 & 53.42 \\
        \midrule
        MasakhaNews & yor & Yorùbá & 43.80 & 45.69 & 74.62 & 75.42 & 81.64 \\
        MasakhaNews & lug & Luganda & 44.90 & 34.71 & 59.98 & 70.54 & 62.82 \\
        MasakhaNews & sna & chiShona & 49.20 & 60.53 & 72.80 & 68.71 & 73.85 \\
        MasakhaNews & ibo & Igbo & 52.50 & 44.32 & 73.79 & 71.69 & 77.21 \\
        MasakhaNews & hau & Hausa & 55.30 & 43.99 & 59.74 & 67.30 & 67.19 \\
        MasakhaNews & amh & Amharic & 58.10 & 62.88 & 81.40 & 82.73 & 84.92 \\
        MasakhaNews & xho & isiXhosa & 58.50 & 33.41 & 65.66 & 58.36 & 63.30 \\
        MasakhaNews & swa & Kiswahili & 63.50 & 52.03 & 67.10 & 75.49 & 71.42 \\
        \midrule
        \multicolumn{4}{c}{\textbf{Pearson Correlation w/ chrF++}} & \textbf{0.416} & \textbf{0.102} & \textbf{0.247} & \textbf{0.238} \\
        \bottomrule
        \end{tabular}
    }
    \caption{Performance of NLLB 1.3B model on FLORES-200 with the machine-translated zero-shot and few-shot ICL performance of XGLM and BLOOMZ using the corresponding NLLB translation.}
    \label{tab:mt-correlation}
\end{table}

We showcase that the MT model performance plays a huge role in determining the language understanding quality through machine translation (MT). We showcase the MT model performance on the devtest subset of FLORES-200~\cite{goyal-etal-2022-flores} along with the zero-shot with MT and few-shot ICL with MT performance in Table~\ref{tab:mt-correlation}. The zero-shot (MT) performance has a low-to-moderate correlation with the machine translation quality (chrF++) of the model (0.416 for XGLM and 0.247 for BLOOMZ), while the few-shot ICL (MT) has a lower correlation (0.102 for XGLM and 0.238 for BLOOMZ) potentially due to the effect of other factors such as the semantic similarity exemplar selection and the quality of the ICL data itself. Our result indicates that, despite being effective for language understanding, the MT-based zero-shot and few-shot inference approach depends on the quality of the machine translation models. Moreover, an MT-based solution might not work as well for cultural-specific tasks which have been addressed in various works~\cite{kabra-etal-2023-multi,koto-etal-2023-large,wibowo2023copalid}.

\newpage
\onecolumn

\section{Effects of Source Languages}
\label{app:source-lang}

We explore alternative source languages for NusaTranslation~\cite{cahyawijaya2023nusawrites}. For NusaTranslation we utilize Indonesian as the source language because Indonesian is the closely related to the languages under study on the corresponding dataset and is widely spoken languages in the respective region. We modify both the prompt language and the source ICL dataset $D^{src}$.

\begin{figure}[!h]
    \centering
    % \includegraphics[width=0.8\linewidth]{images/semantic_similarity/tweetsentimulti.pdf}
    % \includegraphics[width=0.8\linewidth]{images/semantic_similarity/masakhanews.pdf}
    % \hfill\vline\hfill
    \includegraphics[width=0.7\linewidth]{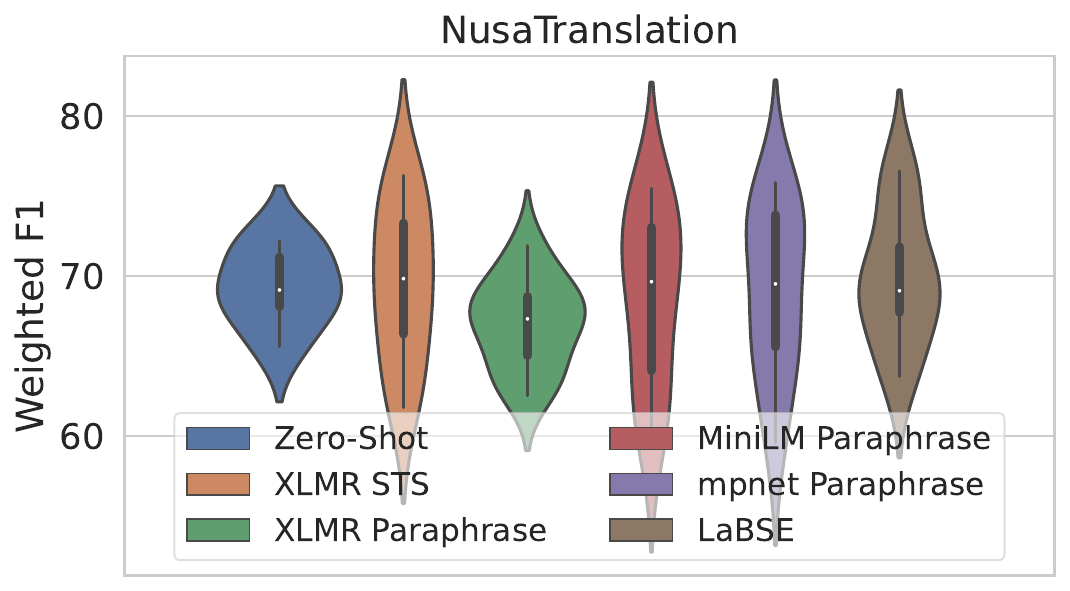}
    \includegraphics[width=0.7\linewidth]{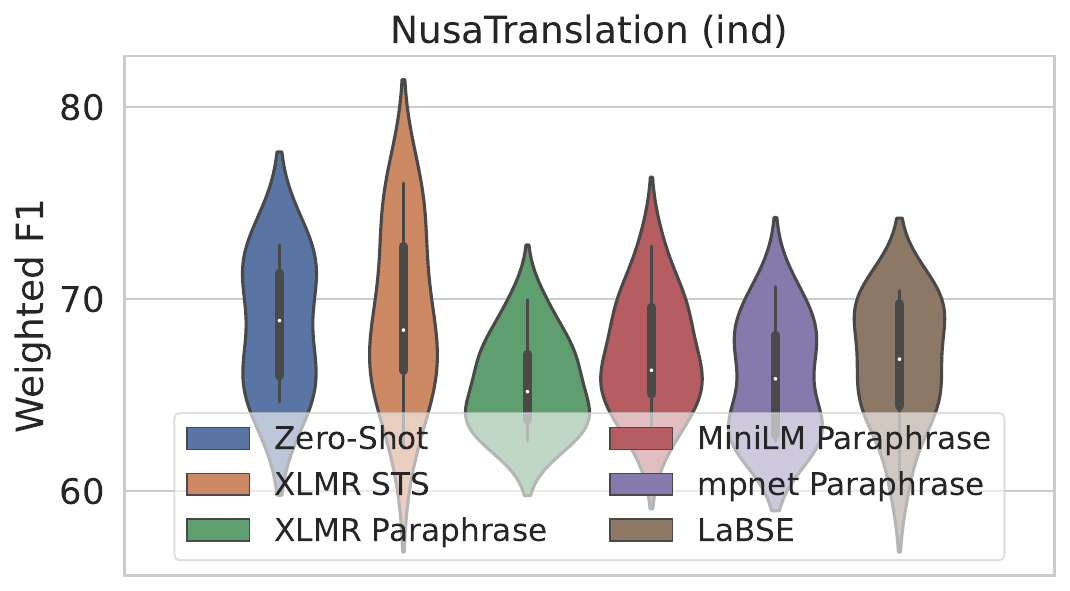}
    \caption{Performance of BLOOM-7B1 on NusaTranslation using \textbf{(top)} English prompt with English ICL exemplars and \textbf{(bottom)} Indonesian prompt with Indonesian ICL exemplars.}
    \vspace{-6pt}
    \label{fig:ind-eng}
\end{figure}

The result is shown in Figure~\ref{fig:ind-eng}. We can clearly see that in most cases, using English as the source language tends to produce better score than these closely related languages. Similar observation is also reported in prior works~\cite{cahyawijaya2023nusacrowd,asai2023buffet} which evaluates the prompt using different prompt language. Our experiment further extend the generalization to the X-ICL setting, where X-ICL using English exemplars outperforms X-ICL with a more closely related languages exemplars.

\newpage
\onecolumn

\section{Visualization of BLOOM Result}
\label{app:bloom_analysis}

\begin{figure*}[!h]
    \centering
    \begin{minipage}{.33\linewidth}
        \centering
        \begingroup
        \includegraphics[trim=0 10em 12em 0, width=\linewidth, clip]{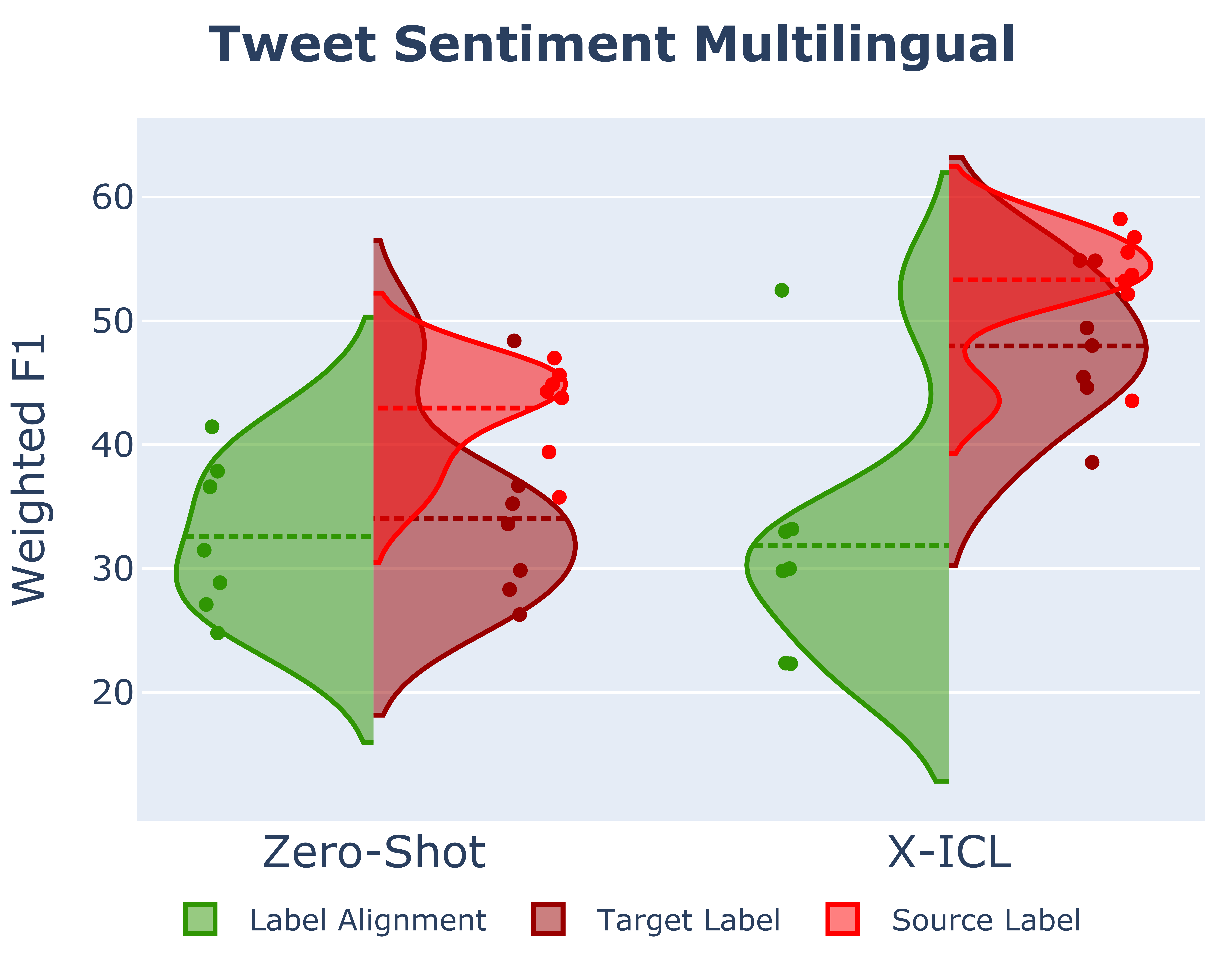}
        \endgroup
    \end{minipage}%
    \hspace{2pt}
    % \hfill\vline\hfill
    \begin{minipage}{.295\linewidth}
        \centering
        \begingroup
        \includegraphics[trim=0 10em 12em 0, width=\linewidth, clip]{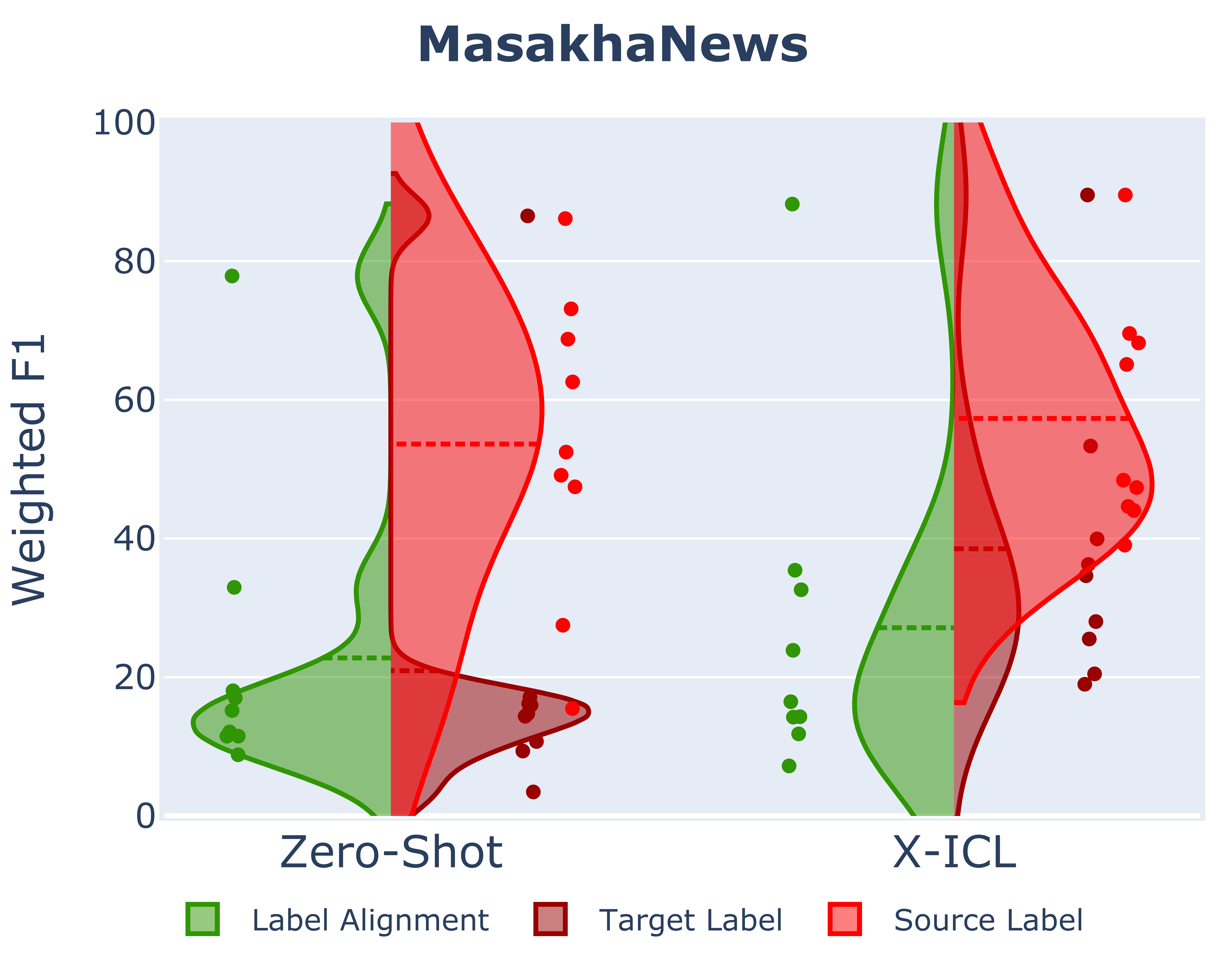}
        \endgroup
    \end{minipage}%
    \hspace{2pt}
    % \hfill\vline\hfill
    \begin{minipage}{.295\linewidth}
        \centering
        \begingroup
        \includegraphics[trim=0 10em 12em 0, width=\linewidth, clip]{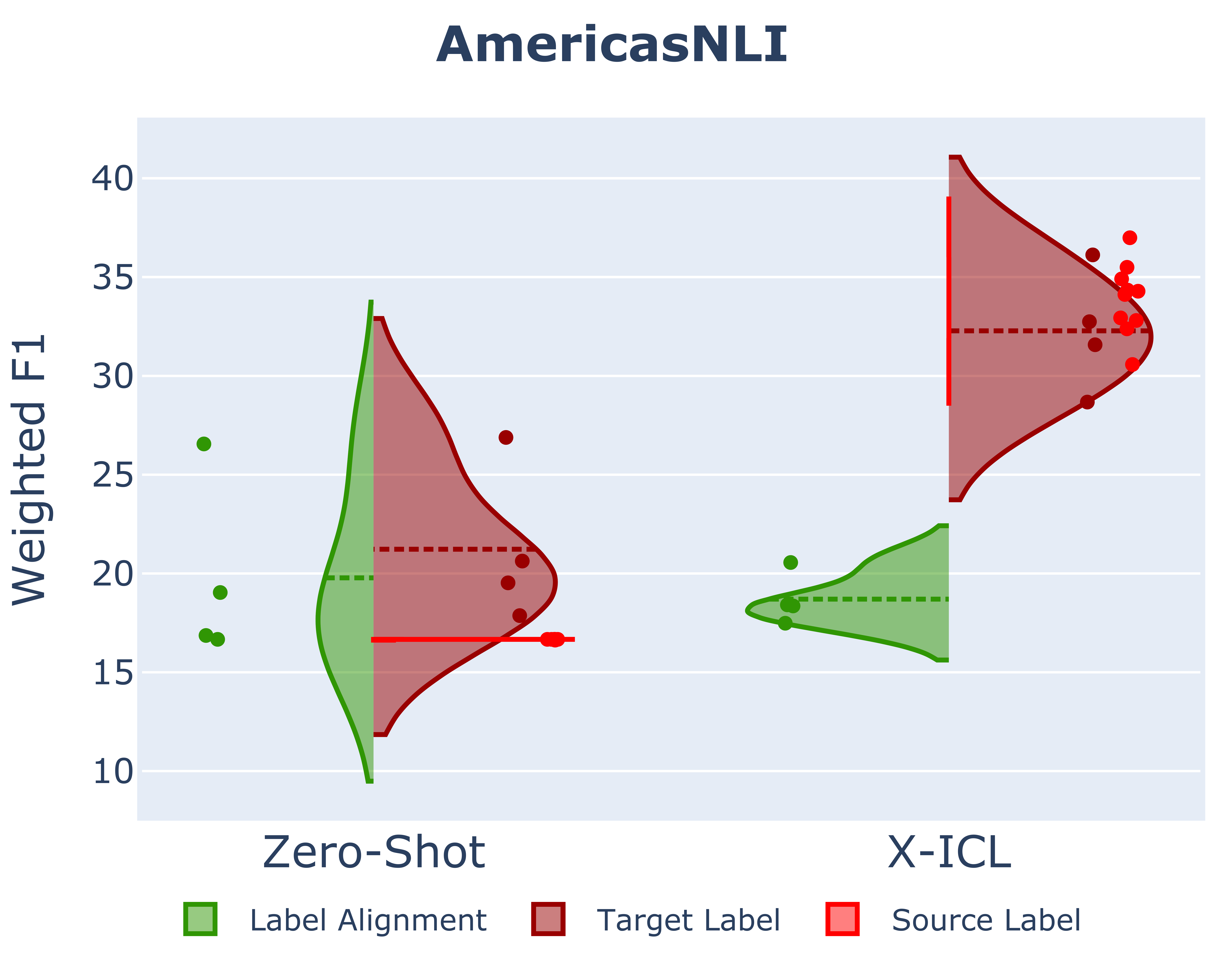}
        \endgroup
    \end{minipage}
    \vspace{-4pt}
    \caption{Performance of BLOOM-7B1 with in-context label alignment, target-only label, and source-only label on \textbf{(left)} higher-resource, \textbf{(center)} low-resource African, and \textbf{(right)} low-resource American languages.}
    \vspace{10pt}
\end{figure*}

\begin{figure*}[!h]
    \centering
    \begin{minipage}{.50\linewidth}
        \centering
        \begingroup
        \includegraphics[trim=0 0 0 0, width=\linewidth, clip]{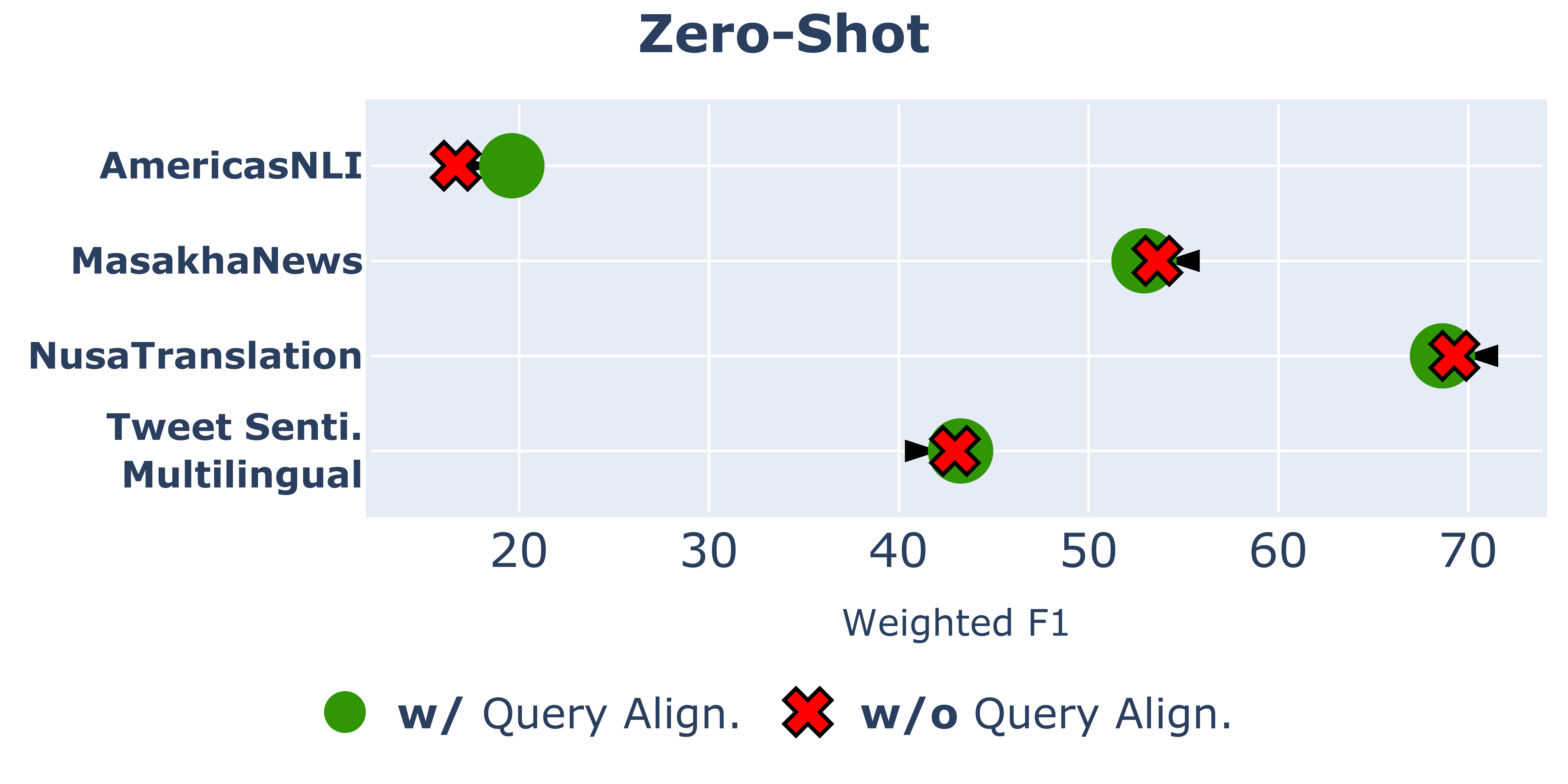}
        \endgroup
    \end{minipage}%
    \hspace{2pt}
    % \hfill\vline\hfill
    \begin{minipage}{.49\linewidth}
        \centering
        \begingroup
        \includegraphics[trim=0 0 0 0, width=\linewidth, clip]{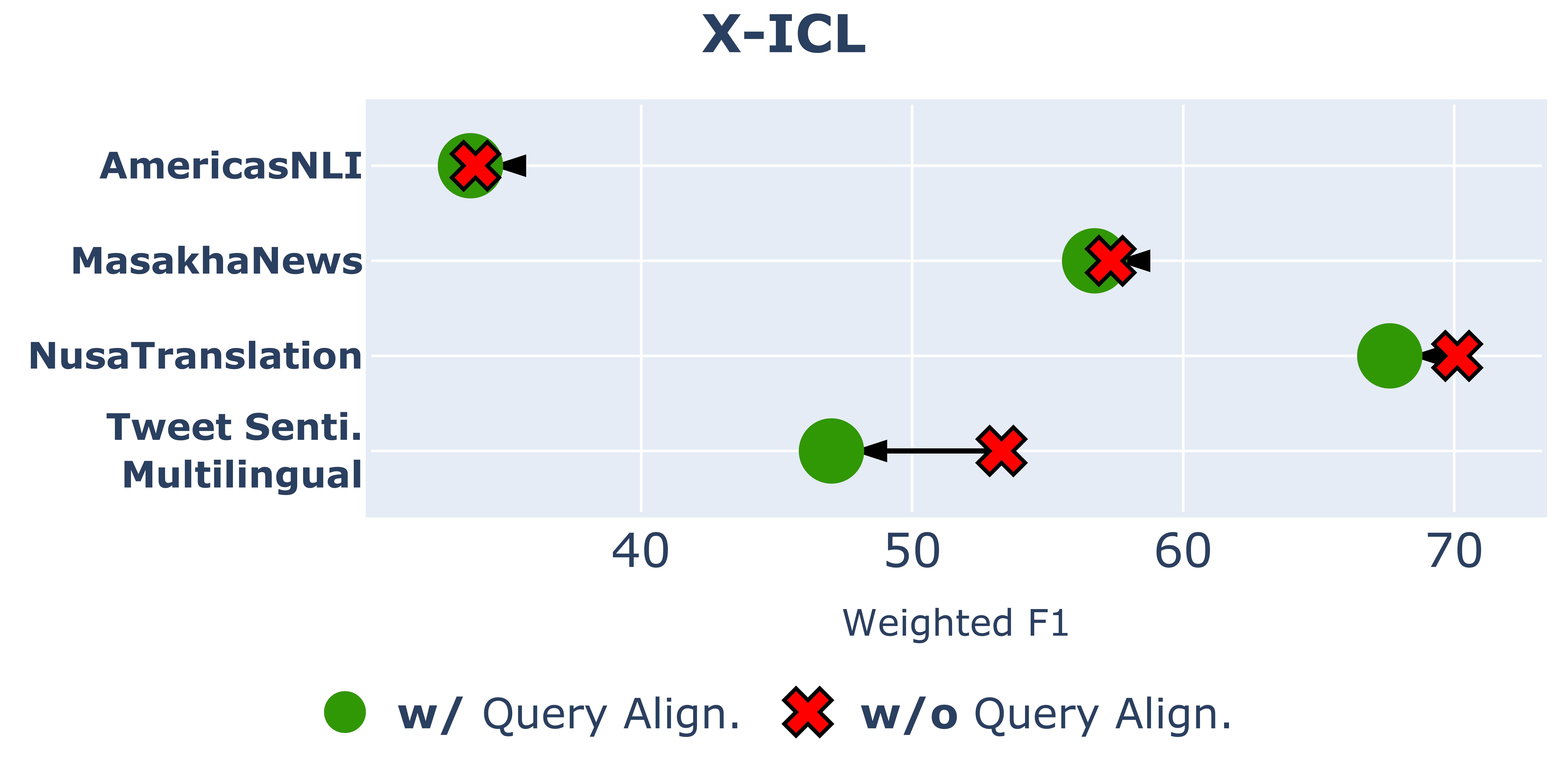}
        \endgroup
    \end{minipage}
    \caption{Performance of BLOOM-7B1 with and without query alignment on \textbf{(left)} higher-resource, \textbf{(center)} low-resource African, and \textbf{(right)} low-resource American languages.}
    \vspace{10pt}
\end{figure*}

\begin{figure}[!h]
    \centering
    \begin{minipage}{.46\linewidth}
        \centering
        \begingroup
        \includegraphics[trim=0 0 5em 0, width=\linewidth, clip]{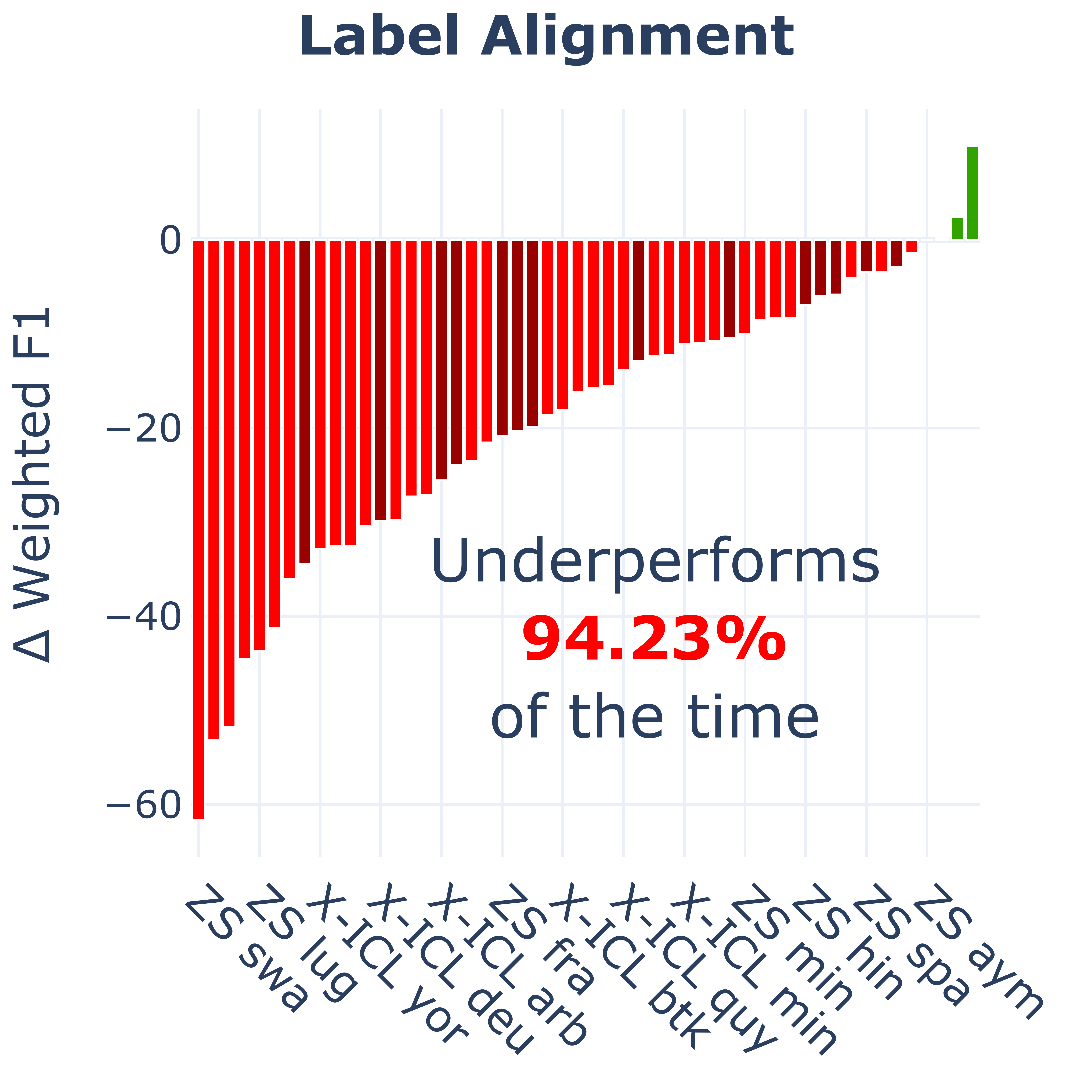}
        \endgroup
    \end{minipage}
    \begin{minipage}{.41\linewidth}
        \centering
        \begingroup
        \includegraphics[trim=27em 0 5em 0, width=\linewidth, clip]{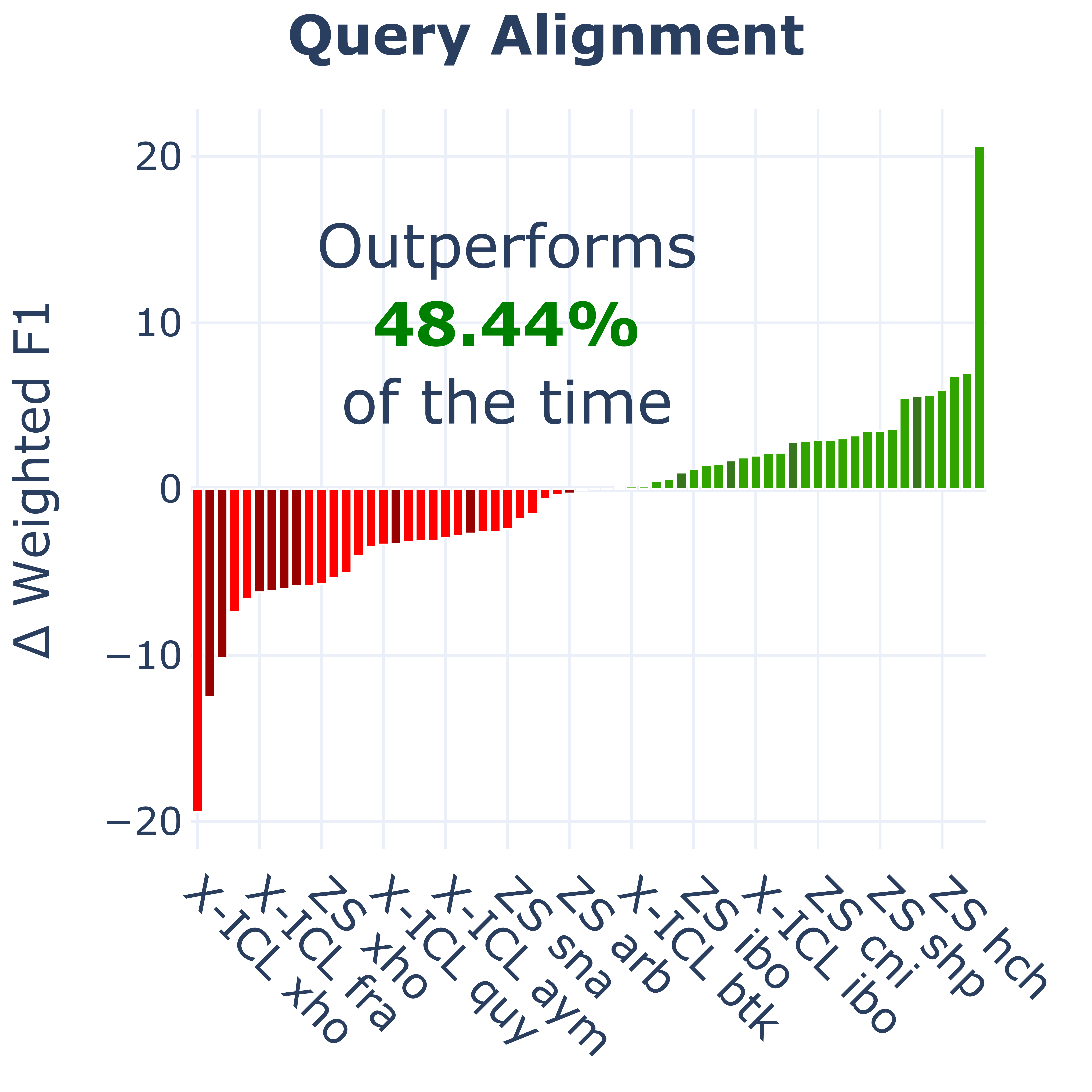}
        \endgroup
    \end{minipage}
    \caption{$\Delta$Weighted F1 of \textbf{(left)} in-context label alignment and \textbf{(right)} in-context query alignment against non-alignment baseline. A score < 0 indicates the in-context alignment degrades the performance.}
    \vspace{-8pt}
    \label{fig:label_vs_input-bloom}
\end{figure}

\begin{figure*}[!h]
    \centering
    \begin{minipage}{.23\linewidth}
        \centering
        \begingroup
        \includegraphics[trim=0 2em 0 0, width=\linewidth]{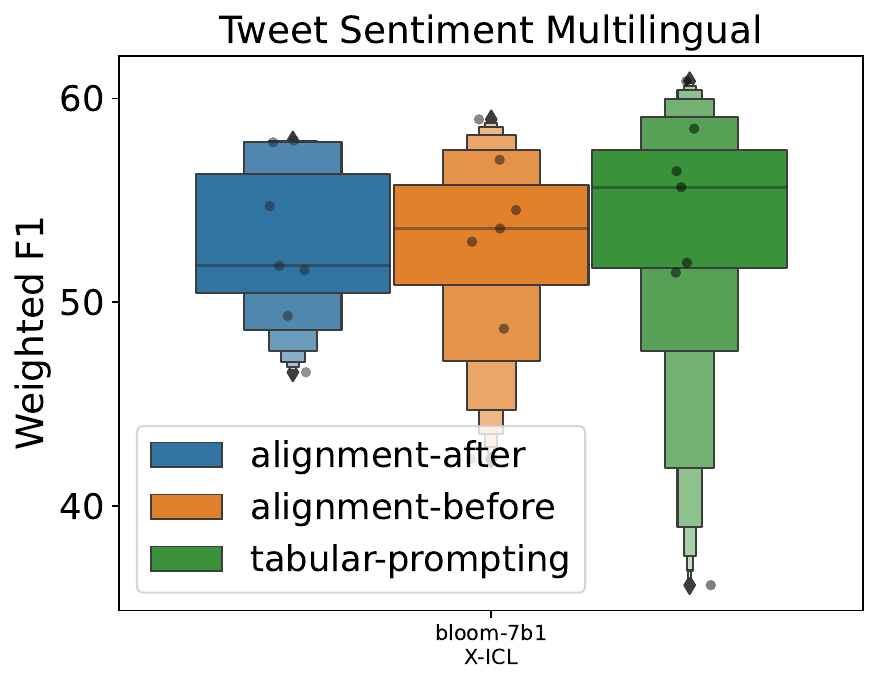}
        \endgroup
    \end{minipage}%
    % \hspace{2pt}
    % \hfill\vline\hfill
    \begin{minipage}{.23\linewidth}
        \centering
        \begingroup
        \includegraphics[trim=0 2em 0 0, width=\linewidth]{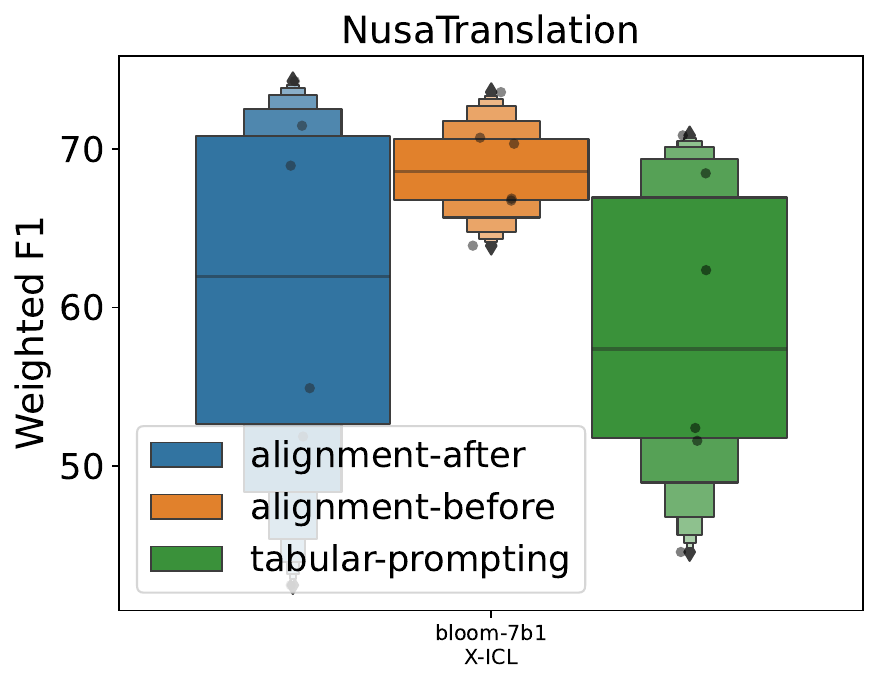}
        \endgroup
    \end{minipage}%
    % \hspace{2pt}
    % \hfill\vline\hfill
    \begin{minipage}{.23\linewidth}
        \centering
        \begingroup
        \includegraphics[trim=0 2em 0 0, width=\linewidth]{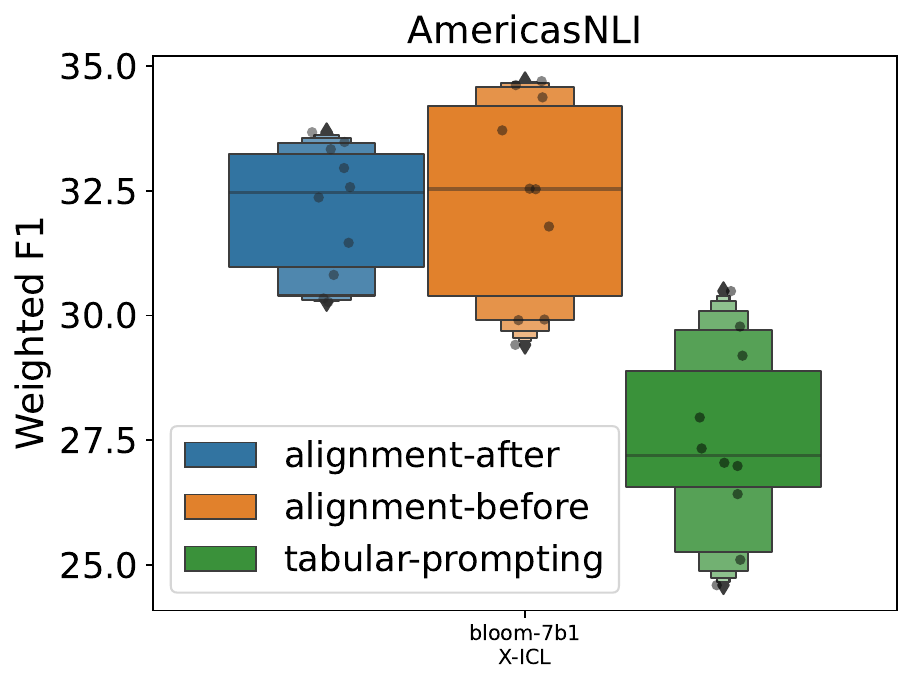}
        \endgroup
    \end{minipage}
    % \hspace{2pt}
    % \hfill\vline\hfill
    \begin{minipage}{.225\linewidth}
        \centering
        \begingroup
        \includegraphics[trim=0 2em 0 0, width=\linewidth]{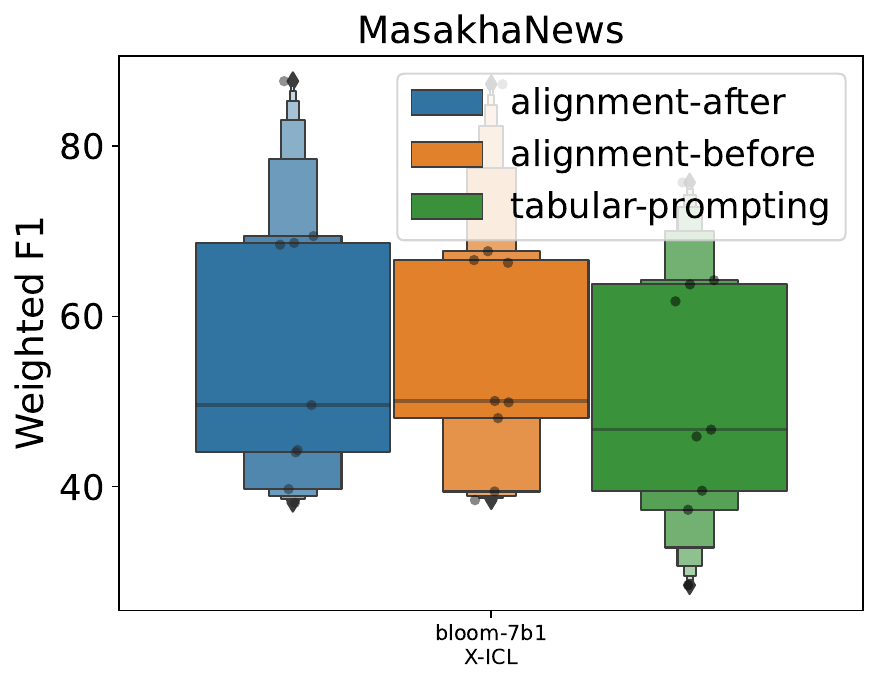}
        \endgroup
    \end{minipage}
    \caption{Performance of BLOOM-7B1 with different alignment formats ordered by the degree of formatting consistency on \textbf{(1)} higher-resource languages, \textbf{(2)} low-resource Indonesian languages, \textbf{(3)} low-resource American languages, and \textbf{(4)} low-resource African languages.}
    \vspace{-4pt}
    \label{fig:result_format-bloom}
\end{figure*}

\begin{figure*}[!h]
    \centering
    \begin{minipage}{.24\linewidth}
        \centering
        \begingroup
        \includegraphics[trim=0 2em 0 0, width=\linewidth]{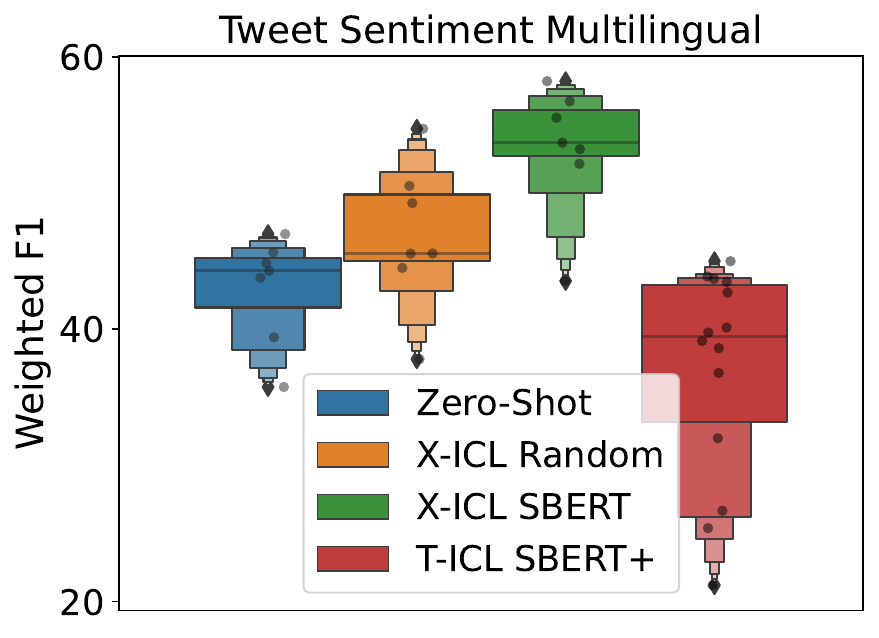}
        \endgroup
    \end{minipage}%
    % \hspace{2pt}
    % \hfill\vline\hfill
    \begin{minipage}{.24\linewidth}
        \centering
        \begingroup
        \includegraphics[trim=0 2em 0 0, width=\linewidth]{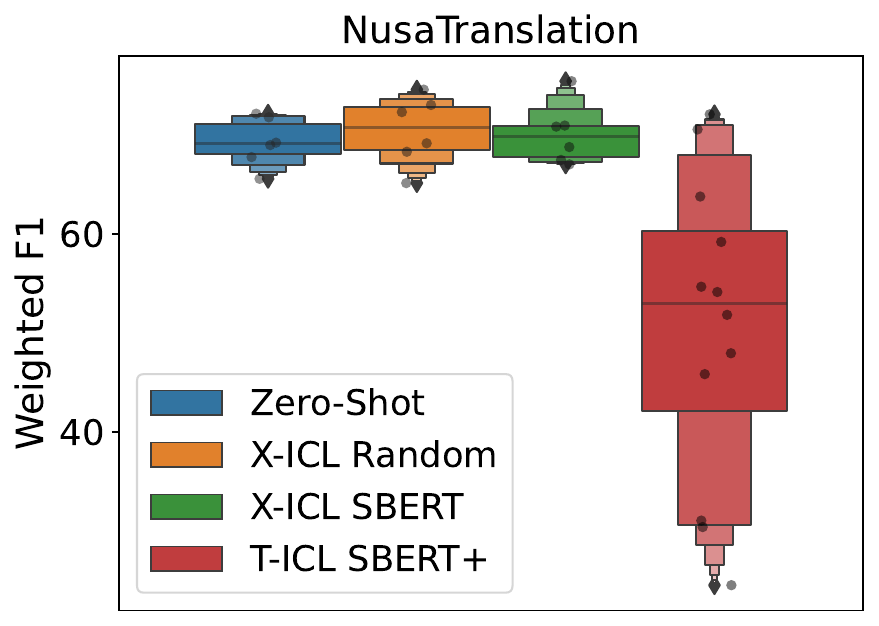}
        \endgroup
    \end{minipage}%
    % \hspace{2pt}
    % \hfill\vline\hfill
    \begin{minipage}{.24\linewidth}
        \centering
        \begingroup
        \includegraphics[trim=0 2em 0 0, width=\linewidth]{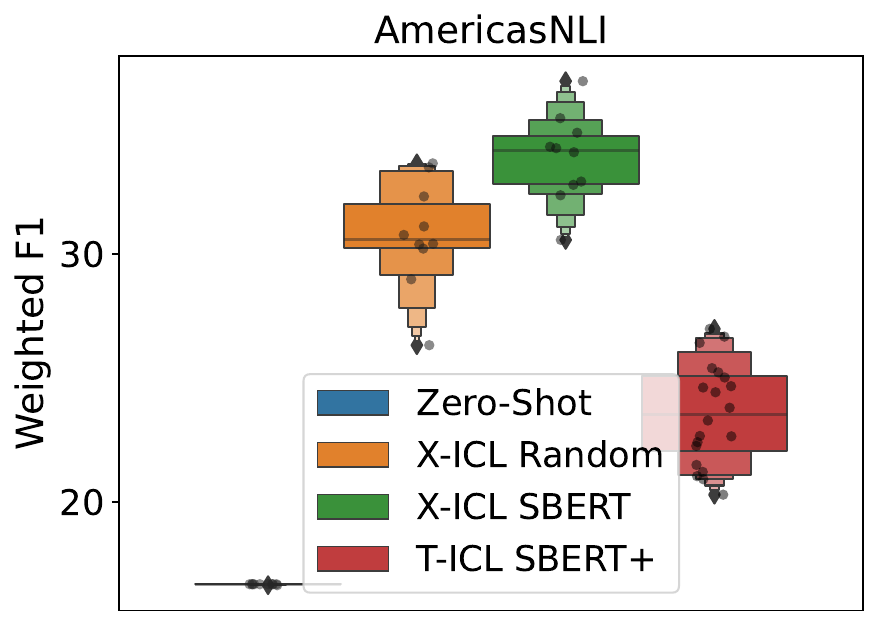}
        \endgroup
    \end{minipage}
    % \hspace{2pt}
    % \hfill\vline\hfill
    \begin{minipage}{.235\linewidth}
        \centering
        \begingroup
        \includegraphics[trim=0 2em 0 0, width=\linewidth]{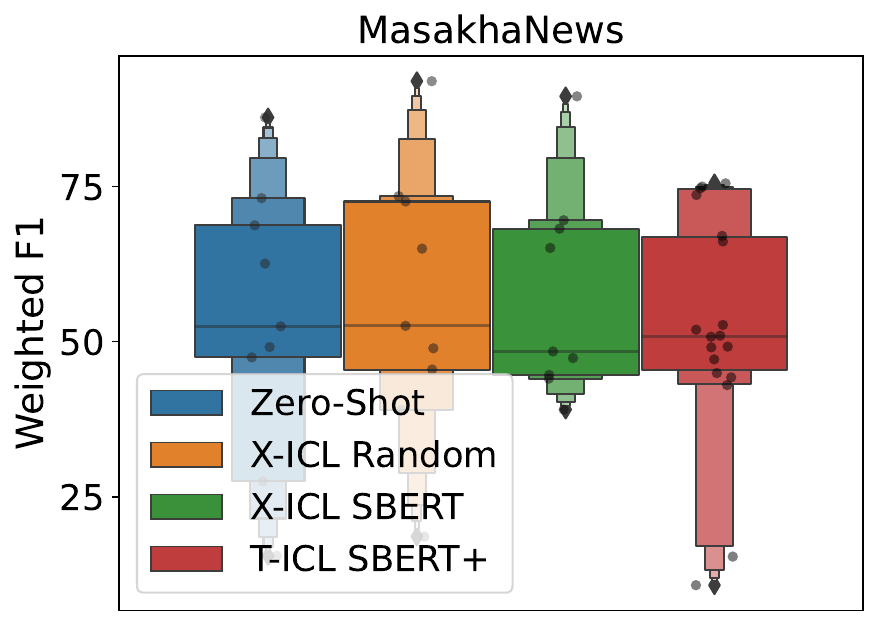}
        \endgroup
    \end{minipage}
    \caption{Performance of BLOOM-7B1 with different in-context learning retrievals covering monolingual, cross-lingual, translation semantic similarity on \textbf{(1)} higher-resource languages, \textbf{(2)} low-resource Indonesian languages, \textbf{(3)} low-resource American languages, and \textbf{(4)} low-resource African languages.}
    \vspace{-6pt}
    \label{fig:result_xss-bloom}
\end{figure*}

\section{Detailed Per Dataset Results}
\label{app:all_results}

The detailed the main results for each different inference type for XGLM-7.5B in Table~\ref{tab:xglm-tsm}, Table~\ref{tab:xglm-masakha},  Table~\ref{tab:xglm-nusa}, and Table~\ref{tab:xglm-anli} for TweetSentimentMultilingual MasakhaNews, NusaTranslation, AmericasNLI, respectively. The detailed results for each different inference type for BLOOM-7B1 in Table~\ref{tab:bloom-tsm}, Table~\ref{tab:bloom-masakha},  Table~\ref{tab:bloom-nusa}, and Table~\ref{tab:bloom-anli} for TweetSentimentMultilingual MasakhaNews, NusaTranslation, AmericasNLI, respectively.

\begin{table}[!h]
    \centering
    \resizebox{0.8\linewidth}{!}{%
        \begin{tabular}{l|c|c|c|c|c|c|c}
        \toprule
        \textbf{Inference Type} &       \textbf{arb} &       \textbf{deu} &       \textbf{fra} &       \textbf{hin} &       \textbf{ita} &       \textbf{por} &       \textbf{spa} \\
        \midrule
        % Zero-Shot & 38.19 & 39.82 & 32.82 & 32.09 & 38.39 & 44.83 & 51.04 \\
        Zero-Shot \\
          \quad Source-Only Label & 38.19 & 39.82 & 32.82 & 32.09 & 38.39 & 44.83 & 51.04 \\
          \quad \quad + Query Alignment & 38.28 & 43.28 & 40.27 & 34.18 & 38.58 & 43.39 & 42.15 \\
          \quad Target-Only Label & 34.44 & 48.99 & 39.86 & 19.12 & 42.25 & 36.85 & 47.88 \\
          \quad Label Alignment & 21.15 & 35.41 & 27.97 & 28.50 & 31.85 & 28.82 & 30.18 \\
        Zero-Shot (MT) & 33.97 & 36.37 & 36.91 & 32.11 & 39.61 & 39.04 & 42.14 \\
        ICL Random & 40.39 & 38.94 & 36.50 & 36.16 & 37.10 & 36.85 & 44.66 \\
        ICL SBERT & 46.60 & 45.56 & 54.04 & 34.02 & 48.43 & 51.01 & 45.90 \\
        ICL SBERT (MT) & 50.66 & 45.07 & 41.87 & 40.43 & 43.39 & 45.02 & 45.38 \\
        X-ICL Random & 35.53 & 40.16 & 37.38 & 32.49 & 40.98 & 39.46 & 39.83 \\
        % X-ICL SBERT & 49.21 & 47.35 & 42.80 & 38.15 & 47.24 & 48.01 & 47.67 \\
        X-ICL SBERT \\
          \quad Source-Only Label & 49.21 & 47.35 & 42.80 & 38.15 & 47.24 & 48.01 & 47.67 \\
          \quad \quad + Query Alignment & 45.28 & 48.85 & 46.35 & 39.62 & 44.83 & 50.65 & 44.20 \\
          \quad Target-Only Label & 47.88 & 45.05 & 43.37 & 37.23 & 42.99 & 46.40 & 42.58 \\
          \quad Label Alignment & 29.51 & 27.15 & 37.97 & 30.10 & 44.50 & 40.91 & 31.96 \\
        \bottomrule
        \end{tabular}
    }
    \caption{Experiment results for XGLM-7.5B on TweetSentimentMultilingual dataset. "-" denotes the experiment is not conducted due to no machine translation system is available.}
    \label{tab:xglm-tsm}
\end{table}

\begin{table}[!h]
    \centering
    \resizebox{\linewidth}{!}{%
        \begin{tabular}{l|c|c|c|c|c|c|c|c|c}
        \toprule
             \textbf{Inference Type} &       \textbf{amh} &       \textbf{hau} &       \textbf{ibo} &       \textbf{lug} &       \textbf{pcm} &       \textbf{sna} &       \textbf{swa} &       \textbf{xho} &       \textbf{yor} \\
        \midrule
        % Zero-Shot & 17.62 & 32.64 & 51.11 & 22.80 & 57.07 & 42.66 & 49.93 & 28.60 & 54.96 \\
        Zero-Shot \\
          \quad Source-Only Label & 17.62 & 32.64 & 51.11 & 22.80 & 57.07 & 42.66 & 49.93 & 28.60 & 54.96 \\
          \quad \quad + Query Alignment & 25.79 & 36.81 & 59.07 & 39.51 & 72.65 & 42.68 & 58.12 & 21.97 & 48.28 \\
          \quad Target-Only Label & 11.92 &  7.36 &  3.72 & 13.94 & 58.59 & 15.25 & 53.29 &  2.11 & 12.88 \\
          \quad Label Alignment & 10.19 &  7.08 &  4.19 & 14.40 & 60.63 & 19.18 & 47.46 & 22.12 & 17.80 \\
        Zero-Shot (MT) & 62.88 & 43.99 & 44.32 & 34.71 & 56.73 & 60.53 & 52.03 & 33.41 & 45.69 \\
        ICL Random & 20.36 & 37.92 & 63.33 & 38.93 & 83.01 & 43.03 & 65.62 & 49.26 & 65.65 \\
        ICL SBERT & 60.75 & 61.39 & 69.86 & 48.23 & 93.02 & 59.56 & 73.07 & 43.79 & 70.84 \\
        ICL SBERT (MT) & 81.40 & 59.74 & 73.79 & 59.98 & 87.20 & 72.80 & 67.10 & 65.66 & 74.62 \\
        X-ICL Random & 24.11 & 38.01 & 62.32 & 46.23 & 85.38 & 51.70 & 58.98 & 47.79 & 66.77 \\
        % X-ICL SBERT & 55.18 & 37.08 & 64.28 & 46.95 & 88.27 & 41.87 & 63.05 & 49.10 & 65.74 \\
        X-ICL SBERT \\
          \quad Source-Only Label & 55.18 & 37.08 & 64.28 & 46.95 & 88.27 & 41.87 & 63.05 & 49.10 & 65.74 \\
          \quad \quad + Query Alignment & 51.43 & 40.53 & 62.05 & 44.70 & 86.61 & 44.58 & 65.59 & 40.27 & 57.24 \\
          \quad Target-Only Label & 53.06 & 19.19 & 27.87 & 27.99 & 88.59 & 25.49 & 37.02 & 25.71 & 31.64 \\
          \quad Label Alignment & 10.19 & 13.79 &  6.40 & 12.35 & 90.88 & 12.71 & 62.73 & 21.41 & 16.00 \\
        \bottomrule
        \end{tabular}
    }
    \caption{Experiment results for XGLM-7.5B on MasakhaNews dataset. "-" denotes the experiment is not conducted due to no machine translation system is available. SBERT denotes exemplar selection using a semantic similarity model.}
    \label{tab:xglm-masakha}
\end{table}

\begin{table}[!h]
    \centering
    \resizebox{0.7\linewidth}{!}{%
        \begin{tabular}{l|c|c|c|c|c|c}
        \toprule
             \textbf{Inference Type} &       \textbf{btk} &       \textbf{jav} &       \textbf{mad} &       \textbf{mak} &       \textbf{min} &       \textbf{sun} \\
        \midrule
        % Zero-Shot & 58.60 & 62.92 & 60.75 & 55.90 & 64.29 & 63.67 \\
        Zero-Shot \\
          \quad Source-Only Label & 58.60 & 62.92 & 60.75 & 55.90 & 64.29 & 63.67 \\
          \quad \quad + Query Alignment & 52.60 & 62.77 & 56.74 & 49.50 & 63.51 & 57.08 \\
          \quad Target-Only Label & 41.63 & 47.52 & 45.85 & 47.53 & 42.20 & 43.12 \\
          \quad Label Alignment & 30.58 & 28.17 & 31.81 & 37.79 & 28.59 & 29.83 \\
        Zero-Shot (MT) &       - & 71.26 &       - & 60.68 & 68.32 & 71.58 \\
        ICL Random & 59.68 & 60.85 & 59.28 & 60.83 & 62.91 & 59.52 \\
        ICL SBERT & 59.59 & 60.84 & 60.81 & 62.39 & 66.11 & 61.68 \\
        ICL SBERT (MT) &       - & 68.35 &       - & 59.61 & 67.28 & 70.78 \\
        X-ICL Random & 60.54 & 61.74 & 63.02 & 58.21 & 63.87 & 61.66 \\
        % X-ICL SBERT & 60.69 & 62.83 & 59.78 & 60.30 & 63.95 & 62.44 \\
        X-ICL SBERT \\
          \quad Source-Only Label & 60.69 & 62.83 & 59.78 & 60.30 & 63.95 & 62.44 \\
          \quad \quad + Query Alignment & 52.41 & 60.38 & 53.06 & 52.77 & 61.36 & 56.62 \\
          \quad Target-Only Label & 56.02 & 59.57 & 56.83 & 48.61 & 64.80 & 59.35 \\
          \quad Label Alignment & 55.60 & 61.13 & 57.45 & 55.10 & 62.52 & 58.42 \\
        \bottomrule
        \end{tabular}
    }
    \caption{Experiment results for XGLM-7.5B on NusaTranslation dataset. "-" denotes the experiment is not conducted due to no machine translation system is available. SBERT denotes exemplar selection using a semantic similarity model.}
    \label{tab:xglm-nusa}
\end{table}

\begin{table}[!h]
    \centering
    \resizebox{\linewidth}{!}{%
        \begin{tabular}{l|c|c|c|c|c|c|c|c|c|c}
        \toprule
             \textbf{Inference Type} &       \textbf{aym} &       \textbf{bzd} &       \textbf{cni} &        \textbf{grn} &       \textbf{hch} &       \textbf{nah} &       \textbf{oto} &       \textbf{quy} &       shp &       \textbf{tar} \\
        \midrule
        % Zero-Shot & 16.68 & 16.66 & 16.66 & 16.61 & 17.68 & 18.88 & 19.31 & 16.66 & 17.62 & 16.66 \\
        Zero-Shot \\
          \quad Source-Only Label & 16.68 & 16.66 & 16.66 & 16.61 & 17.68 & 18.88 & 19.31 & 16.66 & 17.62 & 16.66 \\
          \quad \quad + Query Alignment & 29.56 & 30.79 & 28.04 & 29.15 & 32.07 & 33.05 & 32.23 & 33.77 & 32.33 & 30.82 \\
          \quad Target-Only Label & 19.88 &       - &       - & 17.79 &       - &       - & 17.69 & 22.63 &       - &       - \\
          \quad Label Alignment & 22.31 &       - &       - & 17.90 &       - &       - & 25.52 & 29.17 &       - &       - \\
        Zero-Shot (MT) & 16.94 &       - &       - & 16.66 &       - &       - &       - & 16.66 &       - &       - \\
        ICL Random & 32.43 & 28.66 & 30.42 & 29.91 & 29.15 & 32.70 & 29.63 & 32.98 & 30.28 & 31.74 \\
        ICL SBERT & 34.65 & 28.26 & 30.62 & 34.34 & 31.10 & 33.89 & 28.02 & 32.64 & 28.90 & 30.97 \\
        ICL SBERT (MT) & 34.52 &       - &       - & 34.42 &       - &       - &       - & 37.24 &       - &       - \\
        X-ICL Random & 28.96 & 32.55 & 30.72 & 28.95 & 33.01 & 33.55 & 28.88 & 34.78 & 32.16 & 31.43 \\
        % X-ICL SBERT & 33.20 & 33.99 & 31.99 & 33.88 & 31.00 & 30.80 & 30.97 & 34.24 & 26.95 & 32.74 \\
        X-ICL SBERT \\
          \quad Source-Only Label & 33.20 & 33.99 & 31.99 & 33.88 & 31.00 & 30.80 & 30.97 & 34.24 & 26.95 & 32.74 \\
          \quad \quad + Query Alignment & 35.30 & 32.83 & 35.60 & 32.71 & 33.04 & 28.05 & 31.02 & 34.29 & 30.57 & 32.97 \\
          \quad Target-Only Label & 30.58 &       - &       - & 34.76 &       - &       - & 31.19 & 28.32 &       - &       - \\
          \quad Label Alignment & 25.30 &       - &       - & 17.37 &       - &       - & 25.79 & 25.61 &       - &       - \\
        \bottomrule
        \end{tabular}
    }
    \caption{Experiment results for XGLM-7.5B on AmericasNLI dataset. "-" denotes the experiment is not conducted due to no machine translation system is available.}
    \label{tab:xglm-anli}
\end{table}

\begin{table}[!h]
    \centering
    \resizebox{0.8\linewidth}{!}{%
        \begin{tabular}{l|c|c|c|c|c|c|c}
        \toprule
             \textbf{Inference Type} &       \textbf{arb} &       \textbf{deu} &       \textbf{fra} &       \textbf{hin} &       \textbf{ita} &       \textbf{por} &       \textbf{spa} \\
        \midrule
        % Zero-Shot & 43.77 & 39.40 & 45.62 & 35.75 & 46.98 & 44.28 & 44.84 \\
        Zero-Shot \\
          \quad Source-Only Label & 43.77 & 39.40 & 45.62 & 35.75 & 46.98 & 44.28 & 44.84 \\
          \quad \quad + Query Alignment & 43.51 & 40.38 & 42.96 & 37.45 & 40.98 & 49.85 & 47.64 \\
          \quad Target-Only Label & 33.59 & 28.30 & 29.85 & 26.28 & 36.68 & 35.23 & 48.37 \\
          \quad Label Alignment & 37.86 & 36.60 & 24.80 & 28.86 & 27.10 & 31.47 & 41.44 \\
        Zero-Shot (MT) & 35.73 & 42.98 & 40.22 & 35.09 & 45.04 & 42.21 & 45.47 \\
        ICL Random & 41.71 & 50.44 & 37.72 & 37.24 & 49.86 & 48.58 & 51.10 \\
        ICL SBERT & 51.17 & 55.83 & 57.67 & 38.27 & 51.81 & 57.68 & 60.28 \\
        ICL SBERT (MT) & 55.28 & 51.10 & 55.73 & 45.40 & 54.51 & 53.42 & 55.83 \\
        X-ICL Random & 44.50 & 45.54 & 45.56 & 37.79 & 50.52 & 49.26 & 54.71 \\
        % X-ICL SBERT & 55.52 & 52.14 & 53.22 & 43.53 & 53.69 & 58.20 & 56.73 \\
        X-ICL SBERT \\
          \quad Source-Only Label & 55.52 & 52.14 & 53.22 & 43.53 & 53.69 & 58.20 & 56.73 \\
          \quad \quad + Query Alignment & 45.38 & 46.03 & 47.01 & 43.65 & 41.19 & 52.38 & 53.46 \\
          \quad Target-Only Label & 44.61 & 47.99 & 45.45 & 38.57 & 54.85 & 54.84 & 49.41 \\
          \quad Label Alignment & 29.99 & 22.32 & 32.98 & 33.18 & 29.80 & 52.45 & 22.36 \\
        \bottomrule
        \end{tabular}
    }
    \caption{Experiment results for BLOOM-7B1 model on TweetSentimentMultilingual dataset. "-" denotes the experiment is not conducted due to no machine translation system is available.}
    \label{tab:bloom-tsm}
\end{table}

\begin{table}[!h]
    \centering
    \resizebox{\linewidth}{!}{%
        \begin{tabular}{l|c|c|c|c|c|c|c|c|c}
        \toprule
             \textbf{Inference Type} &       \textbf{amh} &       \textbf{hau} &       \textbf{ibo} &       \textbf{lug} &       \textbf{pcm} &       \textbf{sna} &       \textbf{swa} &       \textbf{xho} &       \textbf{yor} \\
        \midrule
        % Zero-Shot & 15.47 & 47.45 & 62.58 & 52.46 & 86.14 & 49.12 & 73.12 & 27.49 & 68.75 \\
        Zero-Shot \\
          \quad Source-Only Label & 15.47 & 47.45 & 62.58 & 52.46 & 86.14 & 49.12 & 73.12 & 27.49 & 68.75 \\
          \quad \quad + Query Alignment & 43.33 & 45.99 & 71.56 & 51.43 & 89.22 & 38.83 & 71.87 & 24.61 & 73.66 \\
          \quad Target-Only Label & 10.72 &  9.34 & 17.11 & 14.36 & 86.53 & 16.18 & 14.73 &  3.44 & 15.97 \\
          \quad Label Alignment & 12.10 & 11.48 & 18.04 &  8.81 & 77.86 & 32.95 & 11.49 & 15.18 & 17.01 \\
        Zero-Shot (MT) & 82.73 & 67.30 & 71.69 & 70.54 & 84.50 & 68.71 & 75.49 & 58.36 & 75.42 \\
        ICL Random & 26.74 & 42.29 & 73.91 & 45.04 & 85.46 & 49.53 & 73.59 & 36.86 & 72.71 \\
        ICL SBERT & 61.84 & 60.77 & 79.24 & 49.86 & 92.19 & 66.67 & 74.57 & 43.63 & 79.28 \\
        ICL SBERT (MT) & 84.92 & 67.19 & 77.21 & 62.82 & 90.23 & 73.85 & 71.42 & 63.30 & 81.64 \\
        X-ICL Random & 18.57 & 45.50 & 72.59 & 48.92 & 91.98 & 52.54 & 64.99 & 38.98 & 73.45 \\
        % X-ICL SBERT & 47.35 & 39.04 & 69.56 & 48.41 & 89.54 & 44.62 & 65.10 & 44.04 & 68.20 \\
        X-ICL SBERT \\
          \quad Source-Only Label & 47.35 & 39.04 & 69.56 & 48.41 & 89.54 & 44.62 & 65.10 & 44.04 & 68.20 \\
          \quad \quad + Query Alignment & 36.09 & 42.43 & 63.76 & 45.88 & 84.34 & 46.71 & 69.99 & 21.78 & 65.26 \\
          \quad Target-Only Label & 53.33 & 18.98 & 36.25 & 25.50 & 89.54 & 28.02 & 39.94 & 20.47 & 34.63 \\
          \quad Label Alignment & 23.87 & 11.82 & 16.45 &  7.20 & 88.23 & 14.23 & 32.60 & 14.29 & 35.42 \\
        \bottomrule
        \end{tabular}
    }
    \caption{Experiment results for BLOOM-7B1 model on MasakhaNews dataset. "-" denotes the experiment is not conducted due to no machine translation system is available.}
    \label{tab:bloom-masakha}
\end{table}

\begin{table}[!h]
    \centering
    \resizebox{0.7\linewidth}{!}{%
        \begin{tabular}{l|c|c|c|c|c|c}
        \toprule
             \textbf{Inference Type} &       \textbf{btk} &       \textbf{jav} &       \textbf{mad} &       \textbf{mak} &       \textbf{min} &       \textbf{sun} \\
        \midrule
        % Zero-Shot & 65.58 & 69.00 & 67.78 & 69.24 & 72.17 & 71.80 \\
        Zero-Shot \\
          \quad Source-Only Label & 65.58 & 69.00 & 67.78 & 69.24 & 72.17 & 71.80 \\
          \quad \quad + Query Alignment & 65.50 & 71.13 & 67.20 & 61.87 & 72.14 & 73.98 \\
          \quad Target-Only Label & 66.76 & 68.22 & 67.22 & 64.21 & 67.79 & 68.12 \\
          \quad Label Alignment & 61.62 & 60.77 & 59.31 & 58.57 & 62.25 & 60.89 \\        
        Zero-Shot (MT) &       - & 73.89 &       - & 57.87 & 67.26 & 76.31 \\
        ICL Random & 65.68 & 68.40 & 65.33 & 62.84 & 70.45 & 65.32 \\
        ICL SBERT & 62.84 & 72.70 & 64.38 & 61.77 & 76.27 & 75.04 \\
        ICL SBERT (MT) &       - & 78.95 &       - & 67.56 & 76.83 & 80.53 \\
        X-ICL Random & 68.32 & 73.05 & 69.17 & 65.15 & 74.60 & 72.33 \\
        % X-ICL SBERT & 67.04 & 70.86 & 68.79 & 67.49 & 75.45 & 70.97 \\
        X-ICL SBERT \\
          \quad Source-Only Label & 67.04 & 70.86 & 68.79 & 67.49 & 75.45 & 70.97 \\
          \quad \quad + Query Alignment & 67.18 & 68.30 & 66.24 & 64.39 & 70.10 & 69.47 \\
          \quad Target-Only Label & 59.99 & 69.00 & 62.48 & 61.28 & 72.53 & 71.40 \\
          \quad Label Alignment & 48.97 & 43.81 & 47.30 & 34.98 & 64.47 & 55.31 \\
        \bottomrule
        \end{tabular}
    }
    \caption{Experiment results for BLOOM-7B1 model on NusaTranslation dataset. "-" denotes the experiment is not conducted due to no machine translation system is available.}
    \label{tab:bloom-nusa}
\end{table}

\begin{table}[!t]
    \centering
    \resizebox{\linewidth}{!}{%
        \begin{tabular}{l|c|c|c|c|c|c|c|c|c|c}
        \toprule
             \textbf{Inference Type} &       \textbf{aym} &       \textbf{bzd} &       \textbf{cni} &        \textbf{grn} &       \textbf{hch} &       \textbf{nah} &       \textbf{oto} &       \textbf{quy} &       \textbf{shp} &       \textbf{tar} \\
        \midrule
        % Zero-Shot & 16.66 & 16.66 & 16.66 & 16.66 & 16.66 & 16.66 & 16.62 & 16.66 & 16.66 & 16.66 \\
        Zero-Shot \\
          \quad Source-Only Label & 16.66 & 16.66 & 16.66 & 16.66 & 16.66 & 16.66 & 16.62 & 16.66 & 16.66 & 16.66 \\
          \quad \quad + Query Alignment & 19.87 & 18.07 & 19.57 & 18.13 & 22.57 & 19.58 & 18.51 & 19.52 & 20.15 & 20.14 \\
          \quad Target-Only Label & 26.88 &       - &       - & 19.52 &       - &       - & 17.86 & 20.62 &       - &       - \\
          \quad Label Alignment & 16.66 &       - &       - & 19.03 &       - &       - & 26.55 & 16.86 &       - &       - \\
        Zero-Shot (MT) & 16.66 &       - &       - & 16.66 &       - &       - &       - & 16.66 &       - &       - \\
        ICL Random & 32.99 & 30.68 & 30.79 & 33.40 & 28.02 & 32.67 & 33.29 & 30.64 & 32.24 & 31.63 \\
        ICL SBERT & 33.55 & 32.84 & 30.51 & 37.08 & 31.85 & 31.17 & 29.74 & 34.62 & 29.82 & 33.05 \\
        ICL SBERT (MT) & 35.80 &       - &       - & 37.79 &       - &       - &       - & 39.19 &       - &       - \\
        X-ICL Random & 32.33 & 28.98 & 31.12 & 30.42 & 33.67 & 30.39 & 30.77 & 30.22 & 33.50 & 26.32 \\
        % X-ICL SBERT & 36.99 & 34.12 & 34.28 & 32.93 & 34.90 & 32.38 & 30.57 & 34.34 & 32.80 & 35.49 \\
        X-ICL SBERT \\
          \quad Source-Only Label & 36.99 & 34.12 & 34.28 & 32.93 & 34.90 & 32.38 & 30.57 & 34.34 & 32.80 & 35.49 \\
          \quad \quad + Query Alignment & 34.07 & 34.69 & 34.29 & 38.55 & 31.72 & 32.85 & 34.15 & 31.02 & 32.94 & 32.67 \\
          \quad Target-Only Label & 36.12 &       - &       - & 28.67 &       - &       - & 32.74 & 31.57 &       - &       - \\
          \quad Label Alignment & 18.41 &       - &       - & 17.48 &       - &       - & 18.35 & 20.55 &       - &       - \\        
        \bottomrule
        \end{tabular}
    }
    \caption{Experiment results for BLOOM-7B1 model on AmericasNLI dataset. "-" denotes the experiment is not conducted due to no machine translation system is available.}
    \label{tab:bloom-anli}
\end{table}

\end{document}